\pdfoutput=1
\RequirePackage{etex}
\documentclass{arxiv}

\usepackage{graphicx}%
\usepackage{multirow}%
\usepackage{amsmath,amssymb,amsfonts}%
\usepackage{amsthm}%
\usepackage{mathrsfs}%
\usepackage[title]{appendix}%
\usepackage{xcolor}%
\usepackage{textcomp}%
\usepackage{manyfoot}%
\usepackage{booktabs}%
\usepackage{algorithm}%
\usepackage{algorithmicx}%
\usepackage{algpseudocode}%
\usepackage{listings}%
\usepackage{siunitx}%
\usepackage{array}      %
\usepackage{caption}    %
\usepackage[percent]{overpic}
\usepackage{tikz}
\usepackage{lineno}
\usepackage{longtable}
\usepackage{placeins}

\usepackage{makecell}

\theoremstyle{plain}%

\theoremstyle{definition}%

\theoremstyle{remark}%

\begin{document}
\setcounter{Maxaffil}{10}

\title{\Large Foundation Models to Unlock Real-World Evidence from Nationwide Medical Claims}

\author[1,$\ast$]{Fan Ma} %

\author[1,$\ast$]{Yuntian Liu} %
\author[1,$\ast$]{Xiang Lan} %
\author[1]{Weipeng Zhou}
\author[1]{Jun Ni}

\author[2]{Mauro Giuffrè} 

\author[1]{Lingfei Qian}
\author[1]{Xueqing Peng}
\author[1]{Yujia Zhou}
\author[1]{Ruey-Ling Weng}
\author[1]{Huan He}
\author[4]{Lu Li}
\author[5]{Huiyuan Wang}
\author[1]{Qingyu Chen}
\author[1]{Andrew Loza}
\author[3]{Laila Rasmy}
\author[3]{Degui Zhi}
\author[1,6]{Yuan Lu}
\author[8]{Chenjie Zeng}
\author[8,9]{Joshua C Denny}
\author[1,7]{Lee Schwamm}
\author[1]{Daniella Meeker}
\author[1]{{Brian Ondov}}
\author[1]{Lucila Ohno-Machado}
\author[5]{Yong Chen}
\author[1,$\dag$]{Hua Xu}

\affil[1]{\normalsize Department of Biomedical Informatics and Data Science, Yale School of Medicine, Yale University, New Haven, CT, USA \authorcr \vspace{0.1cm}}

\affil[2]{\normalsize Department of Medical, Surgical and Health Sciences, Università degli Studi di Trieste, Trieste, Italy \authorcr \vspace{0.1cm}}

\affil[3]{\normalsize McWilliams School of Biomedical Informatics, UTHealth Houston, Houston, TX, USA
 \authorcr}

\affil[4]{\normalsize Applied Mathematics and Computational Science Graduate Group, University of Pennsylvania, Philadelphia, PA, USA \authorcr \vspace{0.1cm}}

\affil[5]{\normalsize Department of Biostatistics, Epidemiology and Informatics, Perelman School of Medicine, University of Pennsylvania, Philadelphia, PA, USA \authorcr \vspace{0.1cm}}

\affil[6]{\normalsize Department of Internal Medicine (Section of Cardiovascular Medicine), Yale School of Medicine, Yale University, New Haven, CT, USA \authorcr \vspace{0.1cm}}

\affil[7]{\normalsize Department of Neurology, Yale School of Medicine, Yale University, New Haven, CT, USA \authorcr \vspace{0.1cm}}

\affil[8]{\normalsize Precision Health Informatics Section, National Human Genome Research Institute, National Institutes of Health, Bethesda, MD, USA 
\authorcr \vspace{0.1cm}}

\affil[9]{\normalsize All of Us Research Program, National Institutes of Health, Bethesda, MD, USA
\authorcr \vspace{0.1cm}}

\affil[$\ast$]{\normalsize Co-first authors\hspace{1cm}}
\affil[$\dag$]{\normalsize Corresponding author\authorcr
Hua Xu: hua.xu@yale.edu}

\begin{abstract}
{Real-world data increasingly inform regulatory and clinical decisions. Administrative claims, among their major sources, continuously track care and its cost across providers throughout enrollment, yet remain largely unexplored for healthcare foundation models. Here we present ReClaim, a family of generative foundation models built on 43.8 billion claims records from more than 200 million enrollees in the MarketScan Research Databases. ReClaim learns diagnoses, procedures, medications and expenditure jointly through pre-training from scratch, then specializes for downstream prediction tasks through post-training, with a training cutoff at the end of 2022. In predicting more than 1,000 diseases, ReClaim achieved a mean ROC-AUC of 0.769, outperforming disease-specific gradient-boosted models (0.663) and a transformer-based baseline (0.694), and retained this advantage when generalizing to two independent cross-domain electronic health record datasets. ReClaim also outperformed gradient-boosted models in forecasting next-year expenditure and in identifying high-need, high-cost enrollees. ReClaim maintained its advantage in disease prediction and expenditure forecasting on post-cutoff data (2023--2024). Beyond prediction, ReClaim representations used as covariates resulted in improved effect estimates compared with algorithm-derived covariates, when benchmarked against results expected from randomized controlled trials. Together, these results establish nationwide insurance claims as a scalable substrate for healthcare foundation models supporting disease prediction, expenditure forecasting and confounding adjustment in real-world evidence studies.}
\end{abstract}

\maketitle

\section{Introduction}
\label{sec1}

Real-world data (RWD) capturing healthcare delivered in routine practice have become indispensable for complementing randomized controlled trials (RCTs) in regulatory evaluation, clinical decision-making, and health policy development~\cite{sherman2016realworld, liu2022real, hernan2016using, concato2022realworld}.
Among sources of RWD, electronic health records (EHRs) and administrative claims data both capture essential records of healthcare encounters, including diagnoses, procedures, and prescriptions, but differ in their structure and scope. 
EHRs provide detailed clinical information, such as laboratory results, imaging, and clinical notes, but records are often incomplete because patients may receive care across multiple health systems and EHRs lack a defined observation window, making unrecorded events difficult to distinguish from true absence and introducing false-negative bias~\cite{fda2024ehrclaims}.
In contrast, administrative claims data lack granular clinical detail~\cite{liu2022real}, {but provide relatively standardized longitudinal records of billed diagnoses, treatments, healthcare utilization and expenditures across large populations, with continuity across providers while individuals remain enrolled in a payer system}~\cite{fda2024ehrclaims}. Defined enrollment periods further anchor {the observable window, although claims remain records of reimbursed care rather than complete representations of a patient's underlying health}. These two complementary and ever-growing data sources position RWD as a powerful basis for modeling longitudinal healthcare trajectories in real-world settings.

Building on this potential, there has been growing interest in developing foundation models trained on large-scale RWD. Early work, including BEHRT~\cite{li2020behrt} and Med-BERT~\cite{rasmy2021medbert}, demonstrated that structured EHRs can be represented as sequences of clinical events forming longitudinal healthcare trajectories, enabling transformer-based models to learn transferable representations and improve disease prediction tasks~\cite{steinberg2021language, yang2023transformehr}. More recent studies~\cite{steinberg2023motor, pang2024cehrgpt, mcdermott2023eventstreamgpt, wornow2023ehrshot} extend this paradigm to generative modeling of health event trajectories, enabling time-to-event forecasting, simulation of longitudinal healthcare pathways, and zero- or few-shot adaptation across outcomes, as exemplified by models such as Foresight~\cite{kraljevic2024foresight} and ETHOS~\cite{renc2024zero}. {Apollo~\cite{zhang2026apollo} extends longitudinal patient representation learning to multimodal hospital records, integrating structured events, clinical text and images.} {Alongside these advances in patient representation, recent studies have examined prediction across many disease endpoints and the effects of scale. Delphi~\cite{shmatko2025learning} models recorded diagnosis events across more than 1,000 three-character International Classification of Diseases, Tenth Revision (ICD-10) categories, approaching the performance of specialized single-disease models despite modest data and model scale. Complementing this evaluation across disease categories, the Curiosity preprint~\cite{waxler2025generative} reports improved prediction across diverse tasks with increasing data and model size.}

Despite these advances, the real-world applicability of foundation models trained on RWD remains constrained by limited {evidence across populations, data sources and tasks}. First, {most models are trained within individual health systems or comparatively modest cohorts, leaving their performance under temporal and population shifts incompletely characterized}~\cite{shmatko2025learning, wornow2023shaky}. Second, the field remains largely EHR-centric: {claims-based modeling is less developed, and evaluations that transfer a model between claims and EHR data remain uncommon}~\cite{zeng2022claimpt, sahu2024lmm}. {Third, evaluations have focused predominantly on disease prediction benchmarks, with less evidence for whether the same learned representation supports financial outcomes or observational RWE studies. Finally, task-specific post-training has received limited study as a means of adapting a shared longitudinal representation to distinct downstream objectives.}

Here we present ReClaim, {a family of healthcare foundation models trained on nationwide claims data}. Developed on 43.8 billion claims records from more than 200 million enrollees in the US MarketScan databases{, with a training cutoff at the end of 2022}, ReClaim models {longitudinal trajectories containing diagnoses, procedures, medications and expenditure}. {The framework combines large-scale next-token pre-training with task-specific post-training. We evaluate three applications: disease prediction across more than 1,000 three-character categories of the ICD-10 Clinical Modification (ICD-10-CM) used in US claims, next-year healthcare expenditure forecasting and identification of high-need, high-cost enrollees, and use of learned representations as covariates for confounding adjustment in observational RWE studies. Disease prediction is assessed retrospectively and prospectively in held-out MarketScan cohorts and through cross-domain evaluation in two independent EHR datasets; expenditure forecasting is assessed retrospectively and prospectively in held-out MarketScan cohorts; and the RWE studies are evaluated in MarketScan and through cross-domain evaluation at YNHH. Across these settings, ReClaim outperforms disease prediction and expenditure baselines and matches or exceeds expert-curated and algorithm-derived covariates in RWE studies, demonstrating that claims-based foundation models can support disease prediction, expenditure forecasting and confounding adjustment across time, data sources and analytical objectives.}

\section{Results}\label{sec:results}

\begin{figure}[!t]
    \centering
    \includegraphics[width=\linewidth]{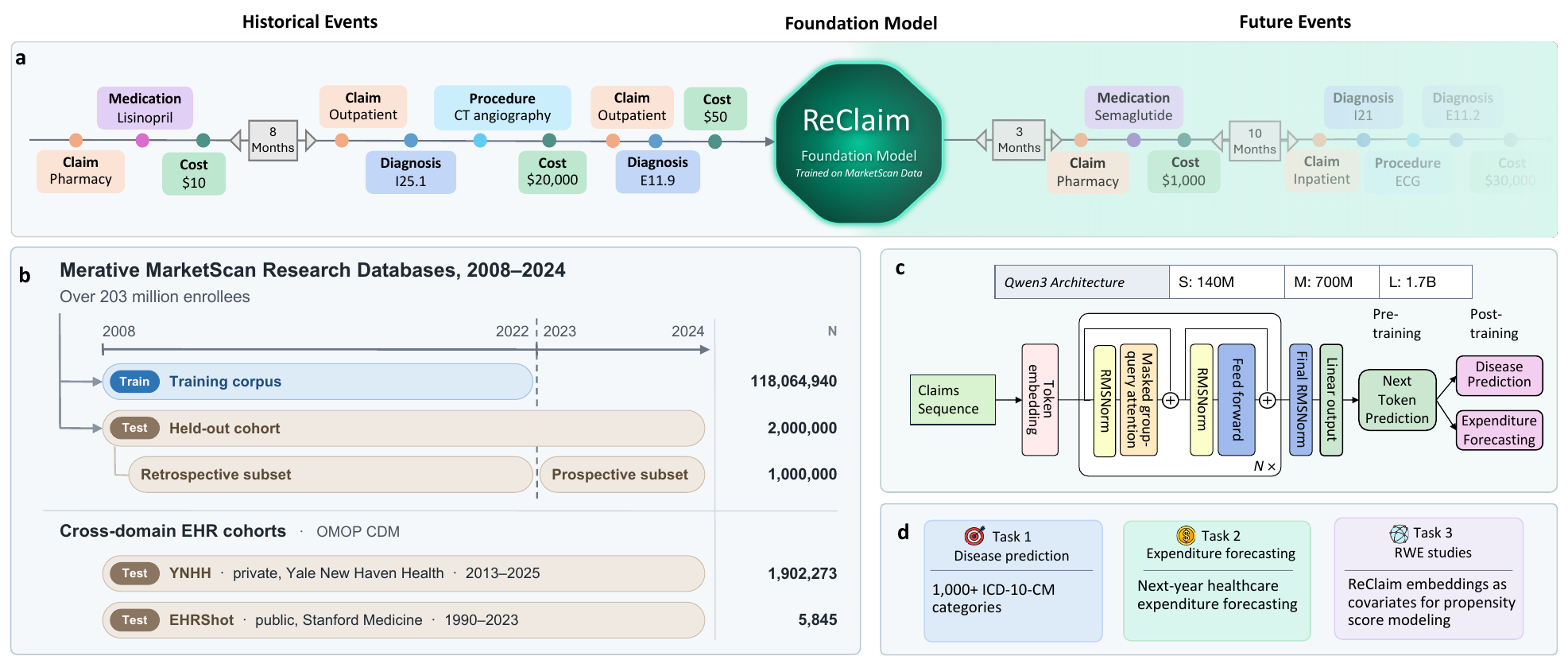}
    \caption{
    \textbf{ReClaim framework and evaluation workflow.} (\textbf{a}), Longitudinal medical events from patient claims are encoded as chronologically ordered trajectories, and ReClaim autoregressively predicts future medical events including diagnoses, procedures, medications, and expenditure. (\textbf{b}),~The study datasets comprise the MarketScan corpus partitioned into a final training set, a held-out internal test cohort for disease and expenditure forecasting with retrospective and prospective temporal subsets, and {two EHR datasets used for cross-domain evaluation: EHRShot and Yale New Haven Health (YNHH). These data sources support retrospective, prospective and cross-domain evaluations.} (\textbf{c}),~Transformer-based ReClaim models (Qwen3 architecture) are trained at three parameter scales through next-token pre-training, followed by task-specific post-training {for disease prediction and for expenditure forecasting}. (\textbf{d}),~ReClaim is evaluated on three downstream tasks: {disease prediction} for over 1,000 ICD-10-CM conditions, next-year healthcare expenditure forecasting, and {confounding adjustment in RWE studies, in which ReClaim embeddings serve as propensity-score covariates in a known-effect validation and in a target trial emulation}.
    }
    \label{fig:framework}
    \vspace{-2.5em}
\end{figure}

\begin{figure}[!t]
    \centering
    \makebox[\linewidth][c]{\small\bfseries Retrospective evaluation (2008--2022)}\par\vspace{0.2em}
    \noindent%
    \tikz[baseline=(img.south west)]{\node[anchor=south west,inner sep=0](img)%
        {\includegraphics[width=0.25\linewidth]{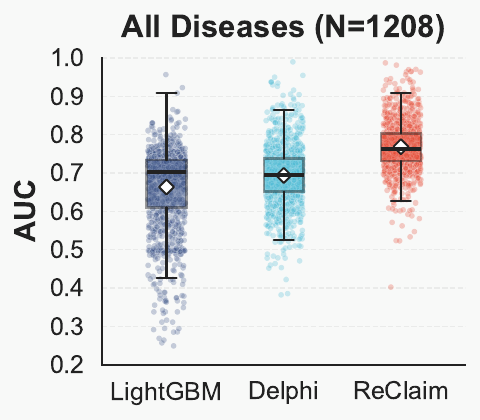}};%
        \node[anchor=north west,font=\large\bfseries] at (img.north west){a};}%
    \tikz[baseline=(img.south west)]{\node[anchor=south west,inner sep=0](img)%
        {\includegraphics[width=0.25\linewidth]{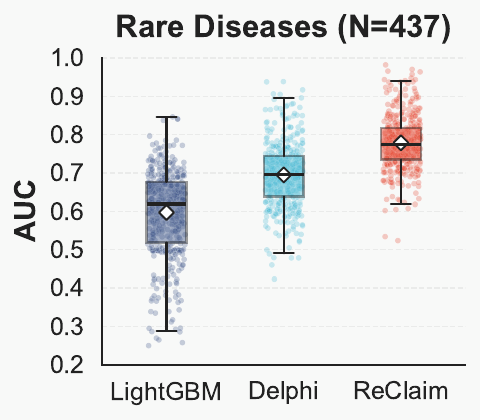}};%
        \node[anchor=north west,font=\large\bfseries] at (img.north west){b};}%
    \tikz[baseline=(img.south west)]{\node[anchor=south west,inner sep=0](img)%
        {\includegraphics[width=0.25\linewidth]{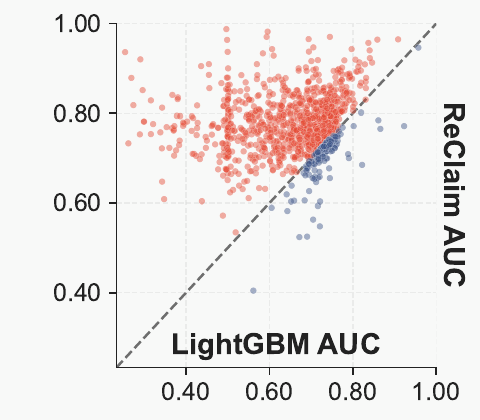}};%
        \node[anchor=north west,font=\large\bfseries] at (img.north west){c};}%
    \tikz[baseline=(img.south west)]{\node[anchor=south west,inner sep=0](img)%
        {\includegraphics[width=0.25\linewidth]{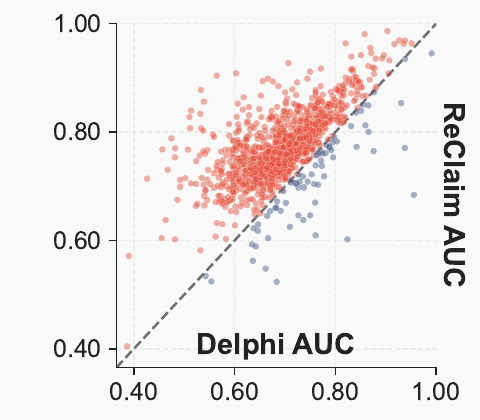}};%
        \node[anchor=north west,font=\large\bfseries] at (img.north west){d};}%
    \par\vspace{0.2em}
    \makebox[\linewidth][c]{\small\bfseries Prospective evaluation (2023 onward)}\par\vspace{0.2em}
    \noindent%
    \tikz[baseline=(img.south west)]{\node[anchor=south west,inner sep=0](img)%
        {\includegraphics[width=0.25\linewidth]{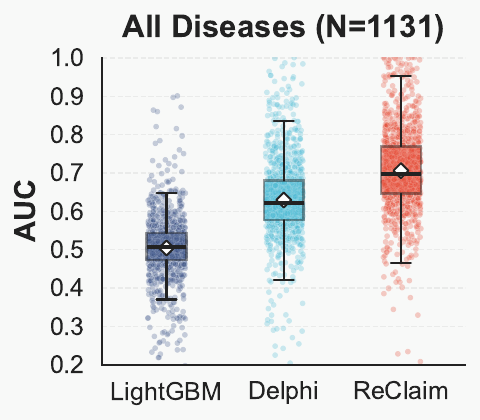}};%
        \node[anchor=north west,font=\large\bfseries] at (img.north west){e};}%
    \tikz[baseline=(img.south west)]{\node[anchor=south west,inner sep=0](img)%
        {\includegraphics[width=0.25\linewidth]{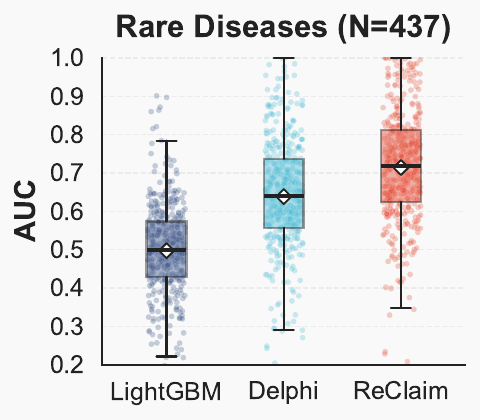}};%
        \node[anchor=north west,font=\large\bfseries] at (img.north west){f};}%
    \tikz[baseline=(img.south west)]{\node[anchor=south west,inner sep=0](img)%
        {\includegraphics[width=0.25\linewidth]{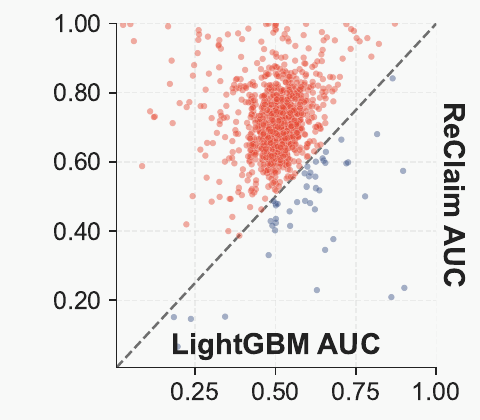}};%
        \node[anchor=north west,font=\large\bfseries] at (img.north west){g};}%
    \tikz[baseline=(img.south west)]{\node[anchor=south west,inner sep=0](img)%
        {\includegraphics[width=0.25\linewidth]{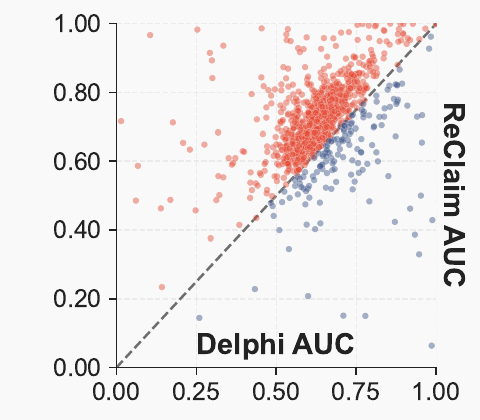}};%
        \node[anchor=north west,font=\large\bfseries] at (img.north west){h};}%
    \caption{{\textbf{ReClaim outperforms LightGBM and Delphi across disease endpoints.} Top row, retrospective (2008--2022); bottom row, prospective (2023 onward). Comparisons are paired disease-level ROC-AUC differences, reported as mean\,\(\Delta\) with 95\% bootstrap confidence intervals (CI) and Holm-adjusted two-sided paired Wilcoxon signed-rank \(\textit{P}\) values; diamonds in box plots mark group means, and in scatter plots points above the diagonal indicate diseases favoring ReClaim (orange) over the comparator. (\textbf{a}),~Disease-level ROC-AUC across 1,208 shared endpoints: ReClaim\,=\,0.769, Delphi\,=\,0.694, LightGBM\,=\,0.663. (\textbf{b}),~Rare-disease subset of 437 endpoints (\(10<n_{\mathrm{diseased}}<500\) per 1 million); ReClaim exceeded LightGBM by +18.17\,pp (95\% CI, 17.07--19.29; \(P_{\mathrm{Holm}}=8.1\times10^{-72}\)) and Delphi by +8.37\,pp (7.76--8.97; \(1.6\times10^{-67}\)). (\textbf{c}),~ReClaim versus LightGBM (87.9\% of endpoints above the diagonal; \(\Delta\)\,=\,+10.51\,pp; 9.86--11.19; \(3.0\times10^{-163}\)). (\textbf{d}),~ReClaim versus Delphi (94.1\%; +7.50\,pp; 7.16--7.84; \(1.1\times10^{-176}\)). \textbf{e--h},~Prospective evaluation (1,131 shared endpoints). (\textbf{e}),~ReClaim\,=\,0.706, Delphi\,=\,0.630, LightGBM\,=\,0.504. (\textbf{f}),~Same rare-disease subset; ReClaim exceeded LightGBM by +21.67\,pp (19.78--23.56; \(7.1\times10^{-60}\)) and Delphi by +7.58\,pp (5.95--9.22; \(3.2\times10^{-25}\)). (\textbf{g}),~ReClaim versus LightGBM (96.0\%; +20.16\,pp; 19.36--20.98; \(1.1\times10^{-172}\)). (\textbf{h}),~ReClaim versus Delphi (87.0\%; +7.64\,pp; 6.93--8.33; \(8.4\times10^{-113}\)).}}
    \label{fig:auc_comp}
\end{figure}

\begin{figure}[!t]
    \centering
    \makebox[\linewidth][c]{\small\bfseries EHRShot}\par\vspace{0.2em}
    \noindent%
    \tikz[baseline=(img.south west)]{\node[anchor=south west,inner sep=0](img)%
        {\includegraphics[width=0.25\linewidth]{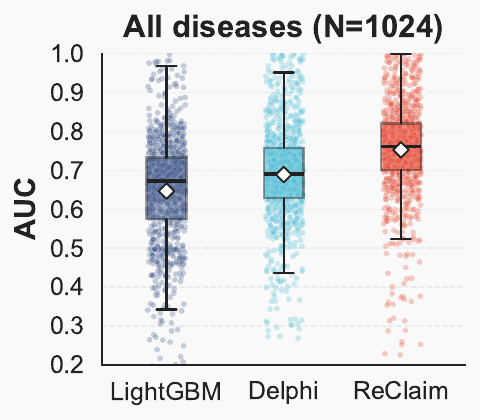}};%
        \node[anchor=north west,font=\large\bfseries] at (img.north west){a};}%
    \tikz[baseline=(img.south west)]{\node[anchor=south west,inner sep=0](img)%
        {\includegraphics[width=0.25\linewidth]{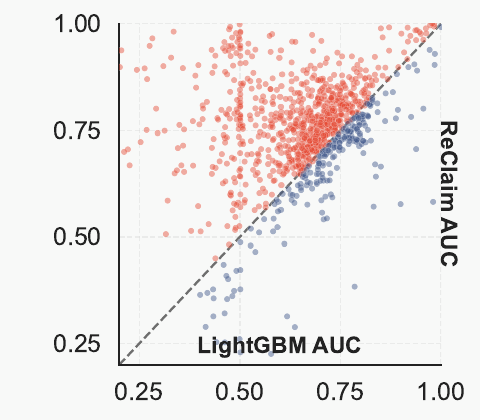}};%
        \node[anchor=north west,font=\large\bfseries] at (img.north west){b};}%
    \tikz[baseline=(img.south west)]{\node[anchor=south west,inner sep=0](img)%
        {\includegraphics[width=0.25\linewidth]{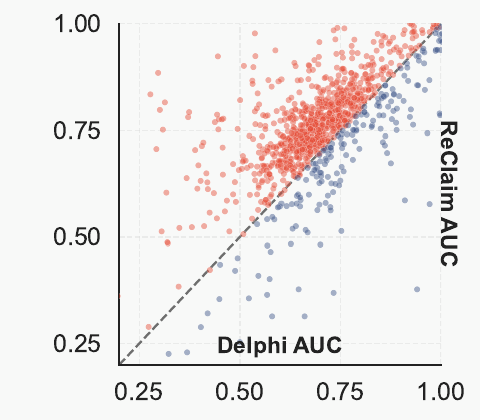}};%
        \node[anchor=north west,font=\large\bfseries] at (img.north west){c};}%
    \tikz[baseline=(img.south west)]{\node[anchor=south west,inner sep=0](img)%
        {\includegraphics[width=0.25\linewidth]{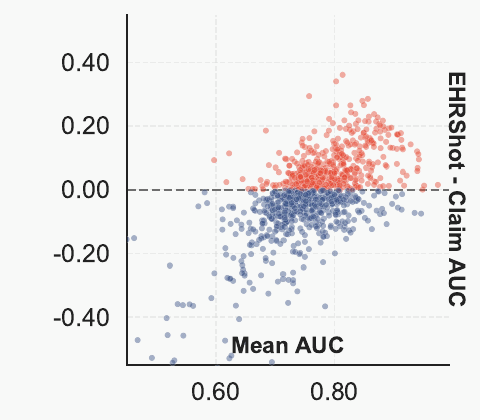}};%
        \node[anchor=north west,font=\large\bfseries] at (img.north west){d};}%
    \par\vspace{0.1em}
    \makebox[\linewidth][c]{\small\bfseries YNHH}\par\vspace{0.1em}
    \noindent%
    \tikz[baseline=(img.south west)]{\node[anchor=south west,inner sep=0](img)%
        {\includegraphics[width=0.25\linewidth]{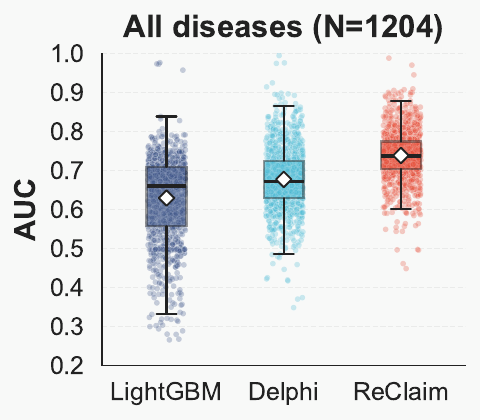}};%
        \node[anchor=north west,font=\large\bfseries] at (img.north west){e};}%
    \tikz[baseline=(img.south west)]{\node[anchor=south west,inner sep=0](img)%
        {\includegraphics[width=0.25\linewidth]{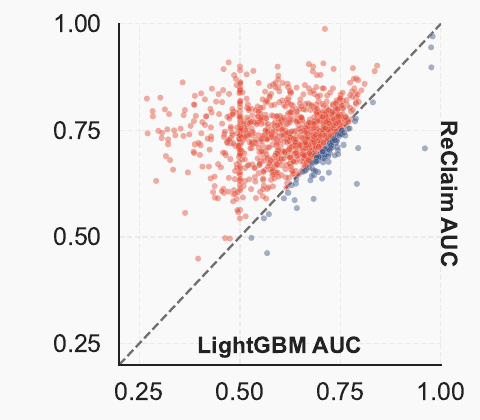}};%
        \node[anchor=north west,font=\large\bfseries] at (img.north west){f};}%
    \tikz[baseline=(img.south west)]{\node[anchor=south west,inner sep=0](img)%
        {\includegraphics[width=0.25\linewidth]{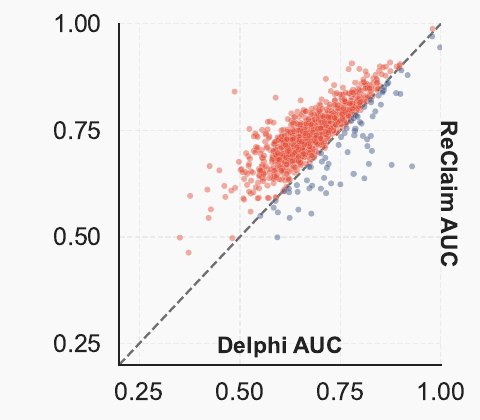}};%
        \node[anchor=north west,font=\large\bfseries] at (img.north west){g};}%
    \tikz[baseline=(img.south west)]{\node[anchor=south west,inner sep=0](img)%
        {\includegraphics[width=0.25\linewidth]{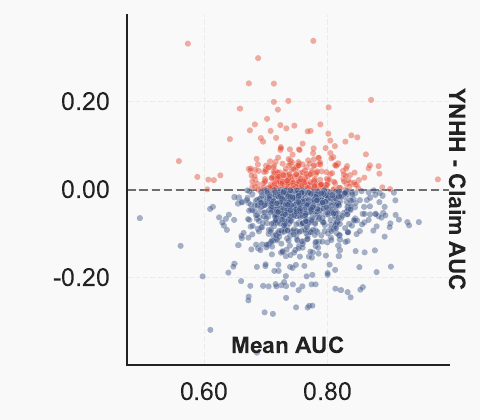}};%
        \node[anchor=north west,font=\large\bfseries] at (img.north west){h};}%
    \caption{{\textbf{Cross-domain evaluation on EHR-based datasets.} Top row, EHRShot; bottom row, YNHH. Paired disease-level ROC-AUC differences are reported as mean\,\(\Delta\) with 95\% bootstrap confidence intervals (CI) and Holm-adjusted two-sided paired Wilcoxon signed-rank \(\textit{P}\) values; orange points above the diagonal in scatter plots favor ReClaim. (\textbf{a}),~Mean ROC-AUC across 1,024 shared EHRShot endpoints: ReClaim\,=\,0.753, Delphi\,=\,0.689, LightGBM\,=\,0.647. (\textbf{b}),~ReClaim versus LightGBM (79.1\% of endpoints above the diagonal; \(\Delta\)\,=\,+10.61\,pp; 95\% CI, 9.56--11.67; \(P_{\mathrm{Holm}}=4.8\times10^{-91}\)). (\textbf{c}),~ReClaim versus Delphi (81.7\%; +6.34\,pp; 5.67--7.01; \(5.4\times10^{-82}\)). (\textbf{d}),~Per-disease ReClaim ROC-AUC difference, EHRShot versus retrospective MarketScan, plotted against mean ROC-AUC (1,024 diseases; \(\Delta\)\,=\,-1.58\,pp; -2.36 to -0.81; \(P_{\mathrm{Holm}}=0.015\)). \textbf{e--h},~Same panels for YNHH (1,204 shared endpoints). (\textbf{e}),~ReClaim\,=\,0.739, Delphi\,=\,0.677, LightGBM\,=\,0.630. (\textbf{f}),~ReClaim versus LightGBM (90.9\%; +10.91\,pp; 10.28--11.55; \(9.0\times10^{-176}\)). (\textbf{g}),~ReClaim versus Delphi (94.1\%; +6.14\,pp; 5.87--6.42; \(4.9\times10^{-169}\)). (\textbf{h}),~Per-disease ReClaim ROC-AUC difference, YNHH versus retrospective MarketScan (1,204 diseases; \(\Delta\)\,=\,-2.97\,pp; -3.36 to -2.59; \(1.3\times10^{-60}\)).}}
    \label{fig:external_validation_auc}
\end{figure}

\subsection{ReClaim Overview and Evaluation Benchmarks}
\enlargethispage{2\baselineskip}
ReClaim converts longitudinal claims records into chronologically ordered patient trajectories that jointly encode enrollment history, clinical events, and healthcare expenditures (Fig.~\ref{fig:framework}a). These trajectories were used to train decoder-only Transformer models from scratch at three parameter scales with a next-token prediction objective, followed by task-specific post-training {for disease prediction and for expenditure forecasting}. {We denote the three model variants as ReClaim-S (140M), ReClaim-M (700M), and ReClaim-L (1.7B) at both stages; unless a size or stage is specified, ReClaim refers to the post-trained ReClaim-L used for the primary disease prediction comparisons.}

ReClaim was first pre-trained on the {records of over 118 million enrollees} collected between 2008 and 2022, and was subsequently evaluated on three downstream task categories across four benchmarks (Fig.~\ref{fig:framework}b; Supplementary Table~\ref{tab:inclusion_criteria}). {The evaluation benchmarks included a one-million held-out MarketScan cohort, consisting of a retrospective subset (2008–2022) and a prospective subset (2023 onward) to assess temporal generalization, as well as two EHR-based cohorts used for cross-domain evaluation: EHRShot~\cite{wornow2023ehrshot} and Yale New Haven Health (YNHH).}
{The three task categories comprised: (1)~\emph{disease prediction} for over 1,000 ICD-10-CM conditions, evaluated across all four benchmarks; (2)~\emph{healthcare expenditure forecasting}, next-year total gross payments and identification of high-need, high-cost enrollees, evaluated retrospectively and prospectively in the held-out MarketScan cohort; and (3)~\emph{RWE studies}, assessing whether ReClaim-derived representations, used as propensity-score covariates, match or exceed conventional covariate specifications for confounding adjustment, evaluated in MarketScan and in a cross-domain evaluation at YNHH.}

\begin{figure}[!ht]
    \centering
    \tikz[baseline=(img.south west)]{\node[anchor=south west,inner sep=0](img)%
        {\includegraphics[width=0.46\linewidth]{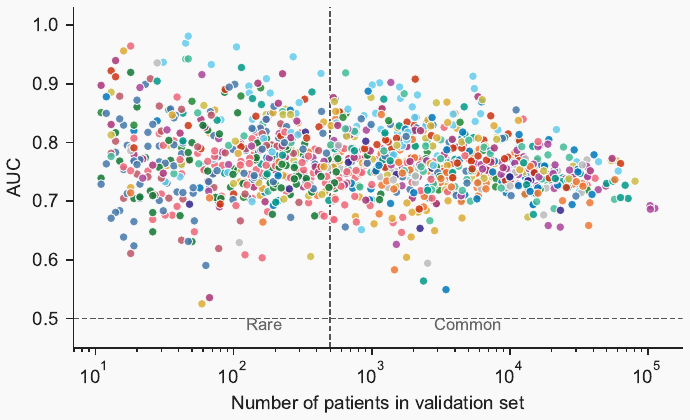}};%
        \node[anchor=north west,font=\large\bfseries] at (img.north west){a};}\hfill%
    \tikz[baseline=(img.south west)]{\node[anchor=south west,inner sep=0](img)%
        {\includegraphics[width=0.52\linewidth]{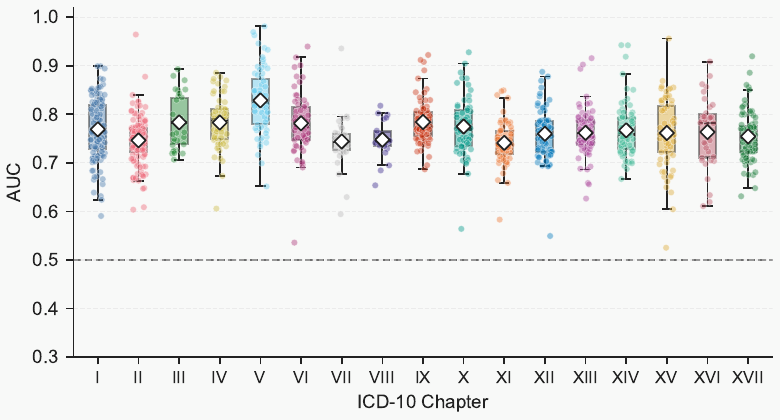}};%
        \node[anchor=north west,font=\large\bfseries] at (img.north west){b};}%
    \par\vspace{0.2em}
    \tikz[baseline=(img.south west)]{\node[anchor=south west,inner sep=0](img)%
        {\includegraphics[width=0.40\linewidth]{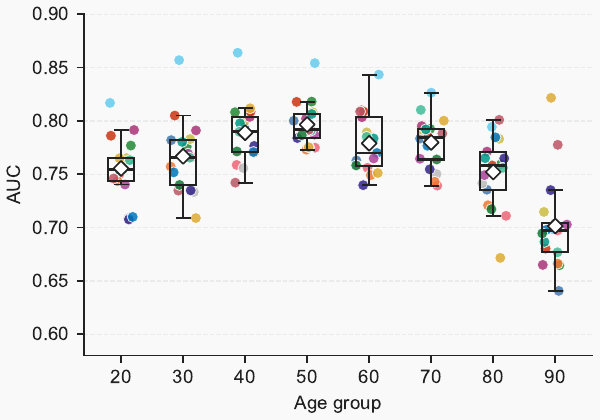}};%
        \node[anchor=north west,font=\large\bfseries] at (img.north west){c};}\hfill%
    \tikz[baseline=(img.south west)]{\node[anchor=south west,inner sep=0](img)%
        {\includegraphics[width=0.22\linewidth]{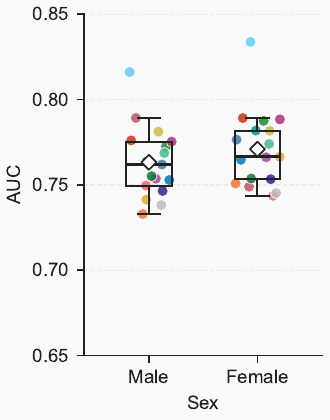}};%
        \node[anchor=north west,font=\large\bfseries] at (img.north west){d};}\hfill%
    \tikz[baseline=(img.south west)]{\node[anchor=south west,inner sep=0](img)%
        {\includegraphics[width=0.34\linewidth]{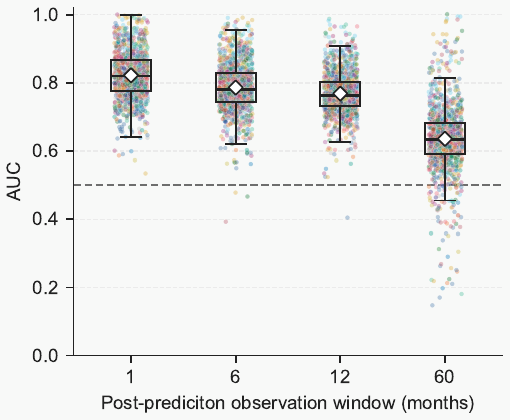}};%
        \node[anchor=north west,font=\large\bfseries] at (img.north west){e};}%
    \par\vspace{0.2em}
    \includegraphics[width=0.95\linewidth]{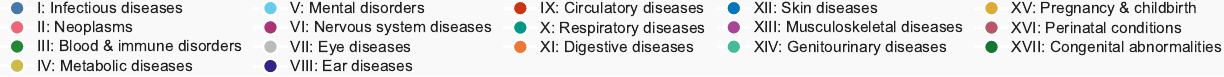}
    \caption{{\textbf{Discrimination performance across disease prevalence, clinical domain, demographic, and temporal strata.} Each point is one disease endpoint colored by ICD-10-CM chapter, and box plots show overlaid individual disease ROC-AUC values with diamonds marking group means. (\textbf{a}),~Disease-level ROC-AUC versus the number of diseased patients in the retrospective held-out set across 1,141 endpoints; dashed lines mark the chance level (ROC-AUC\,=\,0.5) and the rare-disease boundary. Overall mean ROC-AUC\,=\,0.768 (median\,=\,0.762); 91.4\% of endpoints exceed ROC-AUC 0.70 and 25.1\% exceed ROC-AUC 0.80. Rare endpoints (\(n=433\); 0.778) outperform common endpoints (\(n=708\); 0.761). (\textbf{b}),~ROC-AUC distributions across all 17 ICD-10-CM chapters; strongest, V. Mental Disorders (0.829), IX. Circulatory (0.784), III. Blood and Immune Disorders (0.784) and IV. Metabolic (0.783); weakest, XI. Digestive (0.741), VII. Eye (0.744) and II. Neoplasms (0.746). (\textbf{c}),~Category-level mean ROC-AUC by age group (20s--90s), peaking in the 50s (0.797) and declining in the oldest groups (80s, 0.752; 90s, 0.701). (\textbf{d}),~Category-level mean ROC-AUC by sex (female, 0.771; male, 0.763). (\textbf{e}),~Disease-level ROC-AUC across post-prediction observation windows of 1, 6, 12 and 60 months, declining monotonically: 0.822, 0.786, 0.769 and 0.636.}}
    \label{fig:auc_distribution}
\end{figure}

\subsection{{Disease Prediction}}
\label{sec:dp}

{Disease prediction was assessed using each patient's historical trajectory to generate scores for target three-character ICD-10-CM categories first recorded after the prediction time.} ReClaim followed the evaluation protocol of Delphi~\cite{shmatko2025learning} and was benchmarked against both Delphi~\cite{shmatko2025learning} and 1,208 disease-specific LightGBM models~\cite{ke2017lightgbm}. Rare diseases were defined using a prevalence threshold of no more than 5 per 10,000 individuals~\cite{nguengang2020estimating}. {Here, rarity is defined by the observed frequency of each three-character ICD-10-CM category in the evaluation cohort.} As not all datasets contained all endpoints, the number of evaluable diseases varied across the retrospective held-out, prospective temporal, and cross-domain EHR-based evaluations. Performance in each test set was summarized by disease-level ROC-AUC. {Cross-model comparisons used the same evaluable endpoints and shared endpoint-specific case--control cohorts constructed before model inference; endpoint-level auditing confirmed identical case, control and evaluable-stratum counts across all three methods (Supplementary Table~\ref{tab:endpoint_harmonization}).}

\noindent\textbf{Retrospective evaluation (2008--2022)}
In the 1-million retrospective held-out cohort, ReClaim showed a broad improvement over both baselines, with the highest disease-level ROC-AUC distribution across 1,208 shared endpoints (Fig.~\ref{fig:auc_comp}a). The gain was especially evident for rare diseases, where LightGBM degraded more sharply and ReClaim retained discrimination across low-prevalence endpoints (Fig.~\ref{fig:auc_comp}b). Paired disease-level analyses showed that these improvements were not confined to a small subset of conditions: ReClaim outperformed LightGBM and Delphi across most endpoints, with gains distributed over the full performance range (Fig.~\ref{fig:auc_comp}c,d). {These findings were supported by sensitivity analyses: ReClaim-L largely retained its discrimination after excluding related diabetes codes (Supplementary Table~\ref{tab:icd_related_exclusion}) and outperformed LightGBM on all 12 Phecode tasks~\cite{wei2017evaluating} (Supplementary Table~\ref{tab:phecode_sensitivity}).} {Complementing these predictive results, ReClaim disease embeddings aligned more closely with ICD-10-CM chapters than Delphi's under both UMAP and \textit{t}-SNE projections (Supplementary Figs.~\ref{fig:embedding_umap} and~\ref{fig:embedding_tsne} and Supplementary Table~\ref{tab:clustering_metrics}).}

\noindent\textbf{Prospective evaluation (2023 onward)}
The advantage persisted. In the post-cutoff cohort, ReClaim achieved the strongest overall discrimination across the shared disease endpoints (Fig.~\ref{fig:auc_comp}e--h). The separation from LightGBM was larger in this prospective setting, while the comparison with Delphi remained consistently favorable. Together, the retrospective and prospective analyses indicate that claims-scale pre-training followed by {post-training for disease prediction} yields disease-risk estimates that are both more accurate and more temporally robust than supervised endpoint-specific models and the disease-sequence language-model comparator.

\noindent\textbf{{Cross-domain evaluation}}
{We next performed cross-domain evaluation on two EHR-derived cohorts after harmonizing records into the same trajectory format.} On both EHRShot~\cite{wornow2023ehrshot} and YNHH, ReClaim-L retained the highest disease-level ROC-AUC distributions among the three models (Fig.~\ref{fig:external_validation_auc}a,e). Paired comparisons remained favorable against LightGBM and Delphi in both cohorts (Fig.~\ref{fig:external_validation_auc}b,c,f,g). Absolute performance was attenuated relative to in-domain claims testing (Fig.~\ref{fig:external_validation_auc}d,h). These findings suggest that the learned claims representations capture portable longitudinal disease structure while still reflecting expected dataset shift between administrative claims and EHR data. Two compounding shifts account for this gap: {hospital-based recruitment produces marked prevalence enrichment in EHRShot and modest enrichment in YNHH, with median per-disease prevalence 3.04\,$\times$ and 1.12\,$\times$ that of MarketScan, respectively (Supplementary Table~\ref{tab:prevalence_comparison}); EHR records also} differ from claims in coding granularity, capture mechanism and event timing, together altering the trajectory distribution at inference.

\noindent\textbf{Discrimination across disease, demographic, and temporal strata}
We further examined whether discrimination varied with endpoint prevalence, clinical domain, patient demographics and prediction horizon (Fig.~\ref{fig:auc_distribution}a--e). Across 1,141 evaluable diseases, ReClaim-L achieved consistently useful discrimination, with most endpoints exceeding an ROC-AUC of 0.70{; rare diseases showed greater variability, whereas common endpoints converged to a narrower high-performance band} (Fig.~\ref{fig:auc_distribution}a). {We also report PR-AUC for ten selected rare disease categories with high ROC-AUC (Supplementary Table~\ref{tab:rare_disease_prauc}). ReClaim-L achieved higher PR-AUC than LightGBM for all ten categories.}

{Discrimination varied across ICD-10-CM chapters, with the highest mean ROC-AUC for mental disorders and the lowest for digestive diseases (Fig.~\ref{fig:auc_distribution}b). Performance was similar between sexes but lower in the oldest age groups (Fig.~\ref{fig:auc_distribution}c,d), and declined with longer prediction horizons, remaining above chance for most diseases at five years (Fig.~\ref{fig:auc_distribution}e).}

\noindent\textbf{{Model scaling and post-training}}
{On retrospective MarketScan, scaling from ReClaim-S to ReClaim-L modestly increased mean ROC-AUC by 0.57 percentage points before post-training and 2.53 percentage points after post-training (Supplementary Fig.~\ref{fig:scaling_and_posttraining}). Under the same data and training protocol, post-trained GPT-2 and LLaMA-3.1 performed similarly to ReClaim-S but below ReClaim-L (Supplementary Table~\ref{tab:add_baselines}).}

{Task-specific post-training consistently improved discrimination across GPT-2, LLaMA-3.1, ReClaim-S and ReClaim-L (Supplementary Table~\ref{tab:add_baselines}). For ReClaim-L, these gains extended to nearly all evaluated disease endpoints (Supplementary Fig.~\ref{fig:scaling_and_posttraining}). An ablation of post-training sample composition showed that moderate rare-disease resampling gave the highest mean ROC-AUC overall and for rare endpoints among the strategies tested (Supplementary Table~\ref{tab:posttrain_ablations}). On EHRShot, post-trained ReClaim-S outperformed both direct logit inference and Monte Carlo simulation using its pre-trained checkpoint (Supplementary Fig.~\ref{fig:inference_comparison}). ReClaim also outperformed ETHOS on their shared disease categories (Supplementary Table~\ref{tab:ethos_ehrshot}).}

\begin{figure}[!t]
    \centering
    \includegraphics[width=\linewidth]{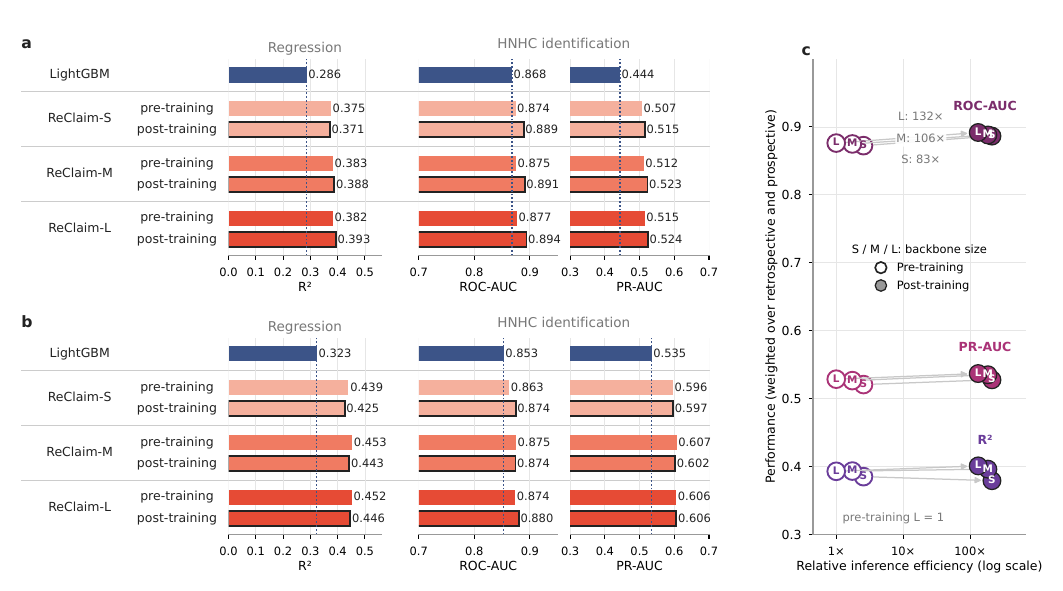}
    \caption{{\textbf{ReClaim surpasses LightGBM in next-year expenditure forecasting, and post-training matches pre-training Monte Carlo simulation at a fraction of its inference cost.}
    Panels \textbf{a} and \textbf{b} show regression of next-year total expenditure ($R^2$) and identification of high-need, high-cost (HNHC) enrollees with a next-year total of at least \$30,000 (ROC-AUC, PR-AUC) in the held-out MarketScan cohort, for the LightGBM baseline fitted on the one-year feature window (dotted line) and for ReClaim-S, -M and -L read out from the pre-trained backbone by 64 Monte Carlo simulations or from the post-trained backbone by a single one-token decode (post-trained inference, outlined bars). (\textbf{a}),~Retrospective evaluation (2008--2022, $n = 356{,}722$). (\textbf{b}),~Prospective evaluation (history through 2022, outcome year 2023, $n = 64{,}658$). (\textbf{c}),~Performance (weighted average of the two evaluations) against inference efficiency relative to Monte Carlo simulation with pre-trained ReClaim-L (ratio of GPU-hours per full evaluation of the held-out cohort on the same hardware and inference engine; log scale). Hollow markers, Monte Carlo simulation; filled markers, post-trained inference; letters, backbone size; arrows, efficiency gain from post-training. Values for all backbone sizes and GPU-hours are given in Supplementary Tables~\ref{tab:expenditure_full_results} and \ref{tab:inference_cost}.}}
    \label{fig:cost_combined}
\end{figure}

\subsection{Healthcare Expenditure Forecasting}
\label{sec:cp}

{Expenditure forecasting was assessed in the 1-million held-out cohort in two settings, regression of next-year total expenditure and identification of high-need, high-cost (HNHC) enrollees at a \$30,000 cutoff, about the 95th percentile of annual spending both nationally (Supplementary Table~\ref{tab:healthcare_expenditures_percentile}) and in our cohort (HNHC prevalence 5.2\%; Supplementary Table~\ref{tab:discretization}). ReClaim here denotes the post-trained model, compared with a LightGBM baseline~\cite{osawa2020cost} using the one-year feature window, which matched or exceeded its full-history version on $R^2$, ROC-AUC and PR-AUC (Methods; Supplementary Table~\ref{tab:expenditure_full_results}).}

\noindent\textbf{Retrospective evaluation (2008--2022)}
In the retrospective subset, {post-trained} ReClaim{-L} outperformed the LightGBM baseline across {both} expenditure tasks (Fig.~\ref{fig:cost_combined}a). For total expenditure regression, ReClaim-L achieved higher explained variance than LightGBM {($R^2$ 0.393 versus 0.286)}. The same pattern extended to risk stratification: ReClaim-L improved HNHC identification{, raising ROC-AUC from 0.868 to 0.894 and PR-AUC from 0.444 to 0.524}.

\noindent\textbf{Prospective evaluation (2023 onward)}
The advantage persisted under temporal generalization. {The prospective HNHC evaluation matches the design of the reference study~\cite{osawa2020cost}, and our LightGBM classifier performed in line with its gradient-boosted tree classifier (ROC-AUC 0.853 versus a reported c-statistic of 0.84), supporting the validity of our baseline implementation. ReClaim-L again outperformed this baseline across regression and HNHC identification (Fig.~\ref{fig:cost_combined}b), with $R^2$ 0.446 versus 0.323, ROC-AUC 0.880 versus 0.853 and PR-AUC 0.606 versus 0.535. These gains suggest} that sequence pre-training captured longitudinal utilization patterns associated with future high-cost care.

\noindent\textbf{{Performance} across payer strata}
Stratified analyses showed that the superiority of {ReClaim} over supervised modeling for expenditure forecasting was not limited to the overall population, but extended across payer groups, where ReClaim improved performance in all strata {(Supplementary Table~\ref{tab:expenditure_by_payer})}. Scaling trends were generally preserved across strata, but remained smaller than the overall gains over LightGBM.

\noindent{\textbf{Model scaling and post-training gains}}
{Performance improved consistently from ReClaim-S to ReClaim-L in both evaluations, indicating modest but systematic scaling benefits beyond the overall separation from LightGBM (Fig.~\ref{fig:cost_combined}a,b). We compared post-trained inference with Monte Carlo simulation using the same backbone and 64 future trajectories, the count at which performance plateaued in our ablation over the number of simulations (Supplementary Fig.~\ref{fig:mc_ablation}, Supplementary Table~\ref{tab:mc_trajectory_ablation}). Post-trained inference matched (within 0.005) or outperformed Monte Carlo simulation on ROC-AUC, PR-AUC and $R^2$ in both evaluations, except that it trailed slightly on prospective $R^2$ (Supplementary Table~\ref{tab:expenditure_full_results}). Post-trained inference generates one token per enrollee against about 16,800 for 64 simulated trajectories, and on identical hardware and inference engine it required 83 to 132 times less inference GPU time across backbone sizes (Fig.~\ref{fig:cost_combined}c; Supplementary Table~\ref{tab:inference_cost}). Post-trained ReClaim-S largely matched pre-trained ReClaim-L at less than 0.5\% of its inference time (0.44 versus 93.4 GPU-hours).}

\begin{figure}[!t]
    \centering
    \includegraphics[width=\linewidth]{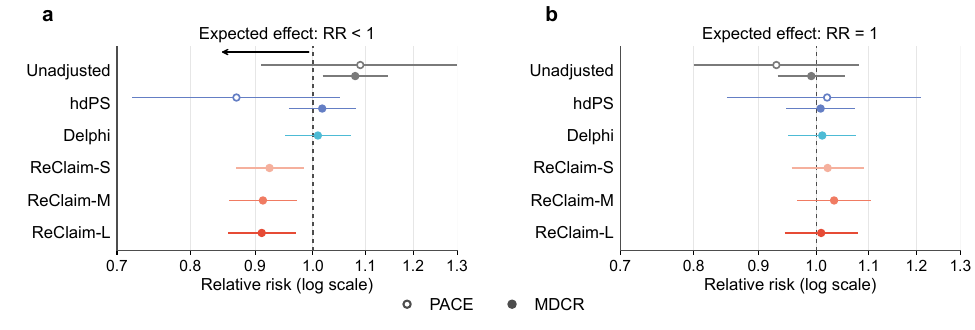}
    \caption{{\textbf{ReClaim representations as propensity-score covariates in a known-effect validation.} (\textbf{a}),~Positive-control validation, cyclooxygenase-2 (COX-2) inhibitors versus nonselective NSAIDs and the risk of gastrointestinal complications, among all MarketScan Medicare Supplemental (MDCR) enrollees ($n = 977{,}551$). Randomized controlled trials show a protective effect, which confounding conceals in the unadjusted estimate; the arrow marks the expected direction (relative risk below 1). (\textbf{b}),~Negative-control validation, influenza vaccination and the risk of hip fracture ($n = 3{,}040{,}767$); the expected relative risk is 1 (dashed line). In both panels, each row is one propensity-score specification; filled circles are our MDCR estimates and hollow circles the reference study's published unadjusted and hdPS estimates in the Pennsylvania Pharmaceutical Assistance Contract for the Elderly (PACE) cohort~\cite{schneeweiss2009high}; bars are 95\% confidence intervals of PS-decile-adjusted relative risks. Every ReClaim specification pairs the embedding with age and sex. hdPS, high-dimensional propensity score; PS, propensity score.}}
    \label{fig:rwe_headline}
\end{figure}

\subsection{Support for Real-World Evidence {Studies}}

{We evaluated whether ReClaim representations can replace algorithm-derived or expert-curated covariates for confounding adjustment in observational RWE studies, through a known-effect validation and a replication of a published target trial emulation on two data sources.}

\noindent{\textbf{Known-effect validation}}
{We replicated the two of the three example studies of the seminal high-dimensional propensity-score (hdPS) study~\cite{schneeweiss2009high} that are feasible in MarketScan data, in all MarketScan MDCR enrollees (Methods; Supplementary Tables~\ref{tab:baseline_controls} and \ref{tab:benchmark_controls}), as a positive control with a protective effect expected from randomized trials~\cite{moore2005tolerability}, cyclooxygenase-2 (COX-2) inhibitors versus nonselective NSAIDs and the risk of gastrointestinal complications, and a negative control with an expected null association, influenza vaccination and the risk of hip fracture. The unadjusted positive-control estimate reproduced the channeling bias of the reference study (relative risk (RR) 1.080, 95\% CI 1.018--1.146; published 1.09, 0.91--1.30), and adjustment reduced it monotonically from hdPS (1.017, 0.957--1.081) and Delphi (1.009, 0.951--1.072) through ReClaim-S and -M to ReClaim-L (0.911, 0.857--0.969) (Fig.~\ref{fig:rwe_headline}a). The ReClaim specifications were the only ones to cross into the protective range with intervals excluding 1. On the negative control every specification preserved the null, with all 95\% CIs spanning 1 (Fig.~\ref{fig:rwe_headline}b). These results held when dimensionality was matched. Compressed by principal component analysis to the dimensions of Delphi (120) and hdPS (500), retaining 82--85\% and 93--96\% of their variance, every ReClaim representation not only preserved the null on the negative control but also still crossed into the protective range with intervals excluding 1 on the positive control, which its dimension-matched baseline did not (Supplementary Table~\ref{tab:t10_pca_compressions}). The results also held when the representations served as outcome models. In a doubly robust analysis, a ReClaim outcome model moved the estimates of the hdPS and Delphi propensity models, which alone stayed near the null, into the protective range with intervals excluding 1, whereas an identically fitted Delphi outcome model did not (Methods; Supplementary Table~\ref{tab:doubly_robust} and Supplementary Fig.~\ref{fig:doubly_robust}). Every combination preserved the null on the negative control, and with a ReClaim outcome model the estimates were stable across propensity models, consistent with double robustness.}

\noindent{\textbf{Target trial emulation}}
{We replicated a recent target trial emulation~\cite{tang2025association} of glucagon-like peptide-1 receptor agonists (GLP-1RA) versus sodium--glucose cotransporter-2 inhibitors (SGLT2i) and versus dipeptidyl peptidase-4 inhibitors (DPP4i) on the MarketScan 2-million held-out split and the Yale New Haven Health (YNHH) EHR, following the reference study's protocol and weighting estimator and varying only the propensity-score specification (Methods; Supplementary Tables~\ref{tab:emulation_cohorts}, \ref{tab:rwe_application_reference_spec} and \ref{tab:rwe_application_definition}). For the two powered outcomes, anxiety and depression, the ReClaim estimates agreed with those of the Expert specification, the reference study's covariate set applied to our data, on both contrasts and both data sources (Supplementary Table~\ref{tab:t4_outcomes_both_sources}). Across specifications, our estimates mostly agreed with the reference study's published estimates, although the anxiety association was attenuated on the YNHH EHR. On the contrast-level diagnostics, the ReClaim specifications mostly had expected absolute systematic error (EASE) comparable to or lower than that of the Expert and hdPS specifications, and had larger effective sample sizes and smaller maximum weights than both on every contrast and source (Supplementary Table~\ref{tab:t5_report_card}). Under weighting, the ReClaim specifications balanced nearly all of their own covariates. On the reference study's covariates, they raised the share with a standardized mean difference below 0.1 from 54--66\% before adjustment to 82--94\%, though below the share under the Expert and hdPS specifications (Supplementary Table~\ref{tab:balance_by_specification}). In the propensity-score matching sensitivity analysis, ReClaim-L again agreed with the Expert specification (Supplementary Table~\ref{tab:t4b_psm_sensitivity}). Overall, ReClaim representations performed comparably to expert-curated covariates across claims and EHR data.}

\section{Discussion}

{In this study, we present ReClaim, a family of healthcare foundation models that are trained on nationwide medical claims data and jointly model diagnoses, procedures, medications and expenditure.} {Through large-scale pre-training and task-specific post-training, ReClaim supports diverse downstream tasks and generalizes across time periods and data sources.}

{In disease prediction, the standard benchmark for healthcare foundation models, ReClaim outperformed disease-specific LightGBM models and the transformer-based Delphi model across more than 1,000 endpoints, with gains that were consistent in retrospective and prospective evaluations and persisted across two independent EHR datasets.} {Beyond this benchmark, ReClaim extended to two tasks largely unexplored for healthcare foundation models, leveraging the strengths of claims data.} {Serving as both regressor and classifier in a single model, it improved over the respective LightGBM baselines in total expenditure forecasting and HNHC identification.} {As propensity-score covariates in RWE studies, ReClaim representations moved the estimate in the expected direction of a protective effect concealed by confounding and preserved an expected null, and performed comparably to expert-curated and algorithm-derived covariates in a target trial emulation on claims and EHR cohorts.}

{These results reflect the complementary contributions of large-scale pre-training, model scale and task-specific post-training.} {During pre-training, performance generally improved with model scale, though modestly.} {Post-training further aligned the models with downstream objectives, improving their performance in both disease prediction and expenditure forecasting relative to their pre-trained checkpoints with substantially less data than pre-training, and the scaling trend persisted after post-training.} {Compared with Monte Carlo simulation, task-specific post-training substantially reduced inference cost in both disease prediction and expenditure forecasting (Supplementary Fig.~\ref{fig:inference_comparison}; Fig.~\ref{fig:cost_combined}c).} {In expenditure forecasting, post-trained ReClaim-L required 132-fold less inference GPU time than its pre-trained counterpart.}
{This saving outweighs the one-time cost of post-training, since post-training plus a single evaluation took 10.8 GPU-hours, about 12\% of the 93.4 GPU-hours of the pre-trained model's Monte Carlo simulation-based evaluation, and the advantage grows with each additional use.}

{Beyond the model and its training, performance also depended on the data and how they were represented.} {While claims provide standardized records across providers during enrollment, offering continuity that is often difficult to achieve with fragmented EHRs, they are sparse, administratively noisy and coded in heterogeneous vocabularies.} {Three design choices addressed these properties.} {Anchor tokens provided absolute temporal reference points and encoded elapsed time during periods without observed healthcare events, allowing the model to learn evolving risk and utilization patterns during both event-rich and clinically silent intervals.} {Given the administrative noise and imperfect temporal precision of claims, monthly aggregation provided a practical balance between noise reduction and temporal fidelity and outperformed daily and weekly alternatives (Supplementary Fig.~\ref{fig:encoding_comparison}).} {Hierarchical ICD encoding has precedents in ETHOS~\cite{renc2024zero} and Curiosity~\cite{waxler2025generative}; we use it as a vocabulary-design choice rather than a methodological novelty. Hierarchical standardization and mapping compressed a raw vocabulary of more than 180,000 ICD-9-CM, ICD-10-CM, NDC and CPT codes while preserving traceability to original codes, rather than excluding long-tail codes to reduce sparsity, at the cost of modestly longer sequences.}

Several limitations should be considered. {First, data limitations arise at three levels, from the raw claims to their representation in ReClaim and their harmonization with EHR data.} {Claims data lack granular clinical information, including laboratory results, imaging findings and narrative notes, which may limit performance for outcomes requiring high-resolution clinical features.} {Claims are generated for billing and reimbursement, and therefore reflect coding practices and financial incentives that may introduce systematic bias and reporting latency.} {Although MarketScan insurance claims have nationwide geographic coverage, the data are not nationally representative, since uninsured individuals are not captured.} {MarketScan is also organized around enrollment episodes rather than complete life courses, so recorded enrollees may not correspond exactly to unique individuals observed over time.} {These properties also shaped the representation choices used to construct ReClaim sequences, which trade information for tractability.} {For example, events are aggregated to monthly timesteps, medications are represented at the level of active ingredients, and continuous expenditure is discretized into finite cost tokens.} {Our ablations do not establish that these representation choices are optimal across tasks, and the small discretization error observed for annual expenditure may not extend to other thresholds or prediction horizons.} {Cross-domain harmonization adds a further layer of approximation.} {Mapping EHR records into the ReClaim schema requires claims-specific variables to be imputed or marked missing, so cross-domain performance reflects both what the model transfers and how faithfully this mapping represents the EHR data.}

{Second, the task definitions limit what the evaluations can establish and how their results can be interpreted.} {For disease prediction, the first recorded diagnosis remains a proxy for documented disease rather than biological onset.} {The primary prediction targets are major ICD-10-CM categories defined by three-character codes, which can encompass clinically distinct diseases or subtypes that the current prediction formulation and evaluation do not resolve, and controls may carry related diagnoses outside the queried category.} {This limitation applies to both common and rare conditions: category-level performance does not establish prediction of a specific subtype or individual rare disease, which would require separate, clinically validated labels and evaluations.} {A sensitivity analysis across 12 clinically curated Phecode tasks~\cite{wei2017evaluating} supported the findings beyond the primary ICD grouping (Supplementary Table~\ref{tab:phecode_sensitivity}), but these phenotypes do not exhaust the endpoint space, and both ICD and Phecode labels remain record-derived rather than independently adjudicated disease states.} {For the rarest endpoints, small event counts yield unstable estimates, DeLong confidence intervals for ROC-AUC may be unreliable, and improvements in discrimination alone do not establish clinical usefulness at a particular decision threshold.} {Post-training performance may also depend on the number and composition of prompt--response pairs; our sampling ablations do not establish an optimal configuration.} {For expenditure forecasting, the target reflects payer and plan design as much as health need, which limits transportability to payer systems with different payment rules.} {Expenditure also shifts with policy and market conditions, and our targets were not adjusted for inflation.} {For the RWE studies, the findings should be interpreted with caution.} {Both replications were assessed mainly by the direction of the estimated effects, and each faced a data constraint.} {The known-effect validation indexes in 2011--2014, the earliest window our data allow, when celecoxib was the only COX-2 inhibitor on the market, unlike in the reference study; in the target trial emulation, four of the six outcomes are event-limited and are not interpreted as findings.} {More generally, as in any observational study, residual confounding cannot be fully excluded, and diagnostics may be sensitive to analytic choices.} {Adequate overlap also does not guarantee complete covariate balance or confounding control.} {Finally, our exploration is confined to using the representations as covariates in conventional confounding adjustment.}

These limitations point to important directions for future research. Integrating claims-based representations with complementary data sources, including EHRs, imaging, and genomics, may enable the development of multimodal healthcare foundation models that capture both breadth and depth of health and healthcare. {The finer-grained diagnostic information already present in the training trajectories could support extending post-training and evaluation to full-length ICD codes and clinically curated subtype-specific endpoints, enabling assessment of disease distinctions beyond the current major-category targets.} {In RWE studies, our results suggest a possible role for foundation-model representations as covariates in confounding adjustment, but they remain preliminary.} {Future work could integrate foundation models more systematically into established causal inference frameworks.} {Such integration would require more extensive validation before routine use in real-world evidence generation.} Advances in model interpretability and fairness will also be critical for safe and equitable deployment in clinical and research settings.

{In conclusion, with further development and validation, claims-based foundation models could ultimately inform healthcare decisions at scale by prioritizing individuals for diagnostic evaluation or earlier screening, including those whose care is fragmented across providers; identifying patients at risk of high future healthcare needs and costs, so that care management and value-based care programs can intervene early~\cite{blumenthal2016caring} and payers and health systems can plan resources; and supporting confounding adjustment in studies that generate real-world evidence for clinical and regulatory decisions.} {ReClaim lays the groundwork for realizing this potential.}

\section{Methods}\label{sec2}

\subsection{Training and Evaluation Data}

We utilized the Merative MarketScan Research Databases (2008–2024), one of the largest longitudinal US administrative claims resources for healthcare research. The dataset contains person-level claims capturing healthcare utilization, expenditures, and enrollment across inpatient, outpatient, and outpatient pharmacy settings, comprising approximately 43.8 billion records. This study includes three primary payer databases: Commercial (CCAE), Medicare {Supplemental} (MDCR), and Medicaid (MDCD), collectively representing over 200 million unique enrollees (Supplementary Table~\ref{tab:marketscan_overview}). After merging the databases and applying inclusion and exclusion criteria (see Supplementary Table~\ref{tab:inclusion_criteria}), followed by the data processing pipeline, a total of 120,064,940 unique enrollees were retained, of whom 54\% were female, with an average of 517 tokens and a mean longitudinal duration of 45 months per enrollee. An example of a processed synthetic enrollee is shown in Supplementary Table~\ref{tab:plain_token_sequence}, with detailed processing steps provided in Supplementary Materials A.1.

\subsubsection{Training data}
\paragraph{Pre-training cohort}
The pre-training corpus comprised 118,064,940 enrollee records, constructed after holding out two non-overlapping evaluation cohorts of 1 million individuals each, with events restricted to those recorded on or before December 31, 2022 (Table~\ref{tab:data_summary}).

\paragraph{Post-training cohort}
{Post-training data were drawn exclusively from the pre-training corpus under the same temporal restriction. For each task, prompt--response pairs were built from eligible enrollee sequences by inserting an instruction token at an eligible position in the trajectory and deriving the response tokens from the record that follows, and 100,000 pairs were randomly sampled per task. The task-specific construction of prompts and responses is described with each task in Sections~\ref{sec:disease_task}--\ref{sec:rwe_task}.}

\begin{table}[!t]
\centering
\caption{{Summary of datasets used for model development and evaluation.} All cohorts include individuals retained after applying predefined inclusion and exclusion criteria and the data processing pipeline. {The held-out cohort is disjoint from the training corpus and serves all downstream evaluations. The predictive benchmarks use one half of its 2{,}000{,}000 enrollees and the RWE studies all of them, with the known-effect validation also drawing on all MarketScan MDCR enrollees, including pre-training enrollees (Methods).}}
\label{tab:data_summary}
\begin{tabular}{c c c c c}
\toprule
Data source & Role & Dataset & Time range & N \\
\midrule

\multirow{2}{*}{\begin{tabular}{c}MarketScan \\ Claims\end{tabular}}
& Training
& Training corpus        & 2008 -- 2022 & 118{,}064{,}940 \\

\cmidrule(lr){2-5}

& Evaluation
& {Held-out cohort} & {2008 -- 2024} & {2{,}000{,}000} \\
\midrule

\multirow{2}{*}{\begin{tabular}{c}Cross-domain \\ EHRs\end{tabular}}
& \multirow{2}{*}{Evaluation}
& EHRShot & 1990 -- 2023 & 5{,}845 \\
& 
& YNHH    & 2013 -- 2025 & {1{,}902{,}273} \\

\bottomrule
\end{tabular}

\end{table}

\subsubsection{Evaluation data}

\paragraph{{In-domain held-out cohorts}}
{Two non-overlapping cohorts of 1 million enrollees each were held out of the pre-training corpus and reserved for downstream evaluation. The RWE studies use both cohorts combined, 2 million enrollees, because their exposures are rare enough to need the larger population (Section~\ref{sec:rwe_task}). The predictive benchmarks, disease prediction and healthcare expenditure forecasting, use one cohort of 1 million enrollees, evaluated in two windows separated by December 31, 2022: (1) a retrospective evaluation restricted to data observed on or before that date, and (2) a prospective evaluation of outcomes occurring from January 1, 2023 onward, with model inputs restricted to each enrollee's history before 2023. Enrollees without adequate pre-2023 history or post-2022 follow-up are excluded from the prospective evaluation, and each task is further restricted to enrollees meeting its own observation and follow-up criteria.}

\paragraph{Cross-domain EHR cohort}
{We incorporate two cross-domain EHR-based cohorts (see Table~\ref{tab:data_summary}).} Yale New Haven Health (YNHH) is a private hospital-based EHR cohort from a five-hospital academic health system spanning Connecticut and Rhode Island. EHRShot is a public benchmark dataset derived from de-identified structured EHR data from Stanford Medicine~\cite{wornow2023ehrshot}. Both datasets are standardized in the Observational Medical Outcomes Partnership Common Data Model (OMOP CDM) and undergo analogous inclusion and exclusion criteria with harmonized preprocessing. Clinical variables are mapped using OHDSI standard vocabularies, and records are transformed into the same longitudinal trajectory schema used for MarketScan, yielding temporally ordered sequences for model inference. For claims-specific variables that are unavailable or differently encoded in EHRs, values are imputed when feasible; expenditure variables, which are not present in EHR data, are assigned the value \texttt{MISSING}. {YNHH serves both the RWE task and disease prediction, and EHRShot serves disease prediction only, each using all eligible patients. Neither cross-domain source supports expenditure forecasting, as EHR data lack financial variables.}

\subsection{Model Pre-training}\label{sec:pre-training}
ReClaim uses decoder-only Transformer architectures following the Qwen3 dense architecture~\cite{qwen3technicalreport}, but all model weights are learned \emph{de novo} rather than initialized from any released Qwen3 checkpoint. All variants operate on the same custom tokenizer and vocabulary. Given an input token sequence, the model computes contextual representations through stacked Transformer blocks with causal self-attention and predicts the next token with a linear output layer tied to the input embedding matrix.

We trained three model scales within this shared architectural family: \textbf{140M}, \textbf{700M} and \textbf{1.7B} parameters (see Supplementary Table~\ref{tab:qwen3_arch_configs} for details). The three variants differ primarily in depth and width while retaining the same overall architecture and vocabulary, so observed performance differences predominantly reflect model capacity rather than changes in tokenization or model design. 

We first pre-train each model with a standard \textbf{next-token prediction} objective on tokenized patient trajectories under our custom vocabulary.
Given a tokenized sequence $x_{1:T}$, we {minimize the cross-entropy loss}
\begingroup
\begin{equation}
\mathcal{L}_{\mathrm{CE}} = -\sum_{t=1}^{T}\log p_{\theta}(x_t \mid x_{<t}),
\end{equation}
\endgroup
{computed} between the predicted distribution and the ground-truth next token with a causal attention mask. In addition to cross-entropy loss $\mathcal{L}_{\mathrm{CE}}$, we apply a $z$-loss regularization term that penalizes the squared log-partition value computed from the logits. 
Let $\mathbf{s}_t \in \mathbb{R}^{|\mathcal{V}|}$ denote the logits at position $t$ over the vocabulary $\mathcal{V}$, and define
\begin{equation}
Z_t = \sum_{v \in \mathcal{V}} \exp(s_{t,v}).
\end{equation}
We then add
\begin{equation}
\mathcal{L}_{z} = \lambda \cdot \mathbb{E}\!\left[(\log Z_t)^2\right], \quad \lambda = 10^{-4},
\label{eq:zloss}
\end{equation}
and the total pre-training loss is
\begin{equation}
\mathcal{L} = \mathcal{L}_{\mathrm{CE}} + \mathcal{L}_{z}.
\end{equation}
This regularizer discourages excessive logit magnitude and global logit drift, which are only weakly constrained by cross-entropy alone. In practice, it improved numerical stability under large-scale training, reduced loss spikes and yielded more stable logit scales during optimization.

\subsection{{Disease Prediction}}\label{sec:disease_task}
\subsubsection{Task formulation}
{We assess disease prediction following the protocol of Delphi~\cite{shmatko2025learning}. Throughout, disease prediction denotes prediction of the first recorded occurrence of a target three-character ICD-10-CM category in the available longitudinal record. This observable, record-derived endpoint should not be interpreted as the biological onset of disease and may reflect delayed recognition or documentation, transfer of care, and differences in coding or healthcare utilization. The original Delphi study defined its endpoints with three-character ICD-10 codes in UK Biobank; because MarketScan and both cross-domain EHR datasets record diagnoses in ICD-10-CM, all endpoints here, including those used to train and evaluate the Delphi baseline, are three-character ICD-10-CM categories.} Each patient is represented as a tokenized longitudinal trajectory containing demographics and chronologically ordered diagnosis, procedure, medication, expenditure, and visit-attribute tokens (Supplementary Table~\ref{tab:plain_token_sequence}). Events are separated by inter-event time tokens (\texttt{<ATT-d>}) and year markers (\texttt{<NY>}); the age in days associated with each token is reconstructed by accumulating the gap tokens forward from the patient's birth-year token.
For each major ICD-10-CM endpoint \(k\) in the model vocabulary (e.g., \texttt{<DX-MAJOR\_E11>} for type 2 diabetes, \texttt{<DX-MAJOR\_I21>} for acute myocardial infarction, and \texttt{<DX-MAJOR\_C50>} for breast cancer), we construct a per-disease case--control cohort stratified by age and sex. {For the retrospective MarketScan and cross-domain EHR evaluations,} prediction times are induced directly from the trajectory rather than selected by an external rule. Specifically, for every event position \(j\) in a patient's history, we precompute the latest prior position \(\pi(j)\) whose age is at least \(\Delta\) days earlier:
\[
\pi(j) = \max\{i < j : \mathrm{age}_i \leq \mathrm{age}_j - \Delta\}.
\]
For a case, the outcome event is defined as the patient's first occurrence of token \(k\). {Patients with an earlier record of that exact token are excluded by construction, and only first-recorded target events are counted.} {Prior occurrence is assessed over all available recorded history before the prediction point, so lookback duration varies by patient. We do not impose a separate fixed disease-free washout interval; the prediction offset $\Delta$ specifies the horizon rather than the lookback duration. In the one-year retrospective and cross-domain evaluations, however, the prediction point must be a recorded encounter at least 365 days before the first occurrence, so every case has an observed record spanning at least one year before its first target-category diagnosis, with no earlier recorded occurrence of that category.} The prediction time is set to
\[
t_{\mathrm{pred}} = \mathrm{age}_{\pi(j)},
\]
where \(j\) denotes the first-occurrence index of token \(k\). For a control, the patient must have no occurrence of token \(k\) anywhere in the full trajectory. {The control prediction time is induced by a non-\(k\) token whose \(t_{\mathrm{pred}}\) falls within the same age group as the cases; comparisons remain stratified by sex and age.}
Specifically, we define age strata from 20 to 100 years in 10-year intervals.

{The prospective MarketScan evaluation instead uses December 31, 2022 as a fixed calendar cutoff. Model inputs contain only events recorded before 2023, and the prediction position is the last retained event before that cutoff. For each target category, patients with a recorded occurrence before the cutoff are excluded; cases have their first recorded occurrence during observed follow-up from January 1, 2023 onward, whereas controls have no recorded occurrence during their observed follow-up. Eligibility requires pre-cutoff history and post-cutoff follow-up, and outcome observation is censored at the end of each patient's available follow-up. Case--control comparisons remain stratified by sex and age at the prediction point. All models are scored on this shared prospective cohort using their retrospectively trained parameters, without fitting or updating them on prospective labels. Thus, unlike the patient-specific offset evaluation above, this analysis tests transfer to a calendar-anchored, incident-recorded-event task.}

{Within each dataset and prediction horizon, disease endpoints are aligned across models by their three-character ICD-10-CM category. Delphi's disease indices are decoded using its disease map, ReClaim's diagnosis-token indices using its vocabulary, and LightGBM's classifier targets using their diagnosis labels. These mappings resolve model-specific identifiers to the same disease category; for example, E11 is matched to E11 across all three models. For each category, all models use the same endpoint-specific case--control cohort, outcome definition, prediction times and sex-by-age strata. Cross-model comparisons retain only categories with non-missing ROC-AUC in all three mapped result sets. This yields 1,208 shared endpoints for retrospective MarketScan, 1,131 for prospective MarketScan, 1,024 for EHRShot and 1,204 for YNHH (Supplementary Table~\ref{tab:endpoint_harmonization}). A sex-by-age stratum is excluded if it contains no cases, no controls or fewer than two patients; an endpoint is evaluable if at least one stratum remains. Case--control cohort construction is performed once per category before inference, and the shared cohort file stores case and control identifiers and each patient's exact trajectory position at which prediction is made. The stored trajectory position is used directly by ReClaim and LightGBM and translated to the corresponding retained-event position for Delphi, with patient-level verification of the reconstructed Delphi sequence. After inference, model-specific indices are resolved back to canonical ICD-10-CM strings and the case count, control count and number of evaluable strata are compared endpoint by endpoint with the shared cohort file. This audit showed complete agreement for all endpoints evaluated in the retrospective MarketScan, prospective MarketScan, EHRShot and YNHH cohorts (Supplementary Table~\ref{tab:endpoint_harmonization}).}

{Rare endpoints are defined as those with more than 10 and fewer than 500 recorded cases in the 1-million held-out cohort (fewer than 5 per 10,000), counting cases from the retrospective and prospective evaluations combined. Applied to the 1,131 endpoints evaluated in both settings, this yields 437 rare endpoints, used for the rare-disease comparisons in Fig.~\ref{fig:auc_comp} and Supplementary Table~\ref{tab:posttraining_stats}; the 433 of them with more than 10 retrospective cases appear in Fig.~\ref{fig:auc_distribution}a.}

At inference, the model receives only tokens up to and including \(t_{\mathrm{pred}}\){ (or the stored pre-cutoff position in the prospective cohort). In the pre-trained versus post-trained ReClaim comparison, the patient cohort, prediction position, truncated history, target-token index and scoring procedure are identical; only the appended inference prompt differs. The pre-trained checkpoint receives ``\texttt{<NY> <ATT-0> <VT-outpatient> <DX-PRINCIPAL>}'', placing the next-token prediction in an outpatient principal-diagnosis context. The post-trained checkpoint receives ``\texttt{<NY> <INSTRUCT-DX>}'', matching its disease prediction instruction. In both cases, a single forward pass supplies} the final-position logit corresponding to vocabulary index \(k\) {as} the risk score. {These disease-specific outputs are read directly from the next-token prediction head, which produces logits $z_j$ over the full vocabulary $\mathcal{V}$.} We report ROC-AUCs with DeLong variance estimates within each stratum and average performance across strata. {For the patient-specific offset evaluations,} risk is evaluated at four clinically meaningful horizons by varying \(\Delta \in \{1,6,12,60\}\) months, corresponding to \(\Delta \in \{30, 182, 365.25, 1826.25\}\) days. The 1-month setting captures near-term risk relevant to immediate intervention.

A caveat is that case--control assignment is based on exact endpoint tokens. Thus, a patient who is negative for the queried token but positive for a clinically related token is still treated as a control. For example, when evaluating \texttt{<DX-MAJOR\_R73>} for prediabetes or abnormal glucose, patients carrying \texttt{<DX-MAJOR\_E11>} for type 2 diabetes are not excluded. The resulting ROC-AUC therefore reflects the model's ability to distinguish coded prediabetes from all other records, rather than from strictly disease-free controls. The same consideration applies to hierarchical or clinically related pairs such as I21 acute myocardial infarction versus I25 chronic ischemic heart disease, or C50 breast cancer versus Z80.3 family history of breast cancer and D05 carcinoma in situ of breast. For endpoints where such hierarchy is material, ROC-AUCs should be interpreted as discrimination of diagnoses as recorded in claims rather than as a pure disease-incidence model. We follow the Delphi convention of evaluating each endpoint identically { and examine the sensitivity of this conclusion to a broader, clinically curated phenotype definition in the Phecode analysis below.}

\paragraph{{Prior-code audit and ICD related-code exclusion}}
{We audited target-category occurrence in the recorded input history before each stored prediction position for all 1,208 retrospective endpoints and found none in any of the 5,800,059 case evaluations (Supplementary Table~\ref{tab:endpoint_harmonization}). For each target in E11 (type 2 diabetes), O24 (diabetes in pregnancy) and R73 (elevated glucose/prediabetes), we excluded patients carrying either of the other two categories anywhere in their observed record, first from controls alone and then from both cases and controls. We evaluated ReClaim-L, LightGBM and Delphi on the same retained cohorts at unchanged prediction positions, without retraining, and averaged ROC-AUC across evaluable sex-by-age strata. ReClaim-L cached logits were rescored, LightGBM used the published models and cached inference features, and Delphi was rerun to obtain per-patient scores. R73 was evaluated only with ReClaim-L because neither baseline supports this endpoint. Endpoint-specific demographic strata and counts are reported in Supplementary Table~\ref{tab:icd_related_exclusion}.}

\paragraph{{Phecode sensitivity analysis}}
{As a complementary robustness analysis, we evaluated 12 clinically curated Phecode phenotypes in the retrospective MarketScan test cohort (Supplementary Tables~\ref{tab:phecode_icd_mapping} and~\ref{tab:phecode_sensitivity}). We mapped the constituent ICD-10-CM codes to each phenotype and treated a patient with any mapped code as phenotype-positive; controls had none of that phenotype's mapped codes during the observed record. Thus, a patient was not classified as a control merely because one constituent ICD category was absent. Both model families used the same phenotype-specific evaluation cases, all eligible controls, and 16 sex-by-age strata. For this sensitivity analysis, ReClaim-L FT used a single jointly fine-tuned 12-label multi-label prediction head, with its label logits scored directly; this is distinct from the ICD-token next-token logit used in the primary benchmark. The bag-of-words LightGBM comparator comprised one independently fitted binary classifier per phenotype. We computed ROC-AUC within each stratum and report the unweighted mean across strata for each phenotype. To assess overlap between the type 2 diabetes (EM\_202.2) and prediabetes (EM\_204.5) endpoints, we repeated their evaluations on mutually exclusive cohorts. For each target, we excluded from both cases and controls every patient with any ICD-10-CM code mapped to the other phenotype anywhere in the observed trajectory. We retained the original prediction positions and rescored both existing models without retraining; all 16 strata remained evaluable for each endpoint. These two sensitivity results were not included in the mean across the original 12 phenotypes. These Phecodes group selected related codes but do not eliminate every possible overlap between clinically related phenotypes or establish biological disease onset.}

\subsubsection{{Post-training for disease prediction}}
Because pre-training optimizes general sequence modeling rather than endpoint-specific risk prediction, we performed a task-specific post-training stage for \textbf{{disease prediction}}. {For each model size, a single post-trained model is used across all target ICD categories.} To construct the post-training data, we transformed each patient trajectory into a prompt--response pair derived from the original longitudinal record.
For each randomly selected sequence with at least two months of records, we randomly selected a position of preceding history and inserted an instruction token at that point. The sequence preceding the instruction token was used as the prompt. {In the subsequent segment, we retained only target diagnosis categories first recorded after the instruction position and absent from the preceding available history; their tokens formed the response. This uses the same first-recorded-category definition as the retrospective, prospective and cross-domain evaluations, with the instruction position as the time anchor. This construction trains the model to map historical longitudinal context to subsequently recorded disease outputs while suppressing unrelated future tokens.}
We post-train the pre-trained model on 100,000 such pairs using causal next-token prediction on the concatenated prompt--response sequence, while masking the loss on prompt tokens so that optimization is confined to the response region.
Given an instruction prompt $p$ and a reference response {$y_{1:M}$, we minimize}
\begingroup
\begin{equation}
\mathcal{L}_{\mathrm{CE}} = -\sum_{m=1}^{M} \log p_{\theta}(y_m \mid p, y_{<m}).
\label{eq:posttrain_ce}
\end{equation}
\endgroup
{The $z$-loss $\mathcal{L}_{z}$ of Eq.~(\ref{eq:zloss}) is retained at the same weight, computed over the response positions, so the post-training objective for disease prediction is}
\begingroup
\begin{equation}
\mathcal{L}_{\mathrm{post}}^{\mathrm{dx}} = \mathcal{L}_{\mathrm{CE}} + \mathcal{L}_{z}.
\label{eq:posttrain_dx_total}
\end{equation}
\endgroup
{This stage aligns the pre-trained model with disease prediction by restricting supervision to the response region, so that the model learns to predict target diagnoses first recorded after the instruction position from the preceding longitudinal context while keeping the tokenizer and vocabulary unchanged from pre-training.}

{To assess the contributions of input representation and post-training, we compared ReClaim-S using full claims or diagnosis-only trajectories before and after post-training (Supplementary Table~\ref{tab:input_posttrain}).}

\subsubsection{{Baselines for disease prediction}}
\paragraph{Disease-specific LightGBM}
We use LightGBM~\cite{ke2017lightgbm} as a tabular baseline, trained on word-count features extracted from each patient's historical record over the observation window, together with demographic variables. This bag-of-words representation preserves event occurrence and frequency information but discards event order and temporality. Model development is performed on a randomly sampled subset of 5 million individuals from the training corpus, consistent with Curiosity~\cite{waxler2025generative}.

{We fit a separate binary classifier for each endpoint. We adopt the core boosting settings of the Curiosity baseline~\cite{waxler2025generative}: a maximum of 10,000 boosting rounds, maximum tree depth of 6, learning rate of 0.1, and early-stopping patience of 100 rounds. The classifier uses a binary objective, ROC-AUC as the evaluation metric, and a feature-sampling fraction of 0.8 per tree. Early stopping monitors ROC-AUC on the validation data supplied to the fitting procedure and selects the best iteration. These settings are fixed across endpoints without an additional hyperparameter search; other parameters retain the LightGBM defaults. For inference-only evaluations, the saved classifiers are loaded without refitting.}

\paragraph{Pre-trained Delphi}
We followed the publicly released Delphi code base~\cite{shmatko2025learning} and used the default settings for training. Delphi is trained on the full available MarketScan corpus. We retain only the demographic tokens and disease-target events required by the public implementation when constructing Delphi inputs. {Delphi is then evaluated on the same retrospective held-out, prospective temporal, and cross-domain disease endpoint sets whenever compatible labels are available.}

\paragraph{{Pre-trained ETHOS}}
{ETHOS (Enhanced Transformer for Health Outcome Simulation) is a decoder-only Transformer trained on tokenized patient health timelines to autoregressively generate future clinical events and support zero-shot outcome prediction~\cite{renc2024zero}. We evaluated its frozen, publicly released checkpoint on EHRShot using 64 Monte Carlo simulations per prediction point, scoring each disease by the fraction of simulated trajectories containing its diagnosis token after the prediction point. Its vocabulary supports 1,010 of the 1,024 ICD-10-CM categories evaluated on EHRShot. The ETHOS comparison restricts ReClaim results to these same 1,010 categories and is reported separately from the full benchmark (Supplementary Table~\ref{tab:ethos_ehrshot}). ETHOS was evaluated only on EHRShot because vocabulary differences prevented reliable endpoint alignment on MarketScan and YNHH.}

\paragraph{{GPT-2}}
{We instantiated a GPT-2-style decoder-only Transformer with 60M parameters. We held the data and learning protocol fixed: the model was trained from scratch with ReClaim's custom tokenizer on the same MarketScan pre-training corpus, then post-trained on the identical 100,000 prompt--response pairs described above. We evaluated both pre-trained and post-trained checkpoints on retrospective MarketScan and the post-trained checkpoint on EHRShot, using the same cohorts and metrics as ReClaim (Supplementary Tables~\ref{tab:add_baselines} and~\ref{tab:ethos_ehrshot}).}

\paragraph{{LLaMA-3.1 }}
{We instantiated a LLaMA-3.1-style causal Transformer with 120M parameters, which replaces GPT-2's learned absolute positions and conventional feed-forward blocks with rotary positional embeddings, RMS normalization and SwiGLU feed-forward layers, and uses grouped-query attention. It was trained from scratch with the same custom vocabulary, tokenizer, pre-training corpus, causal objective and post-training set for disease prediction as the GPT-2 baseline, and evaluated with the same protocol as other baselines.}

\paragraph{{ReClaim inference comparison}}
{We compared three inference approaches using ReClaim-S on EHRShot: diagnosis-token logits from the pre-trained checkpoint, Monte Carlo simulation using that same checkpoint, and diagnosis-token logits from the post-trained checkpoint. Following the simulation strategy of Curiosity~\cite{waxler2025generative}, we sampled 64 future trajectories per prediction point and scored each endpoint by the fraction containing its diagnosis token after the prediction point. All three approaches used the same 1,024 EHRShot endpoints, endpoint-specific case--control cohorts and sex-by-age strata (Supplementary Fig.~\ref{fig:inference_comparison}).}

\subsection{Healthcare Expenditure Forecasting}\label{sec:expenditure_task}
\subsubsection{Task formulation}
We formulate next-year healthcare expenditure forecasting as estimating an enrollee’s total gross payments to all providers in the subsequent calendar year, following prior work~\cite{osawa2020cost}. For each enrollee, {a randomly chosen} eligible prediction point with at least one year of prior history and one year of follow-up partitions the longitudinal sequence into a preceding context segment and a subsequent outcome segment; the former serves as model input, while the latter is used to derive the target, defined as total healthcare expenditure over the following calendar year. {Targets are right-censored at \$1,000,000 (0.007\% of enrollee-years affected; Supplementary Table~\ref{tab:discretization}).} {Performance is evaluated for regression ($R^2$, MAE) and for classification (ROC-AUC, PR-AUC) of HNHC patients, defined by next-year spending of at least \$30,000, approximately the 95th percentile in the 2022 MEPS report~\cite{MEPS_StatisticalBrief560_2025} (Supplementary Table~\ref{tab:healthcare_expenditures_percentile}), following the convention that HNHC patients are the top 5\% of annual spenders, who account for nearly half of the country's annual health care spending~\cite{blumenthal2016caring,osawa2020cost}.} We report performance under both the retrospective and the prospective evaluation on the 1 million held-out cohort.

\subsubsection{{Post-training for healthcare expenditure forecasting}}\label{sec:posttrain}
{To optimize ReClaim for forecasting the annual expenditure target, we performed a post-training stage for healthcare expenditure forecasting analogous to the disease prediction stage.}
{Each enrollee in the supervision pool contributes one prompt--response pair (Supplementary Table~\ref{tab:posttrain_cost_token_sequence}; target distribution in Supplementary Table~\ref{tab:posttrain_cost_stats}). The prompt is the trajectory up to the prediction point, followed by the inserted instruction token \texttt{<INSTRUCT-COST>}. The response is a single token representing the enrollee's total expenditure over the following calendar year, constructed by summing the cost tokens of that year's sequence and re-encoding as a cost token, so that the response draws on the pre-training vocabulary and introduces no new token.}
{Each model is post-trained on these pairs, the loss again masked on prompt tokens (Supplementary Table~\ref{tab:training_details}).}
{Because the cost tokens encode one significant digit and an order of magnitude, they form a value-ordered ladder on which adjacent entries differ by a bounded ratio. We therefore replace the hard next-token target of Eq.~(\ref{eq:posttrain_ce}) by a soft ordinal target, a discretized Gaussian centered on the true cost token with a standard deviation of one position on the ladder, and optimize}
\begingroup
\begin{equation}
\mathcal{L}_{\mathrm{ord}} = -\sum_{j} q(j)\,\log p_{\theta}(c_j \mid p), \qquad q(j) \propto \exp\!\left(-\frac{(j-j^{\ast})^{2}}{2\sigma^{2}}\right),\ \ \sigma = 1,
\label{eq:ordinal}
\end{equation}
\endgroup
{where $c_j$ is the $j$-th cost token on the ladder and $j^{\ast}$ the position of the true token, so that near-misses are penalized less than distant errors.}
{A Huber penalty $\mathcal{L}_{\mathrm{Huber}}$ (weight 0.5) on the distribution's log-dollar expectation ties the expenditure forecast to the continuous target and keeps extreme values from dominating the loss. The closing \texttt{<eos>} position keeps the ordinary cross-entropy $\mathcal{L}_{\mathrm{eos}}$ of Eq.~(\ref{eq:posttrain_ce}), and the z-loss $\mathcal{L}_{z}$ of Eq.~(\ref{eq:zloss}) is retained at the same weight, so the post-training objective for expenditure forecasting is}
\begingroup
\begin{equation}
\mathcal{L}_{\mathrm{post}}^{\mathrm{cost}} = \mathcal{L}_{\mathrm{ord}} + \mathcal{L}_{\mathrm{Huber}} + \mathcal{L}_{\mathrm{eos}} + \mathcal{L}_{z}.
\label{eq:posttrain_total}
\end{equation}
\endgroup
{At inference the cost token is decoded greedily, and the softmax over the cost tokens is a distribution over next-year expenditure: its mean gives the expenditure forecast, and its mass at or above a threshold such as \$30,000 gives the HNHC probability. A single decode therefore supplies both the regression and the classification predictions.}
\subsubsection{{Baselines for healthcare expenditure forecasting}}

\paragraph{{Expenditure-specific LightGBM}}
{The tabular baseline follows the same recipe as the disease prediction LightGBM, with two changes drawn from prior work on expenditure~\cite{osawa2020cost}: prior-year total expenditure is added to the features, and the bag-of-words clinical-code features are built over two windows, the last observed year and the enrollee's entire available history. We fit a regressor for the continuous total expenditure and a classifier for the HNHC label (Supplementary Tables~\ref{tab:expenditure_full_results} and \ref{tab:expenditure_by_payer}).}

\paragraph{{Pre-trained ReClaim, Monte Carlo simulation}}
{We evaluated pre-trained ReClaim models by Monte Carlo simulation}. For each prediction instance, the model is prompted with the input sequence corresponding to the historical context segment. Using the default sampling configuration (temperature = 1, top-$p$ = 1), the model probabilistically generates {Monte Carlo simulations of the future trajectory, with the simulation count varied from 1 to 64}. From each simulation, cost tokens corresponding to the subsequent year are extracted and aggregated to estimate annual healthcare expenditures{, and the fraction of simulations at or above a threshold gives the corresponding probability. Both ReClaim inference methods (pre-training and post-training, Section~\ref{sec:posttrain}) run on the same inference engine (vLLM~\cite{kwon2023efficient}) with identical settings, so their inference costs are directly comparable (Supplementary Table~\ref{tab:inference_cost}).}

\subsection{Support for Real-{W}orld Evidence Studies}\label{sec:rwe_task}
\subsubsection{{Task formulation}}
{Observational RWE studies are threatened by confounding, which propensity-score methods address by balancing the covariates they are given. A study can therefore adjust only for what its covariate set captures, and that set is conventionally chosen by an investigator or derived from the data by an algorithm. We asked whether a foundation-model representation of an enrollee's healthcare trajectory can serve as that covariate set. We formulate the task as estimating a treatment effect from real-world data, in new-user cohorts comparing two treatments, varying the covariate set while carrying every specification through the same cohorts, estimators and diagnostics. We conducted two complementary studies: a known-effect validation, which judges each specification against an effect expected from randomized trials and an expected null association, and a replication of a published target trial emulation, which tests agreement with expert-curated covariates on two data sources.}

\subsubsection{{Propensity-score specifications}}
{Four kinds of covariate set are compared. (1)~Unadjusted, the raw comparison, with no covariates and no propensity-score model. (2)~Expert, the reference study's own domain-expert covariate list, where the study supplies one. (3)~hdPS, demographics plus 500 covariates derived by the high-dimensional propensity-score (hdPS) algorithm~\cite{schneeweiss2009high} with its original settings. (4)~A foundation-model representation, with demographics added separately because left truncation of long histories can cut off the demographic tokens at the start of the sequence. The tokenized pre-index history is encoded by the frozen pre-trained model, with no task-specific training, and the embedding is the final hidden state at the last token, L2-normalized. We use ReClaim-S, -M and -L, and include Delphi as a baseline in the known-effect validation. Propensity models built on foundation-model representations are ridge-penalized logistic regressions, with the penalty chosen by five-fold cross-validation given their dimension; the other specifications use unpenalized logistic regression, as in their original studies.}

\subsubsection{{Known-effect validation}}\label{sec:rwe_study1}
{Following the seminal high-dimensional propensity-score study~\cite{schneeweiss2009high}, we replicated the two of its three example studies that are feasible in MarketScan data, as a positive and a negative control, rule by rule in all eligible MDCR enrollees aged 65 years or older, as in the reference's Medicare-linked PACE cohort (Supplementary Table~\ref{tab:baseline_controls}). The index window is 2011--2014, the earliest period for which MarketScan MDCR supplies the reference's full baseline lookback. The positive control, cyclooxygenase-2 (COX-2) inhibitor versus nonselective NSAID initiation and the risk of severe gastrointestinal complications, has a moderate protective effect expected from randomized trials~\cite{moore2005tolerability}, which confounding may conceal; the negative control, influenza vaccination versus none and the risk of hip fracture, has an expected null association with strong contextual support. A specification is judged by whether its adjusted estimate, a propensity-score-decile-adjusted relative risk~\cite{schneeweiss2009high}, moves below the null in the positive control and stays at 1 in the negative control. The known-effect validation is not run on the YNHH EHR, where records begin in 2013 and the reference's baseline requirement places the earliest index year in 2015, by which time celecoxib had become generic and the exposure contrast was no longer comparable to the reference's.}

{Two further analyses probe the representations. First, each ReClaim representation is compressed by unsupervised principal component analysis to 120 and 500 dimensions, matching Delphi's dimension and the hdPS covariate count, and the specifications refitted on the compressed representations (Supplementary Table~\ref{tab:t10_pca_compressions}). Second, a doubly robust analysis tests the representations as outcome models, crossing the hdPS, Delphi and ReClaim-S propensity models with outcome models of the same ridge-penalized form built on the Delphi or ReClaim-S, -M or -L representation. Because treatment is uncommon in both cohorts, we estimated the effect in the treated by cross-fitted augmented inverse probability weighting (AIPW; Supplementary Table~\ref{tab:doubly_robust} and Supplementary Fig.~\ref{fig:doubly_robust}).}

\subsubsection{{Target trial emulation}}\label{sec:rwe_study2}
{We replicated a recent target trial emulation~\cite{tang2025association} of GLP-1RA versus SGLT2i and versus DPP4i for new-onset psychiatric outcomes on two sources, the MarketScan held-out cohort and the YNHH EHR, following the reference's cohort construction and eligibility criteria (Supplementary Table~\ref{tab:rwe_application_definition}). The claims and EHR sources are harmonized to shared variables and analyzed through the same pipeline, and the resulting cohorts match the reference's arm by arm (Supplementary Table~\ref{tab:emulation_cohorts}). The Expert specification is the reference's own covariate list, built per source as documented in Supplementary Table~\ref{tab:rwe_application_reference_spec}. We report anxiety and depression, the two outcomes with adequate events on both sources, as the main results; the rest are event-limited and are reported with them in Supplementary Table~\ref{tab:t4_outcomes_both_sources}. The primary estimator is the reference's own: stabilized inverse-probability-of-treatment weighting with a weighted Cox model. Following the reference study's sensitivity analysis, the same propensity score is reused with 1:1 nearest-neighbor matching at its caliper of 0.2 standard deviations. Every fit records standardized mean differences on each specification's own covariates and on the reference study's expert-curated covariates, empirical equipoise~\cite{walker2013tool}, and the effective sample size and maximum weight under weighting~\cite{austin2015moving}. Residual systematic bias is quantified with the reference's panel of negative control outcomes (NCOs), outcomes with no plausible causal relationship with the treatments under study but expected to share their sources of bias~\cite{lipsitch2010negative, conover2025objective} (Supplementary Table~\ref{tab:rwe_nco_icd10}); per cell, NCO estimates are used to fit an empirical null distribution, with residual bias summarized by the expected absolute systematic error (EASE)~\cite{schuemie2014interpreting}.}

\section{Data and code availability}
The Merative MarketScan Research Databases used for model development and internal evaluation are not publicly redistributable because of commercial licensing restrictions; access can be obtained from Merative under an appropriate data-use agreement. Yale New Haven Health electronic health record data are not publicly available because they contain protected health information and are subject to institutional, privacy and regulatory restrictions; access requires appropriate institutional approvals and data-use agreements. EHRShot is a public benchmark dataset derived from de-identified structured electronic health records from Stanford Medicine and is available through the original EHRSHOT Redivis release at \url{https://stanford.redivis.com/datasets/53gc-8rhx41kgt}, subject to the EHRSHOT research usage agreement. Documentation for EHRShot is available at \url{https://som-shahlab.github.io/ehrshot-website/docs/intro/dataset/}. 

Code for ReClaim, together with accompanying scripts for data processing, model training, evaluation and figure generation, is available at GitHub (\url{https://github.com/BIDS-Xu-Lab}).

\section{Competing Interests}
The authors declare no competing interests.

\section{Author Contributions}
F.M., Y.L., X.L., W.Z. and H.X. conceived the study.
F.M., Y.L., and X.L. developed the model and experimental framework.
F.M., Y.L., and J.N. did the experiments for {disease prediction}.
Y.L. performed the expenditure prediction and target trial emulation analysis. 
X.L. performed the model pre-training.
X.L., Y.L. performed the model post-training.
J.N. visualized the embeddings for ReClaim and Delphi. 
Y.L., X.L., F.M., and W.Z. performed data processing.
F.M., Y.L., and X.L. drafted the first version manuscript, and all authors contributed to subsequent revisions. 
{B.O. participated in revision discussions, revised the manuscript and checked the data.}
H.X. supervised the study.
All authors approved the final version for submission.

\bibliographystyle{unsrt}
\bibliography{references} %

\newpage
\begin{appendices}
\appendix
\appendix
\renewcommand{\thetable}{A\arabic{table}}
\renewcommand{\thefigure}{A\arabic{figure}}
\renewcommand{\thesubsection}{A.\arabic{subsection}}
\renewcommand{\thesubsubsection}{A.\arabic{subsection}.\arabic{subsubsection}}
\setcounter{table}{0}
\setcounter{figure}{0}

\clearpage
\FloatBarrier
\section{Supplementary Materials}

\subsection*{Supplementary Figures}
\begin{figure}[H]
    \centering
    \tikz[baseline=(img.south west)]{\node[anchor=south west,inner sep=0](img)%
        {\includegraphics[width=0.25\linewidth]{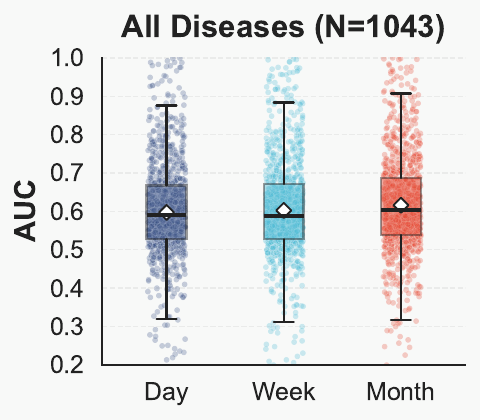}};%
        \node[anchor=north west,font=\large\bfseries] at (img.north west){a};}%
    \tikz[baseline=(img.south west)]{\node[anchor=south west,inner sep=0](img)%
        {\includegraphics[width=0.25\linewidth]{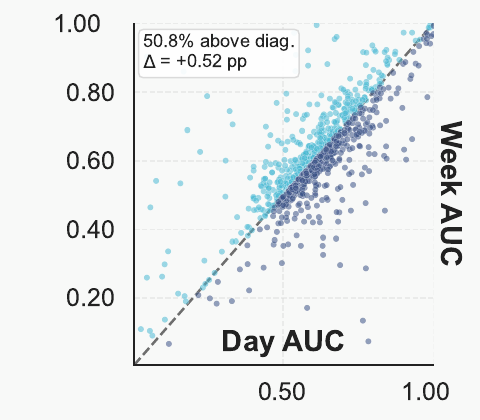}};%
        \node[anchor=north west,font=\large\bfseries] at (img.north west){b};}%
    \tikz[baseline=(img.south west)]{\node[anchor=south west,inner sep=0](img)%
        {\includegraphics[width=0.25\linewidth]{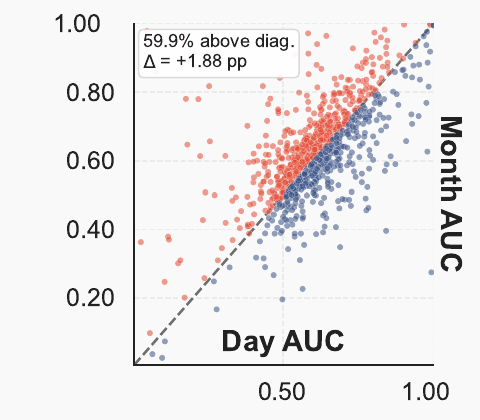}};%
        \node[anchor=north west,font=\large\bfseries] at (img.north west){c};}%
    \tikz[baseline=(img.south west)]{\node[anchor=south west,inner sep=0](img)%
        {\includegraphics[width=0.25\linewidth]{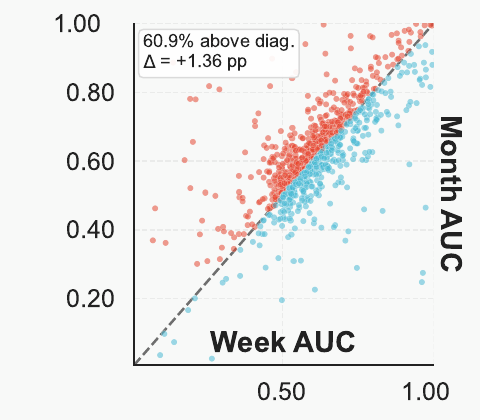}};%
        \node[anchor=north west,font=\large\bfseries] at (img.north west){d};}%
    \caption{\textbf{Coarser temporal-resolution encoding favors the Month variant.} Comparison of three temporal-encoding granularities (Day, Week, Month) used to discretize inter-event intervals in the patient-trajectory representation. All three variants are ReClaim-S models pre-trained on the full MarketScan training corpus and evaluated on a 100,000-patient held-out validation cohort. Disease tokens were harmonized across encodings via the canonical \texttt{<DX-MAJOR\_*>} identifier, yielding 1,043 disease endpoints shared by all three variants; all panels use this 3-way intersection. (\textbf{a}),~Distribution of disease-level ROC-AUC for the Day, Week, and Month encodings, shown as box plots with overlaid jittered scatter; diamonds indicate group means. Mean (median) ROC-AUC was 0.597 (0.590) for Day, 0.603 (0.588) for Week, and 0.616 (0.603) for Month, identifying Month as the strongest variant overall. (\textbf{b}),~Per-disease paired scatter of Week ROC-AUC versus Day ROC-AUC; the two encodings are largely comparable, with Week exceeding Day in 50.8\% of endpoints (mean\,\(\Delta\)\,=\,+0.52\,pp). (\textbf{c}),~Per-disease paired scatter of Month ROC-AUC versus Day ROC-AUC; the Month encoding exceeds the Day encoding in 59.9\% of endpoints (mean\,\(\Delta\)\,=\,+1.88\,pp). (\textbf{d}),~Per-disease paired scatter of Month ROC-AUC versus Week ROC-AUC; the Month encoding exceeds the Week encoding in 60.9\% of endpoints (mean\,\(\Delta\)\,=\,+1.36\,pp). Together, these comparisons identify Month as the preferred temporal-resolution choice: it delivers the highest mean and median ROC-AUC and outperforms both Day and Week in the majority of disease endpoints, while at the same time producing the shortest patient-level token sequence among the three variants and therefore the lowest computational and context-length cost during pre-training and inference.}
    \label{fig:encoding_comparison}
\end{figure}

\clearpage

\begin{figure}[H]
    \centering
    \tikz[baseline=(img.south west)]{\node[anchor=south west,inner sep=0](img)%
        {\includegraphics[width=0.317\linewidth]{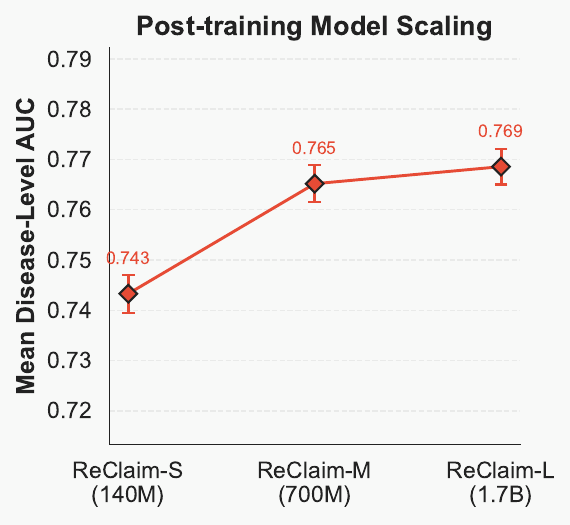}};%
        \node[anchor=north west,font=\large\bfseries] at (img.north west){a};}\hfill%
    \tikz[baseline=(img.south west)]{\node[anchor=south west,inner sep=0](img)%
        {\includegraphics[width=0.267\linewidth]{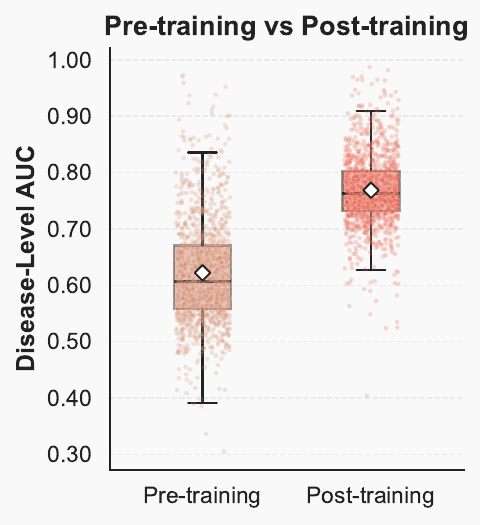}};%
        \node[anchor=north west,font=\large\bfseries] at (img.north west){b};}\hfill%
    \tikz[baseline=(img.south west)]{\node[anchor=south west,inner sep=0](img)%
        {\includegraphics[width=0.376\linewidth]{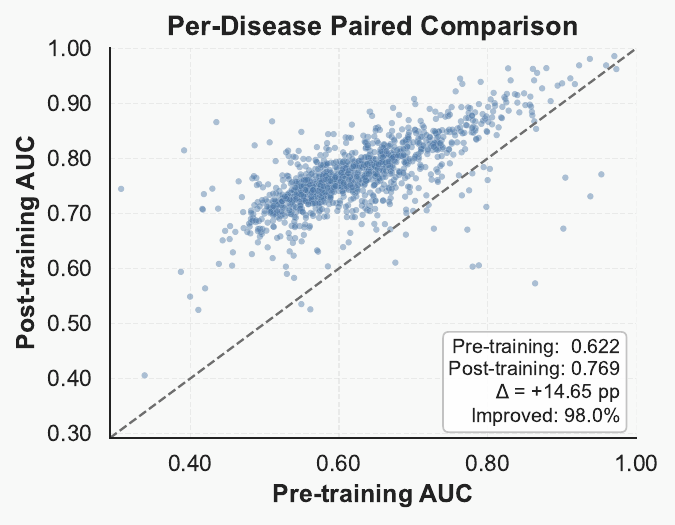}};%
        \node[anchor=north west,font=\large\bfseries] at (img.north west){c};}%
    \caption{{\textbf{Scaling and post-training improve disease prediction.} All panels cover the same 1,208 shared disease endpoints. (\textbf{a}),~Post-trained disease prediction improves monotonically with model scale: mean ROC-AUC\,=\,0.743 (ReClaim-S), 0.765 (ReClaim-M) and 0.769 (ReClaim-L). (\textbf{b}),~Disease-level ROC-AUC before and after post-training for the paired ReClaim-L checkpoints (box plots with overlaid jittered scatter; diamonds mark group means); mean ROC-AUC rises from 0.622 to 0.769. (\textbf{c}),~Paired per-disease scatter of pre- versus post-training ROC-AUC for ReClaim-L; points above the diagonal favor post-training (98.0\% of endpoints; mean $\Delta = +14.65$\,pp; 95\% CI, 14.29--15.01; \(P=2.0\times10^{-191}\), two-sided paired Wilcoxon signed-rank test).}}
    \label{fig:scaling_and_posttraining}
\end{figure}

\clearpage

\begin{figure}[H]
    \centering
    \includegraphics[width=\linewidth]{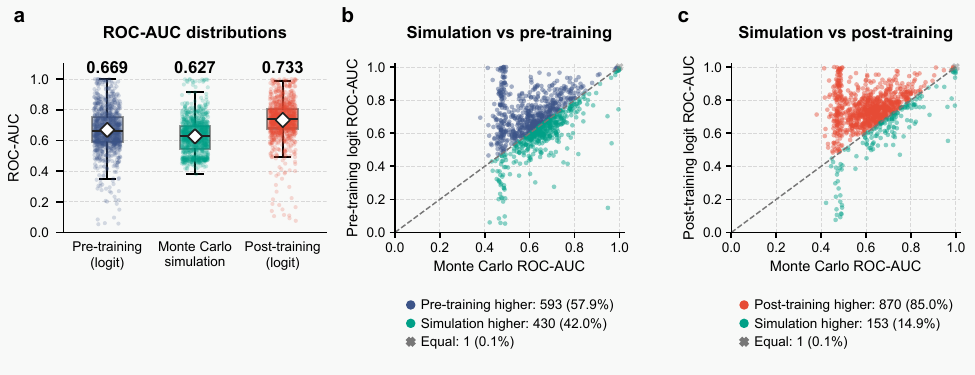}
    \caption{{\textbf{Pre-training, Monte Carlo simulation and post-training for disease prediction on EHRShot.} All three approaches use ReClaim-S and the same 1,024 disease endpoints, endpoint-specific case--control cohorts and sex-by-age strata. (\textbf{a}), From left to right: pre-trained diagnosis-token logits, Monte Carlo simulation with the pre-trained checkpoint, and post-trained diagnosis-token logits. Simulation scores each endpoint by the fraction of 64 sampled future trajectories containing its diagnosis token. Mean ROC-AUCs are 0.669, 0.627 and 0.733, respectively. Each point represents one endpoint; boxes show the median and interquartile range, whiskers extend to the most extreme values within 1.5 times the interquartile range, and diamonds mark unweighted means. (\textbf{b,c}), Paired disease-level comparisons of simulation (horizontal axis) with pre-trained logits (b, vertical axis) or post-trained logits (c, vertical axis). Colors and legend counts indicate which approach achieves higher ROC-AUC; the dashed line marks equal performance. On the same NVIDIA H200 system and prediction points, post-trained inference required 39.9 GPU-seconds versus 20,160 for simulation, approximately $505\times$ lower cumulative GPU time.}}
    \label{fig:inference_comparison}
\end{figure}

\clearpage

\begin{figure}[H]
    \centering
    \includegraphics[width=88mm]{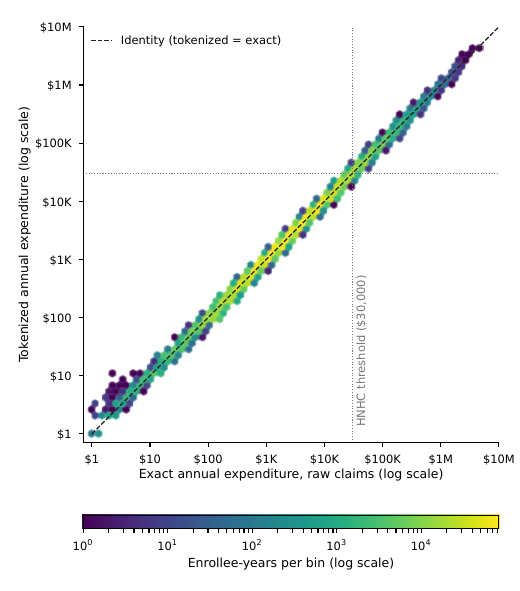}
    \caption{{Cost-token rounding preserves annual expenditure across six orders of magnitude. Two-dimensional histogram of the 2,084,595 enrollee-years contributed by the 359,187 enrollees of the retrospective and prospective evaluation cohorts (2008--2023). For each enrollee-year the horizontal axis is the exact annual expenditure reconstructed from the raw MarketScan inpatient, outpatient and pharmacy claims, and the vertical axis is the same year's total after tokenization: payments are summed per enrollee, month, payer and claim type, each monthly sum is rounded to one significant digit, and the tokenized annual total is the sum of those rounded monthly values. Both axes are logarithmic. Each hexagon is one bin of the plane, colored by the number of enrollee-years it contains on a logarithmic scale. The dashed red line is the identity, on which tokenization changed nothing; the dotted gray lines mark the high-need, high-cost (HNHC) threshold of \$30{,}000 on both axes. The cloud forms a band of near-constant width around the identity, because rounding to one significant digit moves a value by at most about one third (approximately $-0.18$ to $+0.13$ on the $\log_{10}$ axis); the slight periodic wobble is the discretization itself, as values just below a rounding midpoint round down, whereas values at or above it round up. Below about \$10 the encoding is coarse (\$1, \$2, \$3, \ldots), so tokenized totals snap to a few discrete levels. Enrollee-years with exact totals below \$1, and years containing a negative monthly sum (0.15\%; refund or adjustment claims, which the tokenizer maps to \$0), are not shown. R$^2$ of the tokenized against the exact total 0.993; the HNHC label is changed by tokenization in 0.42\% of enrollee-years (Supplementary Table~\ref{tab:discretization}).}}
    \label{fig:discretization}
\end{figure}

\clearpage

\begin{figure}[H]
    \centering
    \includegraphics[width=\linewidth]{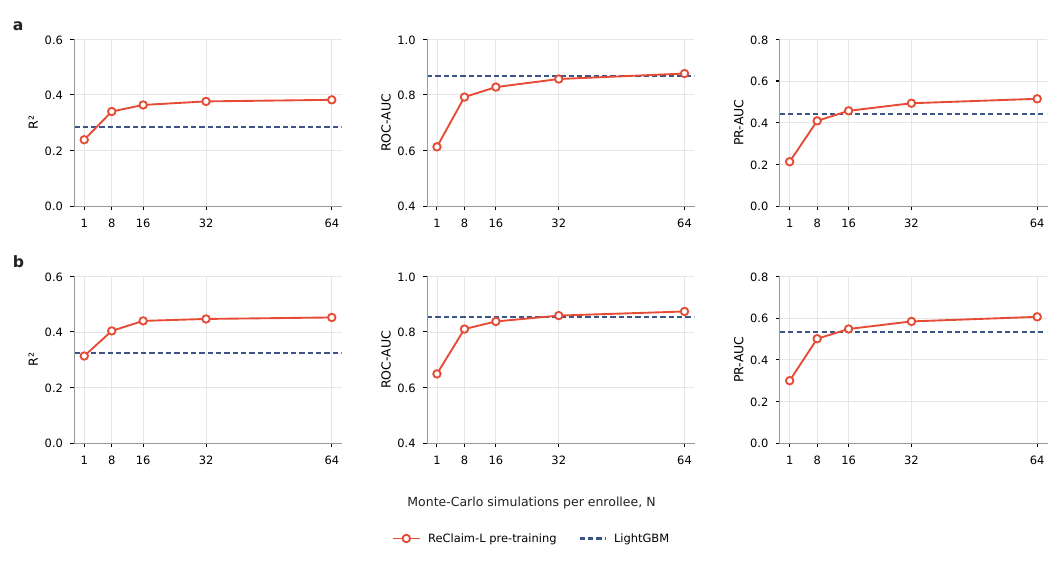}
    \caption{{\textbf{Expenditure forecasting performance by number of Monte Carlo simulations.} (\textbf{a}),~Retrospective and (\textbf{b}),~prospective evaluation in the held-out MarketScan cohort: $R^2$, HNHC ROC-AUC and PR-AUC of pre-trained ReClaim-L read out by $N$ = 1, 8, 16, 32 and 64 simulations of the following year per enrollee. Dashed line, LightGBM. $N$ = 1 is greedy decoding. GPU time grows linearly with $N$; values for all three backbone sizes are in Supplementary Table~\ref{tab:mc_trajectory_ablation}.}}
    \label{fig:mc_ablation}
\end{figure}

\clearpage

\begin{figure}[H]
    \centering
    \includegraphics[width=\linewidth]{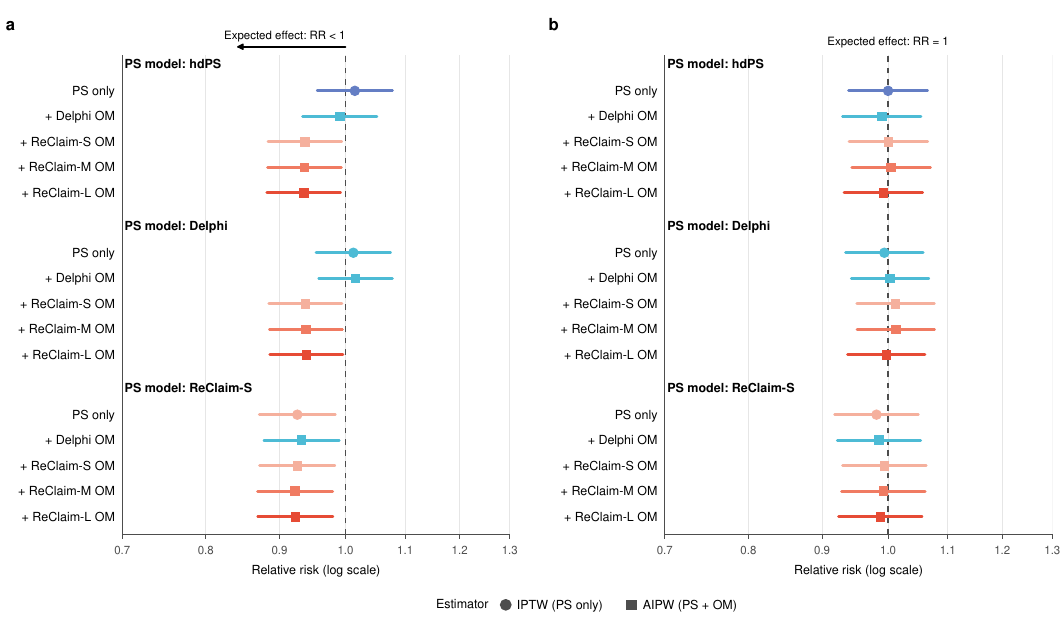}
    \caption{{\textbf{Doubly robust analysis of the known-effect validation.} (\textbf{a}),~Positive control, cyclooxygenase-2 (COX-2) inhibitors versus nonselective NSAIDs and the risk of gastrointestinal complications; the expected effect is protective (arrow). (\textbf{b}),~Negative control, influenza vaccination and the risk of hip fracture; the expected relative risk is 1 (dashed line). Each block is one propensity-score model (hdPS, Delphi or ReClaim-S), alone or paired with an outcome model (OM) built on the Delphi or ReClaim-S, -M or -L representation. Circles, propensity-score weighting for the effect in the treated without an outcome model (IPTW); squares, cross-fitted augmented inverse probability weighting (AIPW); bars are 95\% confidence intervals. Source data in Supplementary Table~\ref{tab:doubly_robust}. hdPS, high-dimensional propensity score; IPTW, inverse probability of treatment weighting; NSAID, nonsteroidal anti-inflammatory drug.}}
    \label{fig:doubly_robust}
\end{figure}

\clearpage

\begin{figure}[H]
    \centering
    
    \tikz[baseline=(img.south west)]{\node[anchor=south west,inner sep=0](img)%
        {\includegraphics[width=0.48\linewidth]{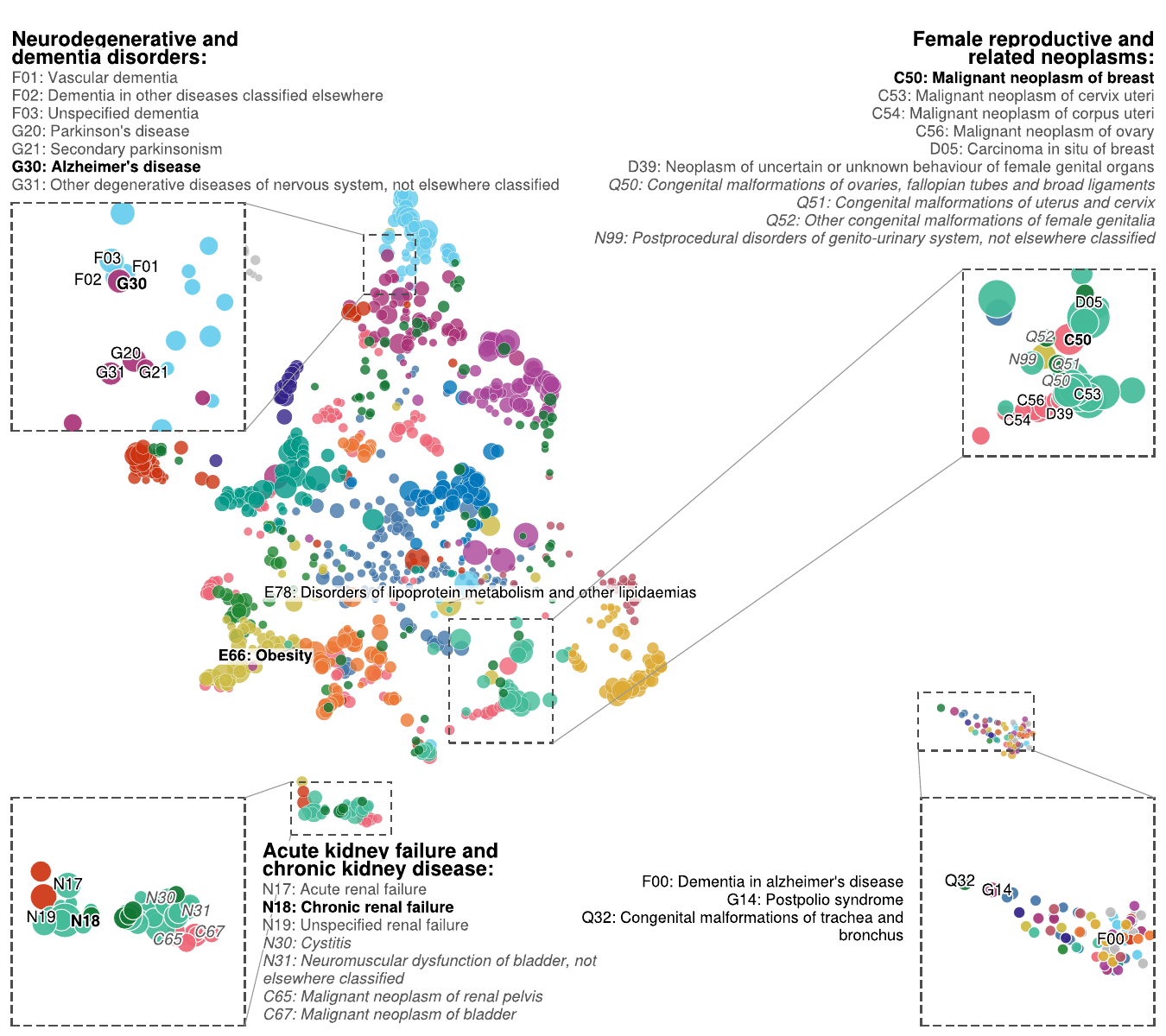}};%
        \node[anchor=south west,font=\normalsize\bfseries] at (img.north west){a. Delphi};}\hfill
    \tikz[baseline=(img.south west)]{\node[anchor=south west,inner sep=0](img)%
        {\includegraphics[width=0.48\linewidth]{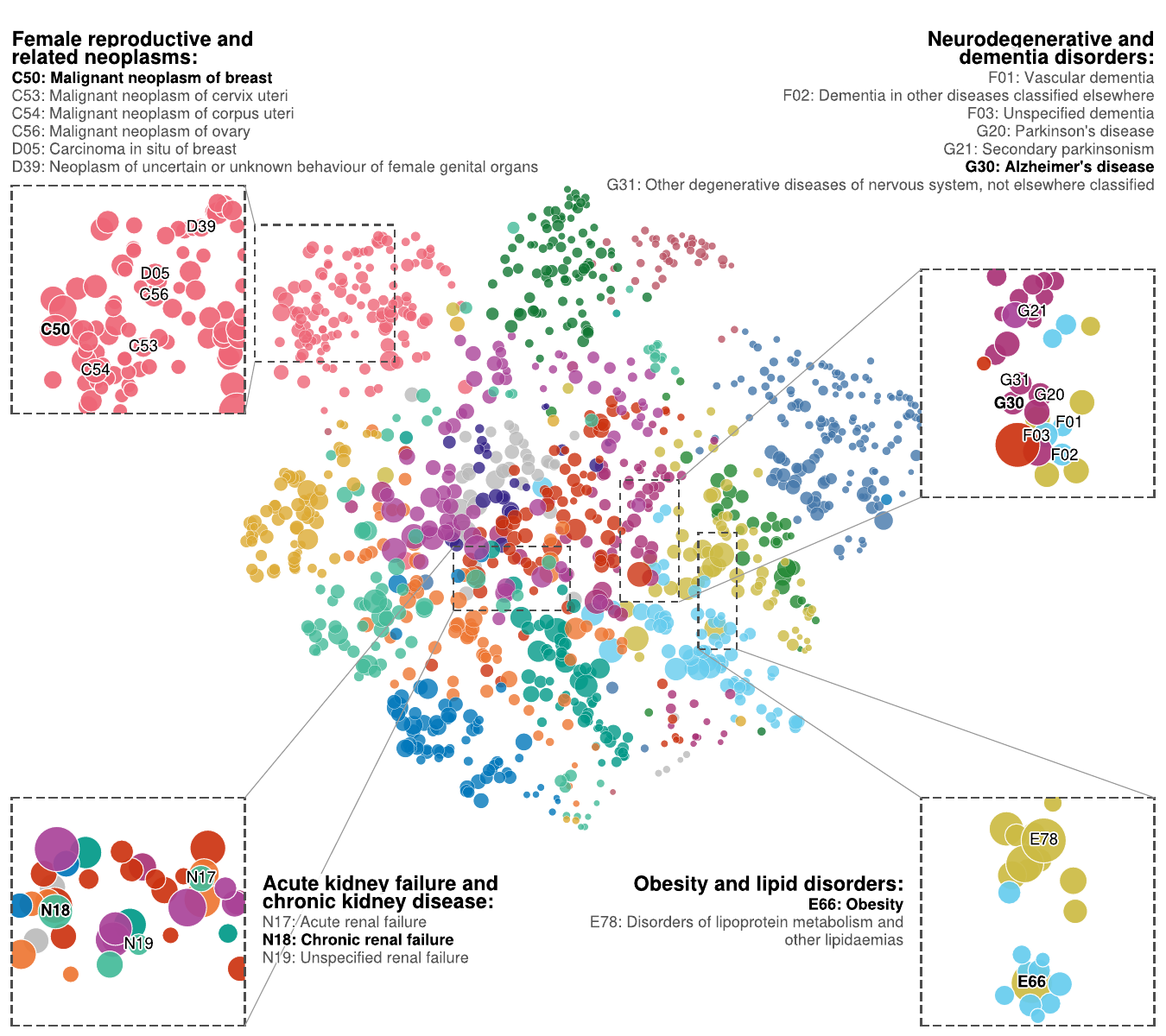}};%
        \node[anchor=south west,font=\normalsize\bfseries] at (img.north west){b. ReClaim};}
    \par\vspace{0.2em}
    \includegraphics[width=0.95\linewidth]{figures/auc_vs_n_diseased_legend.pdf}
    \par\vspace{0.2em}
    \includegraphics[width=0.95\linewidth]{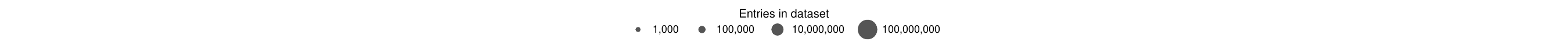}

    \caption{{\textbf{Uniform Manifold Approximation and Projection (UMAP) of learned disease embeddings.} \textbf{a},~Delphi. \textbf{b},~ReClaim. Each point denotes an ICD-10-CM disease token projected from the pre-trained embedding space to two dimensions. Colors indicate ICD-10-CM chapter groups and marker size reflects token frequency in the training corpus. Relative to Delphi, ReClaim showed tighter local neighborhoods and clearer chapter-level structure while preserving continuity between clinically related disease systems.}}
    \label{fig:embedding_umap}
\end{figure}

\clearpage

\begin{figure}[H]
    \centering
    
    \tikz[baseline=(img.south west)]{\node[anchor=south west,inner sep=0](img)%
        {\includegraphics[width=0.48\linewidth]{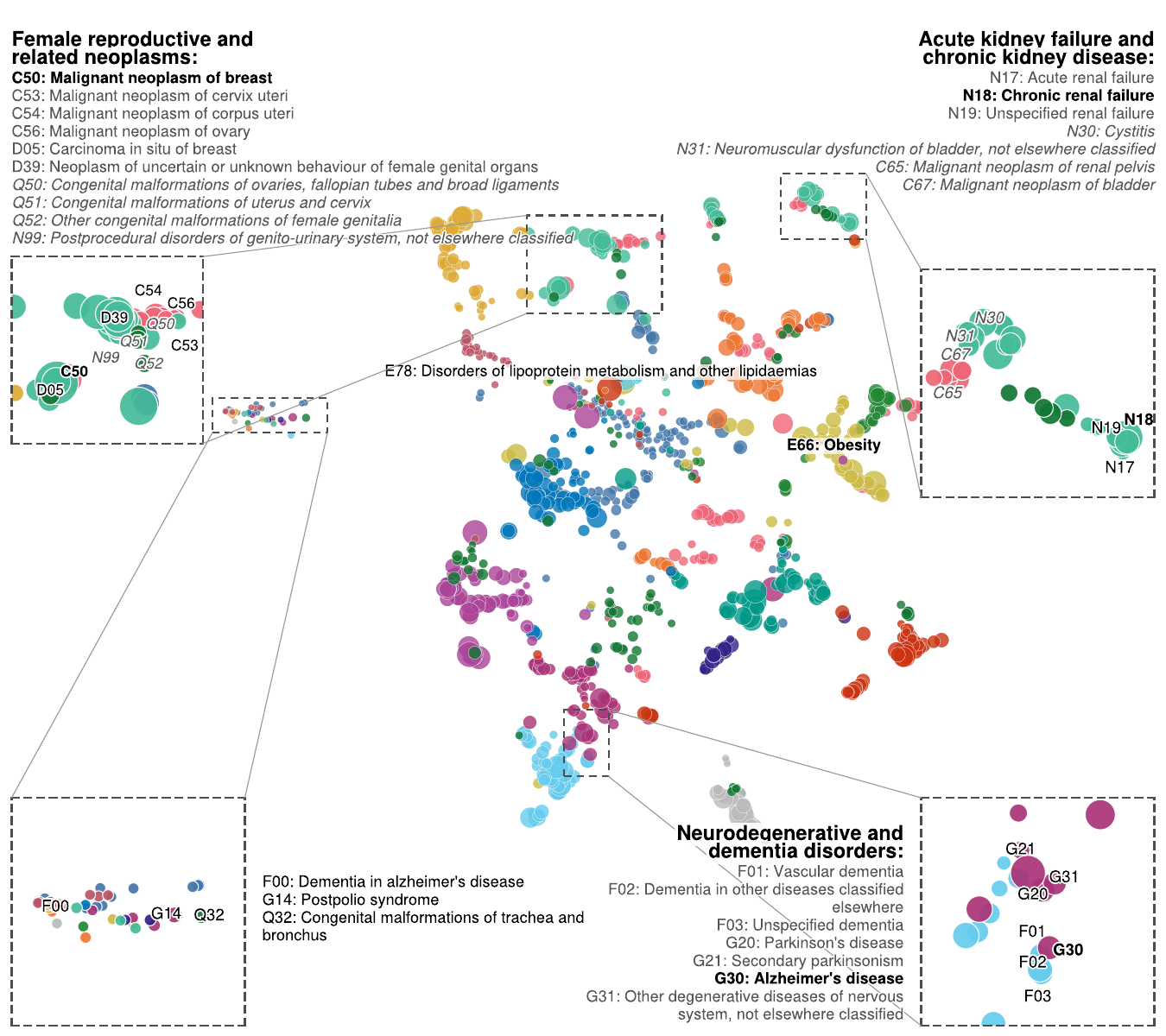}};%
        \node[anchor=south west,font=\normalsize\bfseries] at (img.north west){a. Delphi};}\hfill
    \tikz[baseline=(img.south west)]{\node[anchor=south west,inner sep=0](img)%
        {\includegraphics[width=0.48\linewidth]{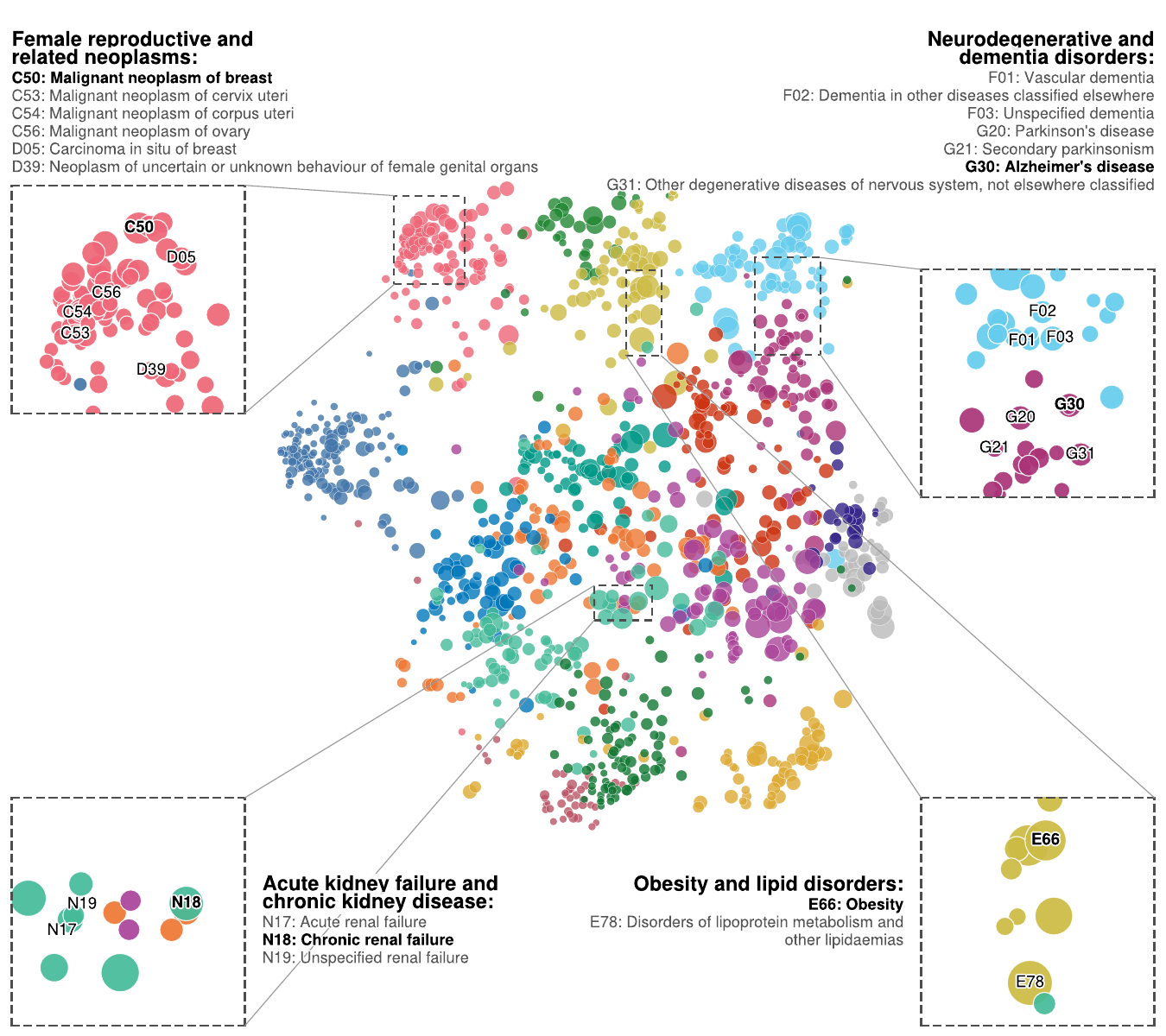}};%
        \node[anchor=south west,font=\normalsize\bfseries] at (img.north west){b. ReClaim};}
    \par\vspace{0.2em}
    \includegraphics[width=0.95\linewidth]{figures/auc_vs_n_diseased_legend.pdf}
    \par\vspace{0.2em}
    \includegraphics[width=0.95\linewidth]{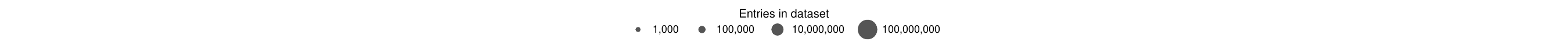}

    \caption{{\textbf{\textit{t}-distributed Stochastic Neighbor Embedding (\textit{t}-SNE) of learned disease embeddings.} \textbf{a},~Delphi. \textbf{b},~ReClaim. Each point denotes an ICD-10-CM disease token projected from the pre-trained embedding space to two dimensions. Colors indicate ICD-10-CM chapter groups and marker size reflects token frequency in the training corpus. Relative to Delphi, ReClaim showed tighter local neighborhoods and clearer chapter-level structure while preserving continuity between clinically related disease systems.}}
    \label{fig:embedding_tsne}
\end{figure}

\FloatBarrier
\clearpage
\subsection*{Supplementary Tables}
\begin{table}[H]
\centering
\caption{Overview of the Merative MarketScan Research Databases.}
\label{tab:marketscan_overview}
\begin{tabular}{
c 
>{\raggedright\arraybackslash}p{0.50\linewidth} 
c 
r
}
\toprule
\textbf{Database} & \textbf{Description} & \textbf{Time range} & \textbf{N} \\
\midrule

CCAE
& Commercial claims for employer-sponsored populations, including employees, early (non-Medicare) retirees, COBRA continuees, and dependents.
& Jan~2008 -- Sep~2024
& 165{,}720{,}281 \\

\midrule

MDCR 
& Medicare claims for Medicare-eligible employees and retirees and their dependents in employer-sponsored supplemental and Medicare Advantage plans; predominantly fee-for-service data.
& Jan~2009 -- Sep~2024
& 11{,}091{,}216 \\

\midrule

MDCD
& Multi-state Medicaid claims capturing healthcare utilization for individuals enrolled in Medicaid programs across participating states, including fee-for-service and managed care plans.
& Jan~2009 -- Jun~2024
& 28{,}787{,}210 \\

\bottomrule
\end{tabular}
\end{table}

\begin{table}[H]
\centering
\caption{Standardized Inclusion and Exclusion Criteria Applied Across Payer Databases}
\label{tab:inclusion_criteria}
\begin{tabular}{p{0.30\textwidth} p{0.60\textwidth}}
\hline
\textbf{Criterion} & \textbf{Description} \\
\hline
Database linkage & CCAE and MDCR databases were merged using shared enrollee identifiers for individuals linkable across payer systems. As a subset of individuals are linkable between CCAE and MDCR via shared enrollee identifiers, whereas MDCD identifiers are unique and non-overlapping across databases.\\[6pt]
Valid enrollee identifier & Consistent enrollee identifier required to enable longitudinal linkage of claims \\[6pt]
Prescription drug coverage & Enrollment periods with documented prescription drug coverage required to ensure complete pharmacy claims for medication exposure assessment \\[6pt]
Minimum enrollment duration & $\geq$180 cumulative days of enrollment required to provide sufficient observation time \\[6pt]
Minimum claims span & $\geq$30-day span between first and last observed claims to exclude transient or fragmented records \\[6pt]
Age restriction & Age at first observed event restricted to 10--110 years to represent realistic clinical populations \\[6pt]
Dual-eligible exclusion & Medicare dual-eligible beneficiaries excluded from the MDCD database to prevent cross-database linkage and data leakage \\
\hline
\end{tabular}
\end{table}

\begin{table}[H]
\centering
\caption{\textbf{Example token sequence for a synthetic enrollee}. Tokenized representation of a synthetic MarketScan enrollee's longitudinal healthcare trajectory, illustrating the output of our data-processing and customized tokenization pipeline. Tokens appear verbatim in chronological order.}
\label{tab:plain_token_sequence}
\scriptsize
\renewcommand{\arraystretch}{1.25}
\begin{tabular}{@{}>{\raggedright\arraybackslash}p{\textwidth}@{}}
\hline
\texttt{<sos> <SEX-1> <DOBYR-1974> <AGE-44>
<ATT-0> <ERLST-CCAE> <PLANTYP-6> <CAP-0> <EGEOLOC-04>
<ATT-12> <NY>
<ATT-12> <NY>
<ATT-10> <VT-outpatient> <DX-PRINCIPAL> <DX-MAJOR\_R07> <DX-MINOR\_9> <DX-MAJOR\_E11> <DX-MINOR\_9> <PROC-PRINCIPAL> <PROC-COMBSTART> <PROC-268400002> <PROC-29303009> <PROC-308561006> <PROC-COMBEND> <PROC-165102003> <PROC-16254007> <PROC-COMBSTART> <PROC-408254005> <PROC-43396009> <PROC-COMBEND> <PROC-CPT499214> <COST-32>
<ATT-0> <VT-pharmacy> <RX-6809> <COST-51>
<ATT-2> <NY>
<ATT-5> <VT-outpatient> <DX-PRINCIPAL> <DX-MAJOR\_I10> <PROC-PRINCIPAL> <PROC-165102003> <PROC-16254007> <PROC-COMBSTART> <PROC-408254005> <PROC-43396009> <PROC-COMBEND> <PROC-45896001> <PROC-34608000> <PROC-CPT499213> <COST-12>
<ATT-0> <VT-pharmacy> <RX-18867> <RX-5487> <COST-51>
<ATT-1> <VT-outpatient> <DX-PRINCIPAL> <DX-MAJOR\_E11> <DX-MINOR\_9> <DX-MAJOR\_E78> <DX-MINOR\_2> <PROC-PRINCIPAL> <PROC-165102003> <PROC-16254007> <PROC-COMBSTART> <PROC-408254005> <PROC-43396009> <PROC-COMBEND> <PROC-63476009> <PROC-45896001> <PROC-34608000> <COST-22>
<ATT-0> <VT-pharmacy> <RX-1991302> <RX-4018> <COST-23>
<ATT-6> <NY>
<ATT-3> <VT-inpatient> <DX-PRINCIPAL> <DX-MAJOR\_R11> <DX-MINOR\_2> <DX-SECONDARY> <DX-MAJOR\_R10> <DX-MINOR\_9> <DX-MAJOR\_R13> <DX-MINOR\_0> <DX-MAJOR\_R13> <DX-MINOR\_10> <DX-MAJOR\_Z98> <DX-MINOR\_84> <PROC-SECONDARY> <PROC-168702005> <PROC-241157000> <PROC-COMBSTART> <PROC-315639002> <PROC-86181006> <PROC-COMBEND> <DS-MISSING> <LS-0> <COST-43>
<ATT-9> <NY>
<ATT-3> <VT-outpatient> <DX-PRINCIPAL> <DX-MAJOR\_E11> <DX-MINOR\_9> <PROC-PRINCIPAL> <PROC-165102003> <PROC-16254007> <PROC-COMBSTART> <PROC-408254005> <PROC-43396009> <PROC-COMBEND> <PROC-45896001> <PROC-COMBSTART> <PROC-117356000> <PROC-43789009> <PROC-COMBEND> <PROC-104154005> <PROC-CPT499214> <COST-22>
<ATT-8> <ERLED-CCAE> <eos>} \\
\hline
\end{tabular}
\end{table}

\begin{table}[H]
\centering
\caption{{\textbf{Example token sequence for a synthetic enrollee in disease prediction post-training}. During disease prediction post-training, an \texttt{<INSTRUCT-DX>} token was randomly inserted into the sequence, and target diagnoses first recorded after this instruction token were used as the prediction targets.}}
\label{tab:posttrain_token_sequence}
\scriptsize
\renewcommand{\arraystretch}{1.25}
\begin{tabular}{@{}>{\raggedright\arraybackslash}p{\textwidth}@{}}
\hline
\texttt{<sos> <SEX-1> <DOBYR-1974> <AGE-44>
<ATT-0> <ERLST-CCAE> <PLANTYP-6> <CAP-0> <EGEOLOC-04>
<ATT-12> <NY>
<ATT-12> <NY>
<ATT-10> <VT-outpatient> <DX-PRINCIPAL> <DX-MAJOR\_R07> <DX-MINOR\_9> <DX-MAJOR\_E11> <DX-MINOR\_9> <PROC-PRINCIPAL> <PROC-COMBSTART> <PROC-268400002> <PROC-29303009> <PROC-308561006> <PROC-COMBEND> <PROC-165102003> <PROC-16254007> <PROC-COMBSTART> <PROC-408254005> <PROC-43396009> <PROC-COMBEND> <PROC-CPT499214> <COST-32>
<ATT-0> <VT-pharmacy> <RX-6809> <COST-51>
<ATT-2> <NY>
<ATT-5> <VT-outpatient> <DX-PRINCIPAL> <DX-MAJOR\_I10> <PROC-PRINCIPAL> <PROC-165102003> <PROC-16254007> <PROC-COMBSTART> <PROC-408254005> <PROC-43396009> <PROC-COMBEND> <PROC-45896001> <PROC-34608000> <PROC-CPT499213> <COST-12>
<ATT-0> <VT-pharmacy> <RX-18867> <RX-5487> <COST-51>
\textcolor{red}{<INSTRUCT-DX>} <DX-MAJOR\_E78> <DX-MAJOR\_R11> <DX-MAJOR\_R10> <DX-MAJOR\_R13> <DX-MAJOR\_Z98>  <eos>} \\
\hline
\end{tabular}
\end{table}

\begin{table}[H]
\centering
\caption{{Example token sequence for a synthetic enrollee in healthcare expenditure post-training. The prompt is the enrollee's trajectory up to the prediction point, followed by the instruction token \texttt{<INSTRUCT-COST>}. The response is a single cost token giving the following calendar year's total expenditure, followed by \texttt{<eos>}; the loss is applied to the response only. Here the outcome year contains two outpatient visits (\texttt{<COST-12>} = \$100, \texttt{<COST-22>} = \$200) and two pharmacy fills (\texttt{<COST-51>} = \$50, \texttt{<COST-23>} = \$2{,}000), so the target is the total \$2{,}350, re-encoded as \texttt{<COST-23>}. Events after the outcome year are not used.}}
\label{tab:posttrain_cost_token_sequence}
\scriptsize
\renewcommand{\arraystretch}{1.25}
\begin{tabular}{@{}>{\raggedright\arraybackslash}p{\textwidth}@{}}
\hline
\texttt{<sos> <SEX-1> <DOBYR-1974> <AGE-44>
<ATT-0> <ERLST-CCAE> <PLANTYP-6> <CAP-0> <EGEOLOC-04>
<ATT-12> <NY>
<ATT-12> <NY>
<ATT-10> <VT-outpatient> <DX-PRINCIPAL> <DX-MAJOR\_R07> <DX-MINOR\_9> <DX-MAJOR\_E11> <DX-MINOR\_9> <PROC-PRINCIPAL> <PROC-COMBSTART> <PROC-268400002> <PROC-29303009> <PROC-308561006> <PROC-COMBEND> <PROC-165102003> <PROC-16254007> <PROC-COMBSTART> <PROC-408254005> <PROC-43396009> <PROC-COMBEND> <PROC-CPT499214> <COST-32>
<ATT-0> <VT-pharmacy> <RX-6809> <COST-51>
<ATT-2> <NY>
\textcolor{red}{<INSTRUCT-COST>} {<COST-23>} <eos>} \\
\hline
\end{tabular}
\end{table}

\begin{table}[H]
\centering

\caption{{\textbf{Disease endpoint coverage in the post-training data.} Positive-example counts in the 100,000 prompt--response pairs used for disease prediction post-training, summarized across all 1,208 evaluation endpoints and the subset of 437 rare endpoints. A positive example contains a target diagnosis first recorded after the instruction position in the response; one response may contain multiple endpoints. Rare endpoints follow the disease prediction benchmark definition (more than 10 and fewer than 500 cases in the 1-million evaluation cohort, retrospective and prospective cases combined; Methods). Coverage columns include endpoints with no positive examples, whereas the median, interquartile range (IQR) and range summarize counts among endpoints with at least one positive example.}}
\label{tab:posttraining_stats}
\begin{tabular}{lcccccc}
\toprule
\makecell{Endpoint\\group} &
\makecell{N\\endpoints} &
\makecell{N with\\$\geq$1 positive} &
\makecell{N with\\0 positives} &
\makecell{Median\\positives} &
IQR &
Range \\
\midrule
All diseases & 1,208 & 1,167 & 41 & 110 & 19-533 & 1-15180 \\
Rare diseases & 437 & 431 & 6 & 17 & 6.5-31 & 1-162 \\
\bottomrule
\end{tabular}
\end{table}

\begin{table}[H]
\centering
\caption{{Distribution of the target in the healthcare expenditure post-training data. One prompt--response pair per enrollee (100,000 pairs, disjoint from the held-out evaluation cohorts). The target is the enrollee's total expenditure in the calendar year after the prediction point, derived by summing the per-claim expenditure tokens of that year and right-censored at \$1{,}000{,}000; it is expressed in the response as a single cost token. IQR, interquartile range; HNHC, high-need, high-cost (next-year total $\geq$ \$30{,}000).}}
\label{tab:posttrain_cost_stats}
\small
\renewcommand{\arraystretch}{1.15}
\begin{tabular}{@{}lr@{}}
\hline
 & \textbf{Next-year total expenditure} \\
\hline
Mean & \$6,365 \\
Median & \$1,060 \\
IQR & \$70--4,400 \\
95th percentile & \$26,091 \\
99th percentile & \$86,400 \\
Zero spending & 22.4\% \\
HNHC ($\geq$ \$30,000) & 4.3\% \\
\hline
\end{tabular}
\end{table}

\begin{table}[H]
\centering
\caption{\textbf{Per-disease prevalence (per 1,000,000) across MarketScan, EHRShot, and YNHH cohorts.} Comparisons are restricted to {1,023} disease endpoints common to all three cohorts. Cohort sizes ($N$) refer to the unique individuals available for {disease prediction} evaluation{; the YNHH analysis cohort excludes 78 individuals with missing sex information to ensure identical evaluation cohorts across models}. Statistical comparisons are paired two-sided Wilcoxon signed-rank tests of per-disease prevalence against MarketScan.}
\label{tab:prevalence_comparison}
\begin{tabular}{lccc}
\toprule
Cohort & Median (IQR) per 1M & Median ratio vs MarketScan & $P$ vs MarketScan \\
\midrule
MarketScan ($N=1{,}000{,}000$) & {1{,}297 (232--5{,}255)}   & 1.00            & --- \\
EHRShot   ($N=5{,}845$)        & {3{,}935 (855--14{,}115)}  & {3.04\,$\times$}  & {$3.9\times10^{-117}$} \\
YNHH      {($N=1{,}902{,}195$)}  & {1{,}341 (257--6{,}049)}   & {1.12\,$\times$}  & {$2.6\times10^{-4}$} \\
\bottomrule
\end{tabular}

\vspace{4pt}
\footnotesize\noindent
{EHRShot prevalence exceeded MarketScan in 89.1\% of overlapping diseases (911/1{,}023), with $\geq 2\times$ in 683 and $\geq 5\times$ in 336; YNHH exceeded MarketScan in 57.1\% (584/1{,}023), with $\geq 2\times$ in 193 and $\geq 5\times$ in 14. Spearman rank correlation was $\rho=0.877$ for MarketScan versus EHRShot and $\rho=0.932$ for MarketScan versus YNHH, indicating that endpoint ranking by prevalence was preserved despite differences in absolute prevalence. Differences across cohorts reflect their distinct sampling frames: MarketScan covers commercially insured individuals, whereas EHRShot and YNHH represent hospital-encountered populations.}
\end{table}

\begin{table}[H]
\centering
\caption{\textbf{Distribution of overlapping diseases by prevalence band (per 1,000,000) in each cohort.} Each cell shows the disease count followed by the percentage of the {1,023} overlapping endpoints that fall within the indicated band.}
\label{tab:prevalence_bands}
\begin{tabular}{lccc}
\toprule
Prevalence band (per 1M) & MarketScan & EHRShot & YNHH \\
\midrule
$<100$           & {132 (12.9\%)} &   0 (0.0\%)  & {131 (12.8\%)} \\
100--499         & {227 (22.2\%)} & {140 (13.7\%)} & {222 (21.7\%)} \\
500--999         & {117 (11.4\%)} & {128 (12.5\%)} & {108 (10.6\%)} \\
1{,}000--4{,}999 & {284 (27.8\%)} & {305 (29.8\%)} & {271 (26.5\%)} \\
5{,}000--9{,}999 & {89 (8.7\%)}  & {120 (11.7\%)} & {124 (12.1\%)} \\
$\geq 10{,}000$  & {174 (17.0\%)} & {330 (32.3\%)} & {167 (16.3\%)} \\
\bottomrule
\end{tabular}

\vspace{4pt}
\footnotesize\noindent
{The EHRShot distribution is shifted toward higher prevalence bands, whereas YNHH is broadly similar to MarketScan. Endpoints with prevalence below 500 per 1{,}000{,}000 account for 35.1\% of MarketScan diseases, 34.5\% of YNHH diseases and 13.7\% of EHRShot diseases; endpoints with prevalence at or above 10{,}000 per 1{,}000{,}000 comprise 17.0\% in MarketScan, 16.3\% in YNHH and 32.3\% in EHRShot.}
\end{table}

\FloatBarrier
\clearpage
\begin{table}[H]
\centering
\caption{Deterministic ordering of simultaneous tokens within the same monthly timestamp. Token ordering follows a two-level hierarchy: (i) event-level order and (ii) within-event token ordering. Event groups occurring within the same monthly timestamp are first ordered according to the event-level order. Within each event, tokens follow a fixed, category-specific sequence listed from top to bottom. For diagnosis (DX) and procedure (PROC) tokens, principal and secondary markers define sub-ordering, and individual codes within each subgroup are ordered by their original dates, after first-occurrence deduplication.}
\label{tab:simultaneous_token_order}
\small
\setlength{\tabcolsep}{4pt}
\renewcommand{\arraystretch}{1.15}

\begin{tabular}{l c l}
\toprule
\textbf{Event type} & \textbf{Event order} & \textbf{Within-event token order} \\
\midrule

\multirow{2}{*}{Anchor} 
& \multirow{2}{*}{1} 
& AGE \\
& & NY \\

\midrule

\multirow{4}{*}{Enrollment start} 
& \multirow{4}{*}{2} 
& ERLST \\
& & PLANTYP \\
& & CAP \\
& & EGEOLOC \\

\midrule

\multirow{8}{*}{Outpatient visit} 
& \multirow{8}{*}{3} 
& VT \\
& & DX-PRINCIPAL \\
& & DX codes (principal) \\
& & DX-SECONDARY \\
& & DX codes (secondary)\\
& & PROC-PRINCIPAL \\
& & PROC codes (principal)\\
& & PROC-SECONDARY \\
& & PROC codes (secondary)\\
& & COST \\

\midrule

\multirow{3}{*}{Pharmacy visit} 
& \multirow{3}{*}{4} 
& VT \\
& & RX \\
& & COST \\

\midrule

\multirow{10}{*}{Inpatient admission} 
& \multirow{10}{*}{5} 
& VT \\
& & DX-PRINCIPAL \\
& & DX codes (principal)\\
& & DX-SECONDARY \\
& & DX codes (secondary)\\
& & PROC-PRINCIPAL \\
& & PROC codes (principal)\\
& & PROC-SECONDARY \\
& & PROC codes (secondary)\\
& & DS \\
& & LS \\
& & COST \\

\midrule

\multirow{1}{*}{Enrollment end} 
& \multirow{1}{*}{6} 
& ERLED \\

\bottomrule
\end{tabular}

\end{table}

\FloatBarrier
\clearpage
\begin{table}[H]
\centering
\caption{\textbf{Vocabulary summary.} Static tokens are not associated with any timestamp. Temporal tokens are aggregated at a specified temporal resolution and associated with a timestamp, so that they can be sorted chronologically. Clinical token counts are data-dependent.}
\label{tab:vocab_summary}
\footnotesize
\setlength{\tabcolsep}{4pt}
\renewcommand{\arraystretch}{1.05}
\begin{tabular}{@{}l l l l l c c@{}}
\toprule
\textbf{Type} & \textbf{Category} & \textbf{\shortstack[l]{Token \\ Name}} & \textbf{Description} & \textbf{\shortstack[l]{Token \\ Example}} & \textbf{\shortstack{Unique \\ Values}} & \textbf{\shortstack{Frequency \\ (Million)}} \\
\midrule
\multirow{2}{*}{Static}
 & Demographic    & SEX      & Enrollee's sex            & <SEX-1>          & 3       & 120 \\
 & Instruct  & INSTRUCT & Instruction for post-training task      & <INSTRUCT-DX>    & 2       &     \\
\midrule
\multirow{16}{*}{Temporal}
 & \multirow{4}{*}{Time}
   & DOBYR   & Birth year                & <DOBYR-1985>    & 116     & 120 \\
 & & AGE     & Age at first event year   & <AGE-45>        & 101     & 120 \\
 & & NY      & New-year marker           & <NY>            & 1       & 379 \\
 & & ATT     & Months to next encounter  & <ATT-5>         & 13      & 4{,}467 \\
\cmidrule(l){2-7}
 & \multirow{5}{*}{Enrollment}
   & ERLST   & Enrollment start flag     & <ERLST-CCAE>    & 4       & 147 \\
 & & PLANTYP & Healthcare Plan type      & <PLANTYP-5>     & 10      & 147 \\
 & & CAP     & Capitation flag           & <CAP-1>         & 3       & 147 \\
 & & EGEOLOC & Geographic location       & <EGEOLOC-18>    & 68      & 147 \\
 & & ERLED   & Enrollment end flag       & <ERLED-CCAE>    & 4       & 147 \\
\cmidrule(l){2-7}
 & \multirow{3}{*}{\shortstack[l]{Clinical \\ code}}
   & DX      & Diagnosis (ICD-10-CM)             & <DX-MAJOR\_E11>   & 5{,}474  & 23{,}054 \\
 & & PROC    & Procedure (SNOMED CT, CPT-4)      & <PROC-229315006>     & 12{,}473 & 17{,}319 \\
 & & RX      & Medication (Ingredient RxNorm)    & <RX-1991302>       & 2{,}473  & 7{,}903 \\
\cmidrule(l){2-7}
 & Expenditure
   & COST    & Cost (aggregated and rounded)     & <COST-23>       & 101     & 3{,}795 \\
\cmidrule(l){2-7}
 & \multirow{3}{*}{\shortstack[l]{Encounter \\ specific}}
   & VT      & visit type of claim       & <VT-outpatient>  & 4       & 3{,}795 \\
 & & DS      & Discharge status           & <DS-1>          & 6       & 34 \\
 & & LS      & Long stay ($\geq$ 7 days)  & <LS-1>          & 3       & 34 \\
\midrule
\multicolumn{5}{@{}l}{\textbf{Total}} & \textbf{20{,}862} & \textbf{$\sim$62{,}000} \\
\bottomrule
\end{tabular}
\end{table}

\FloatBarrier
\clearpage
{\scriptsize
\setlength{\tabcolsep}{4pt}
\setlength{\LTcapwidth}{\textwidth}
\setlength{\LTleft}{0pt}
\setlength{\LTright}{0pt plus 1fill}
\renewcommand{\arraystretch}{1.1}
\begin{longtable}{@{}%
  >{\raggedright\arraybackslash}p{0.08\textwidth}@{\hskip 4pt}%
  >{\centering\arraybackslash}p{0.05\textwidth}@{\hskip 4pt}%
  >{\raggedright\arraybackslash}p{0.16\textwidth}@{\hskip 4pt}%
  >{\raggedright\arraybackslash}p{0.28\textwidth}@{\hskip 4pt}%
  >{\raggedright\arraybackslash}p{0.36\textwidth}@{}}
\caption{\textbf{Walk-through of the example token sequence in Supplementary Table~\ref{tab:plain_token_sequence}.} Each row represents
one logical token group. Rows alternate between time-interval markers (\texttt{ATT-}$N$, with $\Delta$ giving elapsed months) and dated event groups
(with calendar time).}
\label{tab:example_walkthrough_decode} \\
\hline
\textbf{Calendar Time} & \textbf{$\Delta$ (mo)} & \textbf{Event Type} & \textbf{Event Token Segment} & \textbf{Decoded Interpretation} \\
\hline
\endfirsthead
\multicolumn{5}{l}{\textit{Table \thetable\ -- continued from previous page}} \\
\hline
\textbf{Calendar Time} & \textbf{$\Delta$ (mo)} & \textbf{Event Type} & \textbf{Token Segment} & \textbf{Decoded Interpretation} \\
\hline
\endhead
\hline
\multicolumn{5}{r}{\textit{Continued on next page}} \\
\endfoot
\hline
\endlastfoot

          &     & Sequence start    & \texttt{<sos>}                & Start of sequence. \\
          &     & Demographics      & \texttt{<SEX-1>}              & Male. \\
Jan 1974  &     & Birth year        & \texttt{<DOBYR-1974>}         & Year-of-birth anchor, placed at first month. \\
Jan 2018  &     & Age (first event)        & \texttt{<AGE-44>}             & Age 44 at first recorded event (in 2018). \\

          & 0   & Time interval          & \texttt{<ATT-0>}              & Same month (Jan 2018). \\
Jan 2018  &     & Enrollment start  & \texttt{<ERLST-CCAE>}\newline
                                      \texttt{<PLANTYP-6>}\newline
                                      \texttt{<CAP-0>}\newline
                                      \texttt{<EGEOLOC-04>}
          & CCAE plan enrollment begins\newline
            plan type: PPO\newline
            non-capitated\newline
            geographic region: Connecticut \\

          & 12  & Time interval          & \texttt{<ATT-12>}             & Twelve months elapsed. \\
Jan 2019  &     & New-year anchor   & \texttt{<NY>}                 & Calendar-year boundary marker (2019). \\

          & 12  & Time interval          & \texttt{<ATT-12>}             & Twelve months elapsed. \\
Jan 2020  &     & New-year anchor   & \texttt{<NY>}                 & Calendar-year boundary marker (2020). \\

          & 10  & Time interval          & \texttt{<ATT-10>}             & Ten months elapsed. \\
Nov 2020  &     & Outpatient visit  & \texttt{<VT-outpatient>}\newline
                                      \texttt{<DX-PRINCIPAL>}\newline
                                      \texttt{<DX-MAJOR\_R07> <DX-MINOR\_9>}\newline
                                      \texttt{<DX-MAJOR\_E11> <DX-MINOR\_9>}\newline
                                      \texttt{<PROC-PRINCIPAL>}\newline
                                      \texttt{<PROC-COMBSTART> <PROC-268400002>}\newline
                                      \texttt{<PROC-29303009> <PROC-308561006>}\newline
                                      \texttt{<PROC-COMBEND>}\newline
                                      \texttt{<PROC-165102003>}\newline
                                      \texttt{<PROC-16254007>}\newline
                                      \texttt{<PROC-COMBSTART> <PROC-408254005>}\newline
                                      \texttt{<PROC-43396009> <PROC-COMBEND>}\newline
                                      \texttt{<PROC-CPT499214>}\newline
                                      \texttt{<COST-32>}
          & Outpatient visit\newline
            Principal diagnosis:\newline
             - ICD-10-CM: R07.9 Chest pain\newline
             - ICD-10-CM: E11.9 Type 2 diabetes mellitus\newline
            Principal procedure:\newline
             - Atomic SNOMED group mapped from CPT4:93000: Electrocardiogram, routine ECG with at least 12 leads;\newline
             - SNOMED: Metabolic function test\newline
             - SNOMED: Lipid panel\newline
             - Atomic SNOMED group mapped from CPT4:83036: Hemoglobin; glycosylated (A1C) \newline
             - CPT4 fallback: Established outpatient.\newline
            Gross payment $\approx \$300$ \\

          & 0   & Time interval          & \texttt{<ATT-0>}              & Same month (Nov 2020). \\
Nov 2020  &     & Pharmacy visit     & \texttt{<VT-pharmacy>}\newline
                                      \texttt{<RX-6809>}\newline
                                      \texttt{<COST-51>}
          & Pharmacy fill\newline
            - RxNorm: 6809 Metformin\newline
            Gross payment $\approx \$50$. \\

          & 2   & Time interval          & \texttt{<ATT-2>}              & Two months elapsed. \\
Jan 2021  &     & New-year anchor   & \texttt{<NY>}                 & Calendar-year boundary marker (2021). \\

          & 5   & Time interval          & \texttt{<ATT-5>}              & Five months elapsed. \\
Jun 2021  &     & Outpatient visit  & \texttt{<VT-outpatient>}\newline
                                      \texttt{<DX-PRINCIPAL> }\newline
                                      \texttt{<DX-MAJOR\_I10>}\newline
                                      \texttt{<PROC-PRINCIPAL>}\newline
                                      \texttt{<PROC-165102003>}\newline
                                      \texttt{<PROC-16254007>}\newline
                                      \texttt{<PROC-COMBSTART> <PROC-408254005>}\newline
                                      \texttt{<PROC-43396009> <PROC-COMBEND>}\newline
                                      \texttt{<PROC-45896001>}\newline
                                      \texttt{<PROC-34608000>}\newline
                                      \texttt{<PROC-CPT499213>}\newline
                                      \texttt{<COST-12>}
          & Outpatient visit\newline
            Principal diagnosis:\newline
             - ICD-10-CM: I10 Essential Hypertension\newline
            Principal procedure:\newline
             - SNOMED: Metabolic function test\newline
             - SNOMED: Lipid panel\newline
             - Atomic SNOMED group mapped from CPT4:83036: Hemoglobin; glycosylated (A1C) \newline
             - SNOMED: Aspartate aminotransferase \newline
             - SNOMED: Alanine aminotransferase\newline
             - CPT4 fallback: Established outpatient.\newline
            Gross payment $\approx \$100$ \\

          & 0   & Time interval          & \texttt{<ATT-0>}              & Same month. \\
Jun 2021  &     & Pharmacy fill     & \texttt{<VT-pharmacy>}\newline
                                      \texttt{<RX-18867>}\newline
                                      \texttt{<RX-5487>}\newline
                                      \texttt{<COST-51>}
          & Pharmacy visit\newline
            - RxNorm: 18867 Benazepril\newline
            - RxNorm: 5487 Hydrochlorothiazide\newline
            Gross payment $\approx \$50$. \\

          & 1   & Time interval          & \texttt{<ATT-1>}              & One month elapsed. \\
Jul 2021  &     & Outpatient visit  & \texttt{<VT-outpatient>}\newline
                                      \texttt{<DX-PRINCIPAL>}\newline
                                      \texttt{<DX-MAJOR\_E11> <DX-MINOR\_9>}\newline
                                      \texttt{<DX-MAJOR\_E78> <DX-MINOR\_2>}\newline
                                      \texttt{<PROC-PRINCIPAL>}\newline
                                      \texttt{<PROC-165102003>}\newline
                                      \texttt{<PROC-16254007>}\newline
                                      \texttt{<PROC-COMBSTART> <PROC-408254005>}\newline
                                      \texttt{<PROC-43396009> <PROC-COMBEND>}\newline
                                      \texttt{<PROC-63476009>}\newline
                                      \texttt{<PROC-45896001>}\newline
                                      \texttt{<PROC-34608000>}\newline
                                      \texttt{<COST-22>}
          & Outpatient visit\newline
            Principal diagnosis:\newline
             - ICD-10-CM: E11.9 Type 2 diabetes mellitus\newline
             - ICD-10-CM: E78.2 Mixed hyperlipidemia\newline
            Principal procedure:\newline
             - SNOMED: Metabolic function test\newline
             - SNOMED: Lipid panel\newline
             - Atomic SNOMED group mapped from CPT4:83036: Hemoglobin; glycosylated (A1C) \newline
             - SNOMED: Prostate specific antigen \newline
             - SNOMED: Aspartate aminotransferase \newline
             - SNOMED: Alanine aminotransferase\newline
            Gross payment $\approx \$200$ \\

          & 0   & Time interval          & \texttt{<ATT-0>}              & Same month (Jul 2021). \\
Jul 2021  &     & Pharmacy visit     & \texttt{<VT-pharmacy>}\newline
                                      \texttt{<RX-1991302>}\newline
                                      \texttt{<RX-4018>}\newline
                                      \texttt{<COST-23>}
          & Pharmacy fill\newline
            - RxNorm: 1991302 Semaglutide\newline
            - RxNorm: 4018 Ergocalciferol\newline
            Gross payment $\approx \$2{,}000$. \\

          & 6   & Time interval          & \texttt{<ATT-6>}              & Six months elapsed. \\
Jan 2022  &     & New-year anchor   & \texttt{<NY>}                 & Calendar-year boundary marker (2022). \\

          & 3   & Time interval          & \texttt{<ATT-3>}              & Three months elapsed. \\
Apr 2022  &     & Inpatient admission & \texttt{<VT-inpatient>}\newline
                                        \texttt{<DX-PRINCIPAL>}\newline
                                        \texttt{<DX-MAJOR\_R11> <DX-MINOR\_2>}\newline
                                        \texttt{<DX-SECONDARY>}\newline
                                        \texttt{<DX-MAJOR\_R10> <DX-MINOR\_9>}\newline
                                        \texttt{<DX-MAJOR\_R13> <DX-MINOR\_0>}\newline
                                        \texttt{<DX-MAJOR\_R13> <DX-MINOR\_10>}\newline
                                        \texttt{<DX-MAJOR\_Z98> <DX-MINOR\_84>}\newline
                                        \texttt{<PROC-SECONDARY>}\newline
                                        \texttt{<PROC-168702005>}\newline
                                        \texttt{<PROC-241157000>}\newline
                                        \texttt{<PROC-COMBSTART> <PROC-315639002>}\newline
                                        \texttt{<PROC-86181006> <PROC-COMBEND>}\newline
                                        \texttt{<DS-MISSING>}\newline
                                        \texttt{<LS-0>}\newline
                                        \texttt{<COST-43>}
          & Inpatient admission\newline
            Principal diagnosis:\newline
             - ICD-10-CM: R11.2 Nausea with vomiting\newline
            Secondary diagnosis:\newline
             - ICD-10-CM: R10.9 Unspecified abdominal pain\newline
             - ICD-10-CM: R13.0 Aphagia\newline
             - ICD-10-CM: R13.10 Dysphagia\newline
             - ICD-10-CM: Z98.84 Bariatric-surgery status\newline
            Secondary procedure:\newline
             - SNOMED: Plain X-ray of abdomen\newline
             - SNOMED: Upper gastrointestinal tract series\newline
             - Atomic SNOMED group mapped from CPT4:99231: Management of inpatient \newline
            Discharge status: Unknown \newline
            Stay < 7 days\newline
            Gross payment $\approx \$4{,}000$ \\

          & 9   & Time interval          & \texttt{<ATT-9>}              & Nine months elapsed. \\
Jan 2023  &     & New-year anchor   & \texttt{<NY>}                 & Calendar-year boundary marker (2023). \\

          & 3   & Time interval          & \texttt{<ATT-3>}              & Three months elapsed. \\
Apr 2023  &     & Outpatient visit  & \texttt{<VT-outpatient>}\newline
                                      \texttt{<DX-PRINCIPAL>}\newline
                                      \texttt{<DX-MAJOR\_E11> <DX-MINOR\_9>}\newline
                                      \texttt{<PROC-PRINCIPAL>}\newline
                                      \texttt{<PROC-165102003>}\newline
                                      \texttt{<PROC-16254007>}\newline
                                      \texttt{<PROC-COMBSTART> <PROC-408254005>}\newline
                                      \texttt{<PROC-43396009> <PROC-COMBEND>}\newline
                                      \texttt{<PROC-45896001>}\newline
                                      \texttt{<PROC-COMBSTART> <PROC-117356000>}\newline
                                      \texttt{<PROC-43789009> <PROC-COMBEND>}\newline
                                      \texttt{<PROC-104154005>}\newline
                                      \texttt{<PROC-CPT499214>}\newline
                                      \texttt{<COST-22>}
          & Outpatient visit\newline
            Principal diagnosis:\newline
             - ICD-10-CM: E11.9 Type 2 diabetes mellitus\newline
            Principal procedure:\newline
             - SNOMED: Metabolic function test\newline
             - SNOMED: Lipid panel\newline
             - Atomic SNOMED group mapped from CPT4:83036: Hemoglobin; glycosylated (A1C) \newline
             - SNOMED: Aspartate aminotransferase \newline
             - Atomic SNOMED group mapped from CPT4:85027: Obstetric panel \newline
             - SNOMED: Erythrocyte sedimentation rate \newline
             - CPT4 fallback: Established outpatient.\newline
            Gross payment $\approx \$200$ \\

          & 8   & Time interval          & \texttt{<ATT-8>}              & Eight months elapsed. \\
Dec 2023  &     & Enrollment end    & \texttt{<ERLED-CCAE>}         & CCAE enrollment terminates. \\
          &     & Sequence end      & \texttt{<eos>}                & End of sequence. \\

\end{longtable}
}

\FloatBarrier
\clearpage
\begin{table}[H]
\centering
\footnotesize
\setlength{\tabcolsep}{4pt}
\caption{Percentile of population ranked by healthcare expenditures during the year, from the Medical Expenditure Panel Survey Household Component (MEPS-HC). The \$30{,}000 high-need, high-cost threshold used in this study corresponds to the top 5\%.}
\label{tab:healthcare_expenditures_percentile}
\begin{tabular}{lccccc}
\toprule
 & \multicolumn{5}{c}{Annual healthcare expenditures (inflated to 2022 dollars)} \\
\cmidrule(lr){2-6}
Percentile of Population & 2022 & 2021 & 2020 & 2019 & 2018 \\
\midrule
Top 1\%    & \$80,990 or more   & \$92,906 or more   & \$92,920 or more   & \$88,632 or more   & \$83,300 or more   \\
Top 5\%    & \$30,206 or more   & \$32,032 or more   & \$30,650 or more   & \$31,267 or more   & \$30,402 or more   \\
Top 10\%   & \$16,202 or more   & \$17,250 or more   & \$16,175 or more   & \$17,214 or more   & \$16,901 or more   \\
Bottom 50\% & Less than \$1,361 & Less than \$1,471 & Less than \$1,287 & Less than \$1,490 & Less than \$1,519 \\
\bottomrule
\end{tabular}
\end{table}

\FloatBarrier
\clearpage
\begin{table}[H]
\centering
\caption{{Key definitions of the target trial emulation.
Protocol of the reference study~\cite{tang2025association},
implemented identically on both data sources. CI, confidence interval;
DPP4i, dipeptidyl peptidase-4 inhibitor; EASE, expected absolute systematic error; ESRD, end-stage
renal disease; GLP-1RA, glucagon-like peptide-1 receptor agonist; hdPS, high-dimensional propensity
score; IPTW, inverse probability of treatment weighting; MS, MarketScan; NCO, negative control outcome; PS, propensity score; PSM, propensity-score matching;
SE, standard error; SGLT2i, sodium-glucose cotransporter-2 inhibitor; YNHH, Yale New Haven Health.}}
\label{tab:rwe_application_definition}
\footnotesize
\setlength{\tabcolsep}{5pt}
\renewcommand{\arraystretch}{1.15}
\begin{tabular}{@{}p{0.22\textwidth} p{0.74\textwidth}@{}}
\toprule
\textbf{Element} & \textbf{Definition (both sources; the reference's construction)} \\
\midrule
Sources & MS, the MarketScan held-out cohort (CCAE + MDCR + MDCD), disjoint from ReClaim pre-training; YNHH, the Yale New Haven Health EHR (source-ICD-10-CM diagnosis route, the reference's claims semantics) \\
Structure & Two emulations, GLP-1RA vs SGLT2i and GLP-1RA vs DPP4i, each with its own GLP-1RA arm and its own 12-month washout over the two compared classes only (prior use of the third class allowed) \\
Entry & First study-class dispensing (MS) or drug exposure (YNHH) between 2019-01-01 and 2024-09-30; same-day multi-class entries dropped; age 18 or older \\
Eligibility (ever-lookback) & Type 2 diabetes (E11) ever on or before entry required; type 1 diabetes, ESRD or dialysis, any prior occurrence of the six outcomes, and any prior antidepressant or antipsychotic use excluded; at least one medical claim (MS) or recorded diagnosis (YNHH) in the prior 365 days \\
Outcomes & Six new-onset psychiatric outcomes: depression (F32, F33), anxiety disorders (F40 to F45, F48), bipolar disorder (F31), schizophrenia and other psychotic disorders (F20 to F25, F28, F29), other mood disorders (F30, F34, F39), suicidal ideation (R45.851, V62.84); Cox time to first event from entry + 1 day; censoring at disenrollment or 2024-09-30 (MS) and at death or 2024-09-30 (YNHH) \\
Propensity-score specifications (6) & Unadjusted; Expert (the reference's Table S5 covariate set, 58 covariates buildable on both sources, drop-not-substitute); hdPS (demographics + 500 empirically selected covariates); ReClaim-S, -M, -L (age, sex and the ReClaim patient embedding; ridge logistic PS with deterministic 5-fold cross-validation) \\
Estimators (one PS per specification) & Stabilized IPTW with a weighted Cox model and robust SE, weights truncated at the 1st and 99th percentiles (primary); 1:1 nearest-neighbor PSM, caliper 0.2 SD of the PS \\
Negative controls & The reference's 24-outcome NCO panel (Supplementary Table~\ref{tab:rwe_nco_icd10}) with the reference's 0.1\% follow-up-risk rule (23 estimable on MS, 24 on YNHH); an empirical null fitted per cell, from which EASE is computed \\
\bottomrule
\end{tabular}
\end{table}

\FloatBarrier
\clearpage
\begin{table}[H]
\centering
\caption{{Negative control outcomes.
The 24-outcome negative-control panel reported by the reference study~\cite{tang2025association}
(its Supplement Table S4), used verbatim with the reference's rule excluding outcomes below 0.1\%
follow-up risk (23 estimable on claims, 24 on YNHH), with the ICD-10-CM code sets of the pipeline's
lookup table (prefix match on the source diagnosis codes; the same sets on both sources).
NCO, negative control outcome; YNHH, Yale New Haven Health electronic health records.}}
\label{tab:rwe_nco_icd10}
\footnotesize
\setlength{\tabcolsep}{5pt}
\renewcommand{\arraystretch}{1.1}
\begin{tabular}{@{}c p{0.50\textwidth} p{0.36\textwidth}@{}}
\toprule
\textbf{\#} & \textbf{Negative control outcome} & \textbf{ICD-10-CM codes} \\
\midrule
1 & Abnormal pupil & H21.56, H57.03, H57.0 \\
2 & Abrasion and friction burn of multiple sites & T07 \\
3 & Abrasion and friction burn without infection & S00, S10, S20, S30, S40, S50, S60, S70, S80, S90 \\
4 & Acquired hallux valgus & M20.1 \\
5 & Acquired keratoderma & L85.1, L85.0, L11.0 \\
6 & Anal and rectal polyp & K62.0, K62.1 \\
7 & Benign paroxysmal positional vertigo & H81.1 \\
8 & Calcaneal spur & M77.3 \\
9 & Contact dermatitis & L23, L24, L25 \\
10 & Contusion of knee & S80.0 \\
11 & Derangement of knee & M23 \\
12 & Deviated nasal septum & J34.2 \\
13 & Epidermoid cyst & L72.0 \\
14 & Foreign body in ear & T16 \\
15 & Ganglion cyst & M67.4 \\
16 & Hammer toe & M20.4 \\
17 & Impacted cerumen & H61.2 \\
18 & Macular drusen & H35.36 \\
19 & Mechanical complication of internal orthopedic device, implant or graft & T84 \\
20 & Nontoxic multinodular goiter & E04.2 \\
21 & Presbyopia & H52.4 \\
22 & Regular astigmatism & H52.22 \\
23 & Tear film insufficiency & H04.12 \\
24 & Tooth loss & K08.1, K08.4 \\
\bottomrule
\end{tabular}
\end{table}

\FloatBarrier
\clearpage
\begin{table}[H]
\centering
\caption{Configurations for the three ReClaim models trained from scratch with our custom tokenizer and vocabulary.}
\label{tab:qwen3_arch_configs}
\begin{tabular}{lccc}
\hline
\textbf{Configuration} & \textbf{ReClaim-S (140M)} & \textbf{ReClaim-M (700M)} & \textbf{ReClaim-L (1.7B)} \\
\hline
Hidden size ($d_{\text{model}}$) & 1,024 & 2,048 & 2,048 \\
Transformer layers ($L$) & 8 & 16 & 32 \\
Attention heads ($n_{\text{head}}$) & 16 & 24 & 32 \\
Key-value heads ($n_{\text{kv}}$) & 8 & 12 & 16 \\
Intermediate size ($d_{\text{ffn}}$) & 2,048 & 4,096 & 4,096 \\
Max position embeddings & 4,096 & 4,096 & 4,096 \\
Vocabulary size & 20,862 & 20,862 & 20,862 \\
\hline
\end{tabular}
\end{table}

\begin{table}[H]
\centering

\caption{{\textbf{Training configurations and computational costs for ReClaim.} Hardware, token counts, optimization settings and training time are reported for ReClaim-S, -M and -L during pre-training and task-specific post-training for disease prediction and healthcare expenditure forecasting. Each post-training task uses 100,000 prompt--response pairs, with the loss restricted to the response. Global batch size is the total batch across all GPUs. Training time denotes elapsed wall-clock hours; GPU-hours are the product of training time and the number of GPUs, subject to rounding. Because pre-training uses different GPU types across model scales, the reported times are not normalized for hardware performance.}}
\label{tab:training_details}
\resizebox{\textwidth}{!}{%
\begin{tabular}{lccccccccc}
\toprule
&
\makecell{\textbf{GPU}\\\textbf{Model}} &
\makecell{\textbf{No. of}\\\textbf{GPUs}} &
\makecell{\textbf{Training}\\\textbf{Tokens}} &
\makecell{\textbf{Training}\\\textbf{Steps}} &
\textbf{Solver} &
\makecell{\textbf{Learning}\\\textbf{Rate}} &
\makecell{\textbf{Global}\\\textbf{Batch Size}} &
\makecell{\textbf{Training}\\\textbf{Time (h)}} &
\makecell{\textbf{GPU}\\\textbf{Hours}} \\
\midrule

\multicolumn{10}{c}{\textbf{Pre-training}} \\
\midrule
ReClaim-S & B200 & 8  & 56,772,146,947 & 54,380 & AdamW & $5\times10^{-4}$ & 256 & 22.00 & 176.00 \\
ReClaim-M & B200 & 8  & 56,772,146,947 & 54,380 & AdamW & $5\times10^{-4}$ & 256 & 68.00 & 544.00 \\
ReClaim-L & H200 & 24 & 56,772,146,947 & 72,506 & AdamW & $5\times10^{-4}$ & 192 & 60.00 & 1,440.00 \\

\midrule
\multicolumn{10}{c}{\textbf{{Post-training (Disease prediction)}}} \\
\midrule
ReClaim-S & B200 & 8 & 30,385,136 & 3,125 & AdamW & $2\times10^{-5}$ & 32 & 0.13 & 1.04 \\
ReClaim-M & B200 & 8 & 30,385,136 & 3,125 & AdamW & $2\times10^{-5}$ & 32 & 0.25 & 2.00 \\
ReClaim-L & B200 & 8 & 30,385,136 & 3,125 & AdamW & $2\times10^{-5}$ & 32 & 0.50 & 4.00 \\

\midrule
\multicolumn{10}{c}{\textbf{Post-training (Expenditure forecasting)}} \\
\midrule
ReClaim-S & B200 & 8 & 89,850,126 & 6,250 & AdamW & $3\times10^{-5}$ & 32 & 0.28 & 2.20 \\
ReClaim-M & B200 & 8 & 89,850,126 & 6,250 & AdamW & $3\times10^{-5}$ & 32 & 0.76 & 6.10 \\
ReClaim-L & B200 & 8 & 89,850,126 & 3,125 & AdamW & $3\times10^{-5}$ & 64 & 1.27 & 10.10 \\

\bottomrule
\end{tabular}%
}
\end{table}

\FloatBarrier
\clearpage
\begin{table}[H]
\centering

\scriptsize
\setlength{\tabcolsep}{3.2pt}
\renewcommand{\arraystretch}{1.12}
\caption{{\textbf{Harmonization and concordance of the three-model disease prediction evaluation cohorts.} For each dataset and prediction horizon, endpoint-specific case--control cohorts were constructed before model inference and shared by ReClaim, Delphi and LightGBM. Comparisons were restricted to the intersection of mapped endpoints with non-missing ROC-AUC in all three model outputs. Endpoint identity was established using canonical three-character ICD-10-CM category strings rather than model-specific vocabulary indices. Case and control evaluations are patient--endpoint evaluations summed over all eligible endpoint-specific sex-by-age strata. An evaluable stratum contained at least one case, at least one control and at least two patients. Concordant endpoints had identical case counts, control counts and numbers of evaluable strata across the shared cohort file and all three model outputs. Complete concordance confirms that every cross-model comparison used the same endpoints, patients and prediction points. A separate audit of the input histories before the stored prediction positions found prior occurrences of the target category in 0 of 5{,}800{,}059 retrospective case evaluations across all 1{,}208 endpoints.}}
\label{tab:endpoint_harmonization}
\begin{tabular}{p{0.20\textwidth} c r r r c}
\hline
\textbf{Dataset} & \makecell{\textbf{Endpoints}\\(ReClaim/Delphi/LightGBM)} & \makecell[r]{\textbf{Case}\\\textbf{evaluations}} & \makecell[r]{\textbf{Control}\\\textbf{evaluations}} & \makecell[r]{\textbf{Evaluable}\\\textbf{strata}} & \makecell{\textbf{Concordant}\\\textbf{endpoints}} \\
\hline
Retrospective MarketScan & 1{,}208/1{,}208/1{,}208 & 5{,}800{,}059 & 959{,}524{,}463 & 16{,}049 & 1{,}208/1{,}208 (100\%) \\
Prospective MarketScan   & 1{,}131/1{,}131/1{,}131 &   514{,}631 &  91{,}368{,}222 & 11{,}207 & 1{,}131/1{,}131 (100\%) \\
EHRShot                  & 1{,}024/1{,}024/1{,}024 &    76{,}668 &   4{,}621{,}165 &  8{,}250 & 1{,}024/1{,}024 (100\%) \\
YNHH                     & 1{,}204/1{,}204/1{,}204 & 10{,}935{,}889 & 1{,}999{,}015{,}295 & 16{,}739 & 1{,}204/1{,}204 (100\%) \\
\hline
\end{tabular}
\end{table}

\FloatBarrier
\clearpage
\begin{table}[H]
\centering

\scriptsize
\setlength{\tabcolsep}{3.5pt}
\renewcommand{\arraystretch}{1.08}
\caption{{ICD-10-CM endpoints excluded or unavailable in each disease prediction evaluation. For Fig.~\ref{fig:auc_distribution}a, listed endpoints were excluded from the retrospective prevalence-stratified analysis because they had ten or fewer positive cases. For the prospective, EHRShot and YNHH evaluations, listed endpoints were present in the retrospective reference set but absent from the corresponding three-model shared evaluation set because ICD-10-CM code occurrence and case availability differed across datasets.}}
\label{tab:low_count_endpoints}
\begin{tabular}{p{0.20\textwidth} p{0.74\textwidth}}
\hline
\textbf{Evaluation} & \textbf{Excluded or unavailable ICD-10-CM codes} \\
\hline
Fig. 4a (retrospective) & A17, A19, A22, A23, A24, A25, A30, A32, A33, A55, A67, A71, A75, A83, A91, A94, A95, B03, B46, B51, B52, B55, B57, B73, B74, B75, B79, B81, C58, D77, E35, E45, E68, F24, G01, G07, G53, G55, H22, H75, H94, I41, J63, K23, K67, L00, L14, L54, N07, N22, N33, N37, N74, P10, P11, P13, P15, P23, P26, P50, P51, P53, P58, Q56, Q73, Q86, Q97 \\
\hline
Prospective & A01, A17, A20, A22, A23, A25, A30, A32, A33, A43, A67, A71, A75, A78, A81, A83, A89, A91, A95, B03, B26, B42, B47, B52, B73, B74, B75, B77, B79, B90, B91, C12, C13, C33, D77, E15, E45, E52, E68, F24, G01, G02, G07, G59, G94, H28, H94, I01, I02, I32, I41, J64, K23, K67, L00, L14, M91, N07, N33, N37, N74, P10, P13, P23, P25, P26, P39, P51, P53, P58, P71, Q41, Q69, Q73, Q77, Q86, Q97 \\
\hline
EHRShot & A00, A01, A17, A20, A22, A24, A25, A26, A27, A28, A30, A33, A35, A37, A39, A44, A51, A55, A58, A66, A67, A70, A71, A75, A77, A78, A79, A82, A84, A85, A86, A89, A90, A91, A92, A93, A94, A95, B03, B05, B06, B20, B26, B39, B40, B42, B43, B50, B51, B52, B53, B54, B55, B57, B60, B65, B66, B68, B71, B73, B74, B75, B76, B77, B79, B80, B81, B82, B87, B90, C12, C37, C39, C58, C60, D60, D74, D77, D81, E25, E30, E35, E45, E52, E64, E68, E76, E77, F24, F82, G01, G05, G07, G10, G55, G59, H22, H28, H32, H36, H42, H67, H94, I02, I41, J05, J66, K23, K67, L00, L14, L62, L86, L99, M07, M42, M49, M93, N01, N07, N16, N22, N37, N51, N74, N98, O07, O08, O29, O85, O88, O89, P00, P03, P07, P10, P11, P12, P13, P14, P15, P22, P23, P25, P27, P28, P36, P38, P39, P50, P51, P52, P53, P54, P55, P58, P59, P61, P70, P71, P78, P83, P92, P94, Q00, Q02, Q11, Q17, Q30, Q35, Q36, Q37, Q41, Q42, Q50, Q56, Q69, Q70, Q71, Q72, Q73, Q86, Q91, Q98 \\
\hline
YNHH & A33, P26, P53, P58 \\
\hline
\end{tabular}
\end{table}

\FloatBarrier
\clearpage
\begin{table}[H]
\centering

\footnotesize
\setlength{\tabcolsep}{3.5pt}
\renewcommand{\arraystretch}{1.05}
\caption{{Disease endpoints where post-training ROC-AUC is lower than its corresponding pre-training checkpoint (24 of 1{,}208 endpoints; 2.0\%). Sorted by $\Delta$ROC-AUC (post $-$ pre), most negative first. Pre and Post columns report disease-level ROC-AUC for ReClaim-L before and after task-specific post-training; $\Delta$ is the percentage-point change; $N_{+}$ is the number of positive cases in the held-out retrospective test cohort.}}
\label{tab:posttrain_worse_diseases}
\begin{tabular}{l p{0.33\textwidth} p{0.20\textwidth} r r r r}
\hline
\textbf{ICD} & \textbf{Disease} & \textbf{Chapter} & \makecell[r]{\textbf{Pre}\\\textbf{ROC-AUC}} & \makecell[r]{\textbf{Post}\\\textbf{ROC-AUC}} & \textbf{$\Delta$ (pp)} & \textbf{$N_{+}$}\\
\hline
E68 & Sequelae of hyperalimentation & IV. Metabolic Diseases & 0.863 & 0.573 & $-$28.97 & 2 \\
D77 & Other disorders of blood and blood-forming organs in diseases classified elsewhere & III. Blood \& Immune Disorders & 0.901 & 0.673 & $-$22.83 & 2 \\
A94 & Unspecified arthropod-borne viral fever & I. Infectious Diseases & 0.937 & 0.731 & $-$20.59 & 2 \\
A95 & Yellow fever & I. Infectious Diseases & 0.787 & 0.606 & $-$18.13 & 2 \\
P26 & Pulmonary hemorrhage originating in the perinatal period & XVI. Perinatal Conditions & 0.952 & 0.772 & $-$18.03 & 2 \\
P13 & Birth injury to skeleton & XVI. Perinatal Conditions & 0.779 & 0.604 & $-$17.54 & 3 \\
L14 & Bullous disorders in diseases classified elsewhere & XII. Skin Diseases & 0.904 & 0.766 & $-$13.81 & 1 \\
B42 & Sporotrichosis & I. Infectious Diseases & 0.772 & 0.671 & $-$10.06 & 13 \\
C33 & Malignant neoplasm of trachea & II. Neoplasms & 0.797 & 0.712 & $-$8.52 & 30 \\
P25 & Interstitial emphysema and related conditions originating in the perinatal period & XVI. Perinatal Conditions & 0.675 & 0.611 & $-$6.39 & 18 \\
B81 & Other intestinal helminthiases, not elsewhere classified & I. Infectious Diseases & 0.727 & 0.673 & $-$5.36 & 10 \\
K67 & Disorders of peritoneum in infectious diseases classified elsewhere & XI. Digestive Diseases & 0.799 & 0.761 & $-$3.81 & 8 \\
A67 & Pinta [carate] & I. Infectious Diseases & 0.561 & 0.526 & $-$3.50 & 6 \\
P58 & Neonatal jaundice due to other excessive hemolysis & XVI. Perinatal Conditions & 0.774 & 0.742 & $-$3.20 & 9 \\
B91 & Sequelae of poliomyelitis & I. Infectious Diseases & 0.764 & 0.733 & $-$3.10 & 66 \\
A33 & Tetanus neonatorum & I. Infectious Diseases & 0.689 & 0.662 & $-$2.72 & 2 \\
O25 & Malnutrition in pregnancy & XV. Pregnancy \& Childbirth & 0.802 & 0.781 & $-$2.07 & 5{,}318 \\
G14 & Postpolio syndrome & VI. Nervous System Diseases & 0.549 & 0.536 & $-$1.32 & 67 \\
A55 & Chlamydial lymphogranuloma venereum & I. Infectious Diseases & 0.717 & 0.706 & $-$1.08 & 7 \\
H94 & Other disorders of ear in diseases classified elsewhere & VIII. Ear Diseases & 0.864 & 0.854 & $-$1.00 & 5 \\
F72 & Severe mental retardation & V. Mental Disorders & 0.972 & 0.963 & $-$0.92 & 105 \\
E45 & Retarded development following protein-energy malnutrition & IV. Metabolic Diseases & 0.793 & 0.785 & $-$0.88 & 4 \\
A23 & Brucellosis & I. Infectious Diseases & 0.778 & 0.773 & $-$0.52 & 6 \\
O61 & Failed induction of labour & XV. Pregnancy \& Childbirth & 0.720 & 0.716 & $-$0.37 & 458 \\
\hline
\end{tabular}
\end{table}

\FloatBarrier
\clearpage
\begin{table}[H]
\centering

\scriptsize
\setlength{\tabcolsep}{3pt}
\renewcommand{\arraystretch}{1.15}
\caption{{Related-code exclusion sensitivity analysis for ICD-defined diabetes endpoints. For each target in E11 (type 2 diabetes), O24 (diabetes in pregnancy) and R73 (elevated glucose/prediabetes), the other two categories were treated as related codes. Carriers were excluded if either related category occurred anywhere in their observed trajectory. We evaluated exclusion from controls alone and from both cases and controls. ReClaim-L, LightGBM and Delphi were evaluated on the same retained cohorts at unchanged prediction positions, without retraining. ReClaim-L cached logits were rescored, LightGBM used the published models and cached inference features, and Delphi was rerun to obtain per-patient scores. Counts are patient--endpoint evaluations. R73, which is outside the 1,208-endpoint benchmark, was constructed using the same cohort-construction protocol and evaluated only with ReClaim-L; LightGBM has no R73 model and Delphi has no ICD-10 R-chapter endpoint. Dashes indicate unavailable results. All model columns report mean ROC-AUC.}}
\label{tab:icd_related_exclusion}
\resizebox{\linewidth}{!}{\begin{tabular}{l p{0.38\textwidth} r r r r r}
\hline
\textbf{Target} & \textbf{Cohort} & \textbf{Cases} & \textbf{Controls} & \textbf{ReClaim-L} & \textbf{LightGBM} & \textbf{Delphi} \\
\hline
E11 & Original cohort & 26,724 & 794,439 & 0.738 & 0.716 & 0.563 \\
E11 & Exclude related codes from controls & 26,724 & 714,149 & 0.754 & 0.731 & 0.562 \\
E11 & Exclude related codes from cases and controls & 14,415 & 714,149 & 0.728 & 0.700 & 0.562 \\
\hline
O24 & Original cohort & 3,534 & 316,104 & 0.752 & 0.631 & 0.662 \\
O24 & Exclude related codes from controls & 3,534 & 279,979 & 0.765 & 0.633 & 0.671 \\
O24 & Exclude related codes from cases and controls & 2,184 & 279,979 & 0.734 & 0.624 & 0.652 \\
\hline
R73 & Original cohort & 42,669 & 793,275 & 0.748 & --- & --- \\
R73 & Exclude related codes from controls & 42,669 & 696,660 & 0.746 & --- & --- \\
R73 & Exclude related codes from cases and controls & 28,082 & 696,660 & 0.741 & --- & --- \\
\hline
\end{tabular}}
\end{table}

\FloatBarrier
\clearpage
\begin{table}[H]
\centering

\footnotesize
\setlength{\tabcolsep}{3.5pt}
\renewcommand{\arraystretch}{1.05}
\caption{{ICD-10-CM codes mapped to the Phecode-defined phenotype prediction tasks. When a listed code has mapped descendants in the source Phecode hierarchy, the parent code denotes all of those mapped subcategories.}}
\label{tab:phecode_icd_mapping}
\begin{tabular}{p{0.27\textwidth} p{0.67\textwidth}}
\hline
\textbf{Phecode phenotype} & \textbf{Mapped ICD-10-CM codes} \\
\hline
CV\_401: Hypertension & I10, I11, I12, I13, I15, I16, I67.4, N26.2, O10, O11, O13, O16, P29.2 \\
\hline
EM\_239: Hyperlipidemia & E78.0, E78.1, E78.2, E78.3, E78.4, E78.5 \\
\hline
EM\_236.1: Obesity & E66.0, E66.1, E66.2, E66.8, E66.9, O99.21, Z68.3, Z68.4, Z68.54 \\
\hline
RE\_462: Sinusitis & J01, J32 \\
\hline
MB\_286.2: Major depressive disorder & F32, F33 \\
\hline
MB\_288: Anxiety and anxiety disorders & F01.54, F01.A4, F01.B4, F01.C4, F02.84, F02.A4, F02.B4, F02.C4, F03.94, F03.A4, F03.B4, F03.C4, F06.4, F10.180, F10.280, F10.980, F12.180, F12.280, F12.980, F13.180, F13.280, F13.980, F14.180, F14.280, F14.980, F15.180, F15.280, F15.980, F16.180, F16.280, F16.980, F18.180, F18.280, F18.980, F19.180, F19.280, F19.980, F40, F41, F43.22, F43.23, F93.0 \\
\hline
CA\_138: Benign neoplasm of the skin & D22, D23, I78.1, Q82.5 \\
\hline
GI\_511: Gastro-esophageal reflux disease (GERD) & K21, P78.83 \\
\hline
MS\_708: Osteoarthritis & M15, M16, M17, M18, M19 \\
\hline
NS\_331: Headache & G43, G44, R51 \\
\hline
EM\_202.2: Type 2 diabetes & E11, O24.1 \\
\hline
EM\_204.5: Prediabetes & R73.03 \\
\hline
\end{tabular}
\end{table}

\FloatBarrier
\clearpage
\begin{table}[H]
\centering

\scriptsize
\setlength{\tabcolsep}{2.5pt}
\renewcommand{\arraystretch}{1.05}
\caption{{Performance on clinically curated Phecode prediction tasks. For each phenotype, constituent ICD codes were mapped to a common Phecode definition so that a patient with any mapped code was treated as phenotype-positive. Evaluation case and control counts are shown for each endpoint. The reported ROC-AUC is the unweighted mean across 16 sex-by-age strata. For each endpoint, both models were evaluated on the same retrospective MarketScan test cohort using all eligible controls. The baseline comprises one independently fitted bag-of-words LightGBM model per endpoint; ReClaim-L FT was jointly fine-tuned once with a 12-label multi-label head, whose logits were used directly for prediction. The mean row summarizes only the 12 original phenotypes. The two additional diabetes rows are sensitivity evaluations on mutually exclusive cohorts: for each target, patients with any code for the other diabetes phenotype anywhere in their record were excluded from both cases and controls. The existing models were rescored without retraining, and these rows are excluded from the mean.}}
\label{tab:phecode_sensitivity}
\begin{tabular}{l p{0.28\textwidth} r r r r}
\hline
\textbf{Phecode} & \textbf{Phenotype} & \makecell[r]{\textbf{Evaluation}\\\textbf{cases}} & \makecell[r]{\textbf{Evaluation}\\\textbf{controls}} & \makecell[r]{\textbf{LightGBM}\\\textbf{ROC-AUC}} & \makecell[r]{\textbf{ReClaim-L FT}\\\textbf{ROC-AUC}} \\
\hline
CV\_401 & Hypertension & 73,380 & 524,468 & 0.798 & 0.835 \\
EM\_239 & Hyperlipidemia & 84,725 & 570,074 & 0.788 & 0.825 \\
EM\_236.1 & Obesity & 62,167 & 631,661 & 0.832 & 0.877 \\
RE\_462 & Sinusitis & 84,496 & 416,534 & 0.835 & 0.871 \\
MB\_286.2 & Major depressive disorder & 52,347 & 316,574 & 0.878 & 0.905 \\
MB\_288 & Anxiety disorders & 71,745 & 266,710 & 0.882 & 0.909 \\
CA\_138 & Benign neoplasm of skin & 58,407 & 401,183 & 0.854 & 0.873 \\
GI\_511 & Gastroesophageal reflux disease & 61,014 & 682,864 & 0.827 & 0.851 \\
MS\_708 & Osteoarthritis & 58,360 & 751,835 & 0.815 & 0.846 \\
NS\_331 & Headache & 68,273 & 279,769 & 0.850 & 0.884 \\
EM\_202.2 & Type 2 diabetes & 28,461 & 633,143 & 0.842 & 0.861 \\
EM\_204.5 & Prediabetes & 17,182 & 738,182 & 0.840 & 0.860 \\
\hline
\textbf{Mean} & \textbf{All 12 phenotypes} & -- & -- & \textbf{0.837} & \textbf{0.866} \\
\hline
\multicolumn{6}{l}{\textit{Mutually exclusive diabetes cohorts (sensitivity analysis)}} \\
EM\_202.2 & Type 2 diabetes, excluding prediabetes & 23,117 & 613,478 & 0.839 & 0.856 \\
EM\_204.5 & Prediabetes, excluding type 2 diabetes & 10,983 & 738,182 & 0.805 & 0.830 \\
\hline
\end{tabular}
\end{table}

\FloatBarrier
\clearpage
\begin{table}[H]
\centering

\footnotesize
\setlength{\tabcolsep}{2.5pt}
\renewcommand{\arraystretch}{1.05}
\caption{{Precision--recall performance for ten diagnostic categories selected by first restricting to ReClaim-L ROC-AUC $>0.85$, then ranking eligible categories by observed positive-case frequency in ascending order. Among several eligible categories tied at one positive case, P53 was randomly selected as a single representative. Metrics use the same pooled case--control cohort and model scores as the ROC-AUC analysis (offset, 365.25 days; ages, 20--100 years). PR-AUC is average precision computed over the full control set without negative subsampling or a fixed operating threshold. Prevalence is $N_{+}/(N_{+}+N_{-})$, the expected PR-AUC of a random classifier. The PR-AUC ratio is ReClaim-L PR-AUC divided by LightGBM PR-AUC. All three models were evaluated on the identical cohort and split; rows are ordered by the number of positive cases.}}
\label{tab:rare_disease_prauc}
\resizebox{\linewidth}{!}{\begin{tabular}{l p{0.25\textwidth} r r r r r r}
\hline
\textbf{ICD} & \textbf{Disease} & \textbf{$N_{+}$} & \textbf{Prevalence} & \makecell[r]{\textbf{LightGBM}\\\textbf{PR-AUC}} & \makecell[r]{\textbf{Delphi}\\\textbf{PR-AUC}} & \makecell[r]{\textbf{ReClaim-L}\\\textbf{PR-AUC}} & \makecell[r]{\textbf{PR-AUC ratio}\\\textbf{ReClaim-L / LGBM}} \\
\hline
P53$^{\dagger}$ & Hemorrhagic disease of newborn & 1 & $1.20\times10^{-5}$ & $1.20\times10^{-5}$ & $8.56\times10^{-5}$ & $8.86\times10^{-5}$ & 7.4 \\
N22$^{\dagger}$ & Urinary calculus in diseases classified elsewhere & 4 & $2.67\times10^{-5}$ & $1.70\times10^{-4}$ & $4.38\times10^{-3}$ & $2.64\times10^{-3}$ & 15.5 \\
N07 & Hereditary nephropathy, not elsewhere classified & 10 & $1.55\times10^{-5}$ & $6.55\times10^{-5}$ & $2.59\times10^{-3}$ & 0.012 & 179.2 \\
H42 & Glaucoma in diseases classified elsewhere & 28 & $4.67\times10^{-5}$ & $4.00\times10^{-4}$ & $3.66\times10^{-3}$ & $1.52\times10^{-3}$ & 3.8 \\
F21 & Schizotypal disorder & 31 & $4.34\times10^{-5}$ & $1.98\times10^{-4}$ & $2.03\times10^{-3}$ & $3.26\times10^{-3}$ & 16.5 \\
F88 & Other disorders of psychological development & 45 & $7.12\times10^{-5}$ & $4.42\times10^{-4}$ & 0.013 & 0.014 & 31.4 \\
F78 & Other intellectual disabilities & 47 & $8.06\times10^{-5}$ & $9.37\times10^{-4}$ & 0.031 & 0.046 & 49.4 \\
F72 & Severe intellectual disabilities & 105 & $1.24\times10^{-4}$ & 0.038 & 0.068 & 0.110 & 2.9 \\
J66 & Airway disease due to specific organic dust & 114 & $1.28\times10^{-4}$ & $5.30\times10^{-4}$ & $1.99\times10^{-3}$ & 0.049 & 93.1 \\
F70 & Mild intellectual disabilities & 270 & $3.03\times10^{-4}$ & $3.91\times10^{-3}$ & 0.047 & 0.075 & 19.2 \\
\hline
\multicolumn{8}{p{0.97\textwidth}}{$^{\dagger}$Fewer than five positive cases; PR-AUC is shown for completeness but is not statistically stable and should be interpreted as indicative only.} \\
\end{tabular}}
\end{table}

\FloatBarrier
\clearpage
\begin{table}[H]
\centering

\caption{{\textbf{Pre-training and post-training comparisons on retrospective MarketScan.} Mean disease-level ROC-AUCs are reported for the same 1,208 endpoints before and after disease-prediction post-training. Each checkpoint pair uses identical endpoint-specific case--control counts and sex-by-age strata. GPT-2 and LLaMA-3.1 were trained from scratch using the same MarketScan corpus, tokenizer and pre-training objective as ReClaim, then post-trained on the same 100,000 prompt--response pairs. Gains are post-training minus pre-training in percentage points (pp), calculated from unrounded values. Differences in parameter count mean that cross-model comparisons do not isolate architecture from model capacity. EHRShot results, including ETHOS, are reported separately in Supplementary Table~\ref{tab:ethos_ehrshot}.}}
\label{tab:add_baselines}
\small
\begin{tabular}{lcccc}
\toprule
\textbf{Model} & \textbf{Parameters} & \textbf{Pre-training} & \textbf{Post-training} & \textbf{Gain (pp)} \\
\midrule
GPT-2 & 60M & 0.670 & 0.741 & +7.11 \\
LLaMA-3.1 & 120M & 0.616 & 0.744 & +12.80 \\
ReClaim-S & 140M & 0.616 & 0.743 & +12.69 \\
ReClaim-L & 1.7B & 0.622 & 0.769 & +14.65 \\
\bottomrule
\end{tabular}

\end{table}

\begin{table}[H]
\centering

\caption{{\textbf{Effect of rare-disease sampling strategies on disease prediction performance.} Mean ROC-AUC on the retrospective MarketScan evaluation is reported for three post-training settings, each containing 100,000 prompt--response pairs. Resampling constructs a new sample with rare-disease examples accounting for 20\% or 50\% of the pairs. Rare-example repetition adds one copy of each rare-disease example in the existing post-training sample, then randomly removes common-disease examples to keep the total number of pairs unchanged. ROC-AUC is averaged across evaluable sex-by-age strata within each endpoint and then equally across endpoints. The rare subset comprises 437 endpoints with more than 10 and fewer than 500 cases across the retrospective and prospective evaluations combined, following the benchmark definition in Methods. The remaining 771 endpoints are reported separately.}}
\label{tab:posttrain_ablations}
\begin{tabular}{lccc}
\toprule
Post-training setting &
\makecell{All endpoints\\($n=1{,}208$)} &
\makecell{Rare endpoints\\($n=437$)} &
\makecell{Other endpoints\\($n=771$)} \\
\midrule
Resampling (20\% rare) & 0.772 & 0.781 & 0.766 \\
Resampling (50\% rare) & 0.762 & 0.764 & 0.761 \\
Rare-example repetition ($2\times$) & 0.760 & 0.768 & 0.755 \\
\bottomrule
\end{tabular}
\end{table}

\FloatBarrier
\clearpage
\begin{table}[H]
\centering
\caption{{Complete benchmark results for next-year healthcare expenditure forecasting. Held-out cohort, retrospective (n = 356,722; HNHC prevalence 4.4\%) and prospective (n = 64,658; 7.6\%) evaluations. Regression: total next-year expenditure censored at \$1,000,000 (R$^2$, MAE and RMSE in dollars). HNHC identification: next-year total $\geq$ \$30,000; ROC-AUC and PR-AUC are threshold-free (computed from each model's HNHC probability: the classifier's output, the post-trained model's bin mass at or above \$30,000, or the fraction of Monte Carlo simulations at or above \$30,000); F1, precision and recall are at one fixed operating point per model family: predicted dollars $\geq$ \$30,000 for every dollar-emitting model (persistence, LightGBM regressor, ReClaim pre-train and post-train) and the argmax class of the per-cutoff probabilities for the LightGBM classifier. A classifier emits no dollars (no regression metrics); the train-mean anchor never flags HNHC. ReClaim pre-train = 64-trajectory Monte Carlo simulation of the pre-trained backbone; post-train = the post-trained model (100,000 pairs; one response token, the next-year total). LightGBM trains one model per target.}}
\label{tab:expenditure_full_results}   %
\scriptsize
\renewcommand{\arraystretch}{1.12}
\begin{tabular}{@{}lrrrrrrrr@{}}
\hline
 & \multicolumn{3}{c}{\textbf{Regression (total expenditure)}} & \multicolumn{5}{c}{\textbf{HNHC identification ($\geq$ \$30,000)}} \\
\cmidrule(lr){2-4}\cmidrule(lr){5-9}
\textbf{Model} & \textbf{R$^2$} & \textbf{MAE} & \textbf{RMSE} & \textbf{ROC-AUC} & \textbf{PR-AUC} & \textbf{F1} & \textbf{Precision} & \textbf{Recall} \\
\hline
\multicolumn{9}{@{}l}{\textit{Retrospective evaluation}} \\
Train-mean anchor & 0.000 & 8,201 & 23,756 & -- & -- & -- & -- & -- \\
Persistence (last-year cost) & 0.117 & 5,862 & 22,329 & -- & -- & 0.397 & 0.431 & 0.368 \\
LightGBM regressor, 1-yr window & 0.286 & 5,918 & 20,072 & -- & -- & 0.409 & 0.559 & 0.322 \\
LightGBM regressor, all history & 0.244 & 6,188 & 20,657 & -- & -- & 0.379 & 0.523 & 0.297 \\
LightGBM classifier, 1-yr window & -- & -- & -- & 0.868 & 0.444 & 0.367 & 0.745 & 0.243 \\
LightGBM classifier, all history & -- & -- & -- & 0.868 & 0.409 & 0.324 & 0.710 & 0.210 \\
ReClaim-S, pre-train (MC) & 0.375 & 4,940 & 18,777 & 0.874 & 0.507 & 0.467 & 0.667 & 0.360 \\
ReClaim-M, pre-train (MC) & 0.383 & 4,902 & 18,668 & 0.875 & 0.512 & 0.470 & 0.672 & 0.361 \\
ReClaim-L, pre-train (MC) & 0.382 & 4,897 & 18,671 & 0.877 & 0.515 & 0.472 & 0.674 & 0.363 \\
ReClaim-S, post-train & 0.371 & 5,382 & 18,844 & 0.889 & 0.515 & 0.495 & 0.589 & 0.428 \\
ReClaim-M, post-train & 0.388 & 5,574 & 18,588 & 0.891 & 0.523 & 0.501 & 0.554 & 0.458 \\
\textbf{ReClaim-L, post-train} & 0.393 & 5,301 & 18,513 & 0.894 & 0.524 & 0.502 & 0.580 & 0.443 \\
\hline
\multicolumn{9}{@{}l}{\textit{Prospective evaluation}} \\
Train-mean anchor & -0.015 & 10,623 & 32,231 & -- & -- & -- & -- & -- \\
Persistence (last-year cost) & 0.221 & 8,253 & 28,241 & -- & -- & 0.459 & 0.517 & 0.412 \\
LightGBM regressor, 1-yr window & 0.323 & 8,030 & 26,320 & -- & -- & 0.456 & 0.637 & 0.355 \\
LightGBM regressor, all history & 0.241 & 9,594 & 27,883 & -- & -- & 0.435 & 0.468 & 0.407 \\
LightGBM classifier, 1-yr window & -- & -- & -- & 0.853 & 0.535 & 0.432 & 0.806 & 0.295 \\
LightGBM classifier, all history & -- & -- & -- & 0.845 & 0.469 & 0.421 & 0.672 & 0.307 \\
ReClaim-S, pre-train (MC) & 0.439 & 7,558 & 23,957 & 0.863 & 0.596 & 0.538 & 0.723 & 0.428 \\
ReClaim-M, pre-train (MC) & 0.453 & 7,265 & 23,663 & 0.875 & 0.607 & 0.540 & 0.722 & 0.432 \\
ReClaim-L, pre-train (MC) & 0.452 & 7,420 & 23,688 & 0.874 & 0.606 & 0.542 & 0.718 & 0.435 \\
ReClaim-S, post-train & 0.425 & 7,076 & 24,254 & 0.874 & 0.597 & 0.556 & 0.697 & 0.463 \\
ReClaim-M, post-train & 0.443 & 7,728 & 23,892 & 0.874 & 0.602 & 0.561 & 0.607 & 0.520 \\
\textbf{ReClaim-L, post-train} & 0.446 & 7,491 & 23,811 & 0.880 & 0.606 & 0.563 & 0.630 & 0.509 \\
\hline
\end{tabular}
\end{table}

\FloatBarrier
\clearpage
\begin{table}[H]
\centering
\caption{{Sensitivity analysis: next-year total expenditure forecasting stratified by payer. Held-out cohort split by the enrollee's payer database: CCAE+MDCR (commercial and Medicare Supplemental; retrospective n = 305,211, prospective n = 59,697) and MDCD (Medicaid; n = 51,511 and 4,961). Regression on the censored total (R$^2$, MAE in dollars); HNHC identification (next-year total $\geq$ \$30,000; ROC-AUC, PR-AUC, threshold-free). Model definitions as in Supplementary Table~\ref{tab:expenditure_full_results}.}}
\label{tab:expenditure_by_payer}   %
\scriptsize
\renewcommand{\arraystretch}{1.12}
\begin{tabular}{@{}lrrrrrrrr@{}}
\hline
 & \multicolumn{4}{c}{\textbf{CCAE+MDCR}} & \multicolumn{4}{c}{\textbf{MDCD}} \\
\cmidrule(lr){2-5}\cmidrule(lr){6-9}
\textbf{Model} & \textbf{R$^2$} & \textbf{MAE} & \textbf{ROC-AUC} & \textbf{PR-AUC} & \textbf{R$^2$} & \textbf{MAE} & \textbf{ROC-AUC} & \textbf{PR-AUC} \\
\hline
\multicolumn{9}{@{}l}{\textit{Retrospective evaluation}} \\
Train-mean anchor & 0.000 & 8,284 & -- & -- & -0.017 & 7,707 & -- & -- \\
Persistence (last-year cost) & 0.089 & 6,344 & -- & -- & 0.380 & 3,005 & -- & -- \\
LightGBM regressor, 1-yr window & 0.274 & 6,252 & -- & -- & 0.392 & 3,936 & -- & -- \\
LightGBM regressor, all history & 0.236 & 6,507 & -- & -- & 0.308 & 4,300 & -- & -- \\
LightGBM classifier, 1-yr window & -- & -- & 0.854 & 0.421 & -- & -- & 0.938 & 0.631 \\
LightGBM classifier, all history & -- & -- & 0.853 & 0.389 & -- & -- & 0.942 & 0.575 \\
ReClaim-S, pre-train (MC) & 0.367 & 5,320 & 0.863 & 0.487 & 0.451 & 2,693 & 0.935 & 0.682 \\
ReClaim-M, pre-train (MC) & 0.374 & 5,280 & 0.863 & 0.492 & 0.457 & 2,663 & 0.938 & 0.689 \\
ReClaim-L, pre-train (MC) & 0.374 & 5,274 & 0.865 & 0.495 & 0.455 & 2,668 & 0.941 & 0.690 \\
ReClaim-S, post-train & 0.362 & 5,803 & 0.876 & 0.493 & 0.449 & 2,882 & 0.953 & 0.701 \\
ReClaim-M, post-train & 0.378 & 6,043 & 0.878 & 0.501 & 0.478 & 2,797 & 0.953 & 0.705 \\
\textbf{ReClaim-L, post-train} & 0.383 & 5,733 & 0.882 & 0.502 & 0.482 & 2,738 & 0.954 & 0.708 \\
\hline
\multicolumn{9}{@{}l}{\textit{Prospective evaluation}} \\
Train-mean anchor & -0.016 & 10,692 & -- & -- & -0.005 & 9,801 & -- & -- \\
Persistence (last-year cost) & 0.205 & 8,501 & -- & -- & 0.448 & 5,275 & -- & -- \\
LightGBM regressor, 1-yr window & 0.319 & 8,195 & -- & -- & 0.389 & 6,051 & -- & -- \\
LightGBM regressor, all history & 0.243 & 9,735 & -- & -- & 0.202 & 7,893 & -- & -- \\
LightGBM classifier, 1-yr window & -- & -- & 0.848 & 0.530 & -- & -- & 0.904 & 0.609 \\
LightGBM classifier, all history & -- & -- & 0.841 & 0.469 & -- & -- & 0.894 & 0.499 \\
ReClaim-S, pre-train (MC) & 0.435 & 7,683 & 0.859 & 0.590 & 0.506 & 6,048 & 0.912 & 0.680 \\
ReClaim-M, pre-train (MC) & 0.448 & 7,416 & 0.871 & 0.600 & 0.517 & 5,438 & 0.920 & 0.696 \\
ReClaim-L, pre-train (MC) & 0.447 & 7,579 & 0.869 & 0.599 & 0.526 & 5,508 & 0.922 & 0.700 \\
ReClaim-S, post-train & 0.420 & 7,242 & 0.870 & 0.589 & 0.494 & 5,071 & 0.917 & 0.698 \\
ReClaim-M, post-train & 0.438 & 7,938 & 0.870 & 0.595 & 0.510 & 5,202 & 0.913 & 0.699 \\
\textbf{ReClaim-L, post-train} & 0.442 & 7,675 & 0.876 & 0.599 & 0.503 & 5,267 & 0.917 & 0.708 \\
\hline
\end{tabular}
\end{table}

\FloatBarrier
\clearpage
\begin{table}[H]
\centering
\caption{{Inference cost of pre-trained versus post-trained ReClaim on the expenditure forecasting task. GPU-hours to produce annual total-expenditure forecasts for the full held-out expenditure cohort (421,380 forecasts: 356,722 retrospective and 64,658 prospective, for 372,310 unique enrollees), recorded by the evaluation pipeline for the runs reported in this paper on identical hardware and software: 8 $\times$ NVIDIA B200 (data-parallel), the same vLLM engine configuration for both inference methods, one job at a time, wall-clock time $\times$ GPUs including engine start-up. Pre-train is the Monte Carlo simulation of the pre-trained model, which runs 64 Monte Carlo simulations of the following year per enrollee (mean 262 generated tokens each) and aggregates their costs; post-train is the expenditure-post-trained model, which decodes one cost token greedily. Mean generated tokens per enrollee = total tokens generated by the engine divided by the number of enrollees, over the full cohort (all 64 simulations counted for the pre-trained model; exactly one token for every enrollee under the post-trained model). Ratio = pre-train divided by post-train within each block.}}
\label{tab:inference_cost}
\small
\renewcommand{\arraystretch}{1.15}
\begin{tabular}{@{}lrrr@{}}
\hline
\textbf{Backbone} & \textbf{Pre-train} & \textbf{Post-train} & \textbf{Ratio} \\
\hline
\multicolumn{4}{@{}l}{\textit{Mean generated tokens per enrollee}} \\
ReClaim-S & 16,770 & 1 & 16,770$\times$ \\
ReClaim-M & 16,770 & 1 & 16,770$\times$ \\
ReClaim-L & 16,799 & 1 & 16,799$\times$ \\
\hline
\multicolumn{4}{@{}l}{\textit{GPU-hours per evaluation}} \\
ReClaim-S & 36.61 & 0.44 & 83$\times$ \\
ReClaim-M & 53.71 & 0.51 & 106$\times$ \\
ReClaim-L & 93.44 & 0.71 & 132$\times$ \\
\hline
\end{tabular}
\end{table}

\FloatBarrier
\clearpage
\begin{table}[H]
\centering
\caption{{\textbf{Effect of cost-token discretization on expenditure labels.} For every enrollee with at least one claim in the retrospective and prospective evaluation cohorts (359,187 of 372,310 unique enrollees), exact annual expenditure was reconstructed from the raw MarketScan inpatient, outpatient and pharmacy claims and compared with the tokenized total, in which each monthly sum per claim type is rounded to one significant digit before summation, overall and by payer. R$^2$: coefficient of determination of the tokenized annual total against the exact total. HNHC: high-need, high-cost, annual total $\geq$ \$30{,}000; the label is changed by tokenization when it differs between the exact and the tokenized total. Enrollee-years above \$1{,}000{,}000, the censoring point of the expenditure targets, are counted before and after tokenization.}}
\label{tab:discretization}
\small
\renewcommand{\arraystretch}{1.15}
\begin{tabular}{@{}lrrr@{}}
\hline
 & \textbf{All} & \textbf{CCAE + MDCR} & \textbf{MDCD} \\
\hline
Enrollees & 359,187 & 313,491 & 45,696 \\
Enrollee-years & 2,084,595 & 1,834,235 & 250,360 \\
Enrollee-years above \$1{,}000{,}000, exact & 150 (0.007\%) & 130 (0.007\%) & 20 (0.008\%) \\
Enrollee-years above \$1{,}000{,}000, tokenized & 147 (0.007\%) & 129 (0.007\%) & 18 (0.007\%) \\
R$^2$, tokenized vs exact & 0.993 & 0.993 & 0.989 \\
HNHC prevalence, exact & 5.24\% & 5.35\% & 4.44\% \\
HNHC prevalence, tokenized & 5.21\% & 5.32\% & 4.43\% \\
HNHC label changed by tokenization & 0.42\% & 0.44\% & 0.27\% \\
\hline
\end{tabular}
\end{table}

\FloatBarrier
\clearpage
\begin{table}[H]
\centering
\caption{{Performance and cost of the pre-trained model's Monte Carlo simulation by number of Monte Carlo simulations. Pre-trained ReClaim-S, -M and -L read out by $N$ simulations of the following year per enrollee, in the retrospective (n = 356,722) and prospective (n = 64,658) evaluations: R$^2$ of the total, and HNHC ROC-AUC and PR-AUC with the fraction of simulations at or above \$30,000 as the score. GPU-hours are for one pass over the retrospective evaluation on 8 $\times$ NVIDIA B200 with vLLM as the inference engine. $N$ = 1 is greedy decoding of a single simulation. The last row of each block is the post-trained one-token prediction of the same backbone (Supplementary Table~\ref{tab:expenditure_full_results}).}}
\label{tab:mc_trajectory_ablation}
\scriptsize
\renewcommand{\arraystretch}{1.12}
\begin{tabular}{@{}lrrrrrrr@{}}
\hline
 & \multicolumn{3}{c}{\textbf{Retrospective}} & \multicolumn{3}{c}{\textbf{Prospective}} & \\
\cmidrule(lr){2-4}\cmidrule(lr){5-7}
\textbf{$N$} & \textbf{R$^2$} & \textbf{ROC-AUC} & \textbf{PR-AUC} & \textbf{R$^2$} & \textbf{ROC-AUC} & \textbf{PR-AUC} & \textbf{GPU-hours} \\
\hline
\multicolumn{8}{@{}l}{\textit{ReClaim-S}} \\
1 & 0.236 & 0.609 & 0.204 & 0.317 & 0.648 & 0.300 & -- \\
2 & 0.188 & 0.696 & 0.240 & 0.211 & 0.724 & 0.329 & 1.2 \\
4 & 0.272 & 0.747 & 0.329 & 0.338 & 0.768 & 0.422 & 2.2 \\
8 & 0.330 & 0.788 & 0.397 & 0.392 & 0.805 & 0.490 & 4.2 \\
16 & 0.359 & 0.826 & 0.449 & 0.421 & 0.831 & 0.541 & 8.3 \\
32 & 0.370 & 0.855 & 0.486 & 0.434 & 0.855 & 0.576 & 16.3 \\
64 & 0.375 & 0.874 & 0.507 & 0.439 & 0.863 & 0.596 & 32.3 \\
post-trained inference & 0.371 & 0.889 & 0.515 & 0.425 & 0.874 & 0.597 & 0.38 \\
\hline
\multicolumn{8}{@{}l}{\textit{ReClaim-M}} \\
1 & 0.237 & 0.613 & 0.212 & 0.314 & 0.649 & 0.303 & -- \\
2 & 0.189 & 0.699 & 0.248 & 0.247 & 0.729 & 0.337 & 1.9 \\
4 & 0.296 & 0.749 & 0.337 & 0.349 & 0.771 & 0.429 & 3.3 \\
8 & 0.339 & 0.791 & 0.404 & 0.407 & 0.810 & 0.499 & 6.4 \\
16 & 0.361 & 0.828 & 0.454 & 0.426 & 0.840 & 0.550 & 12.3 \\
32 & 0.377 & 0.856 & 0.492 & 0.445 & 0.858 & 0.582 & 24.2 \\
64 & 0.383 & 0.875 & 0.512 & 0.453 & 0.875 & 0.607 & 47.4 \\
post-trained inference & 0.388 & 0.891 & 0.523 & 0.443 & 0.874 & 0.602 & 0.43 \\
\hline
\multicolumn{8}{@{}l}{\textit{ReClaim-L}} \\
1 & 0.239 & 0.613 & 0.213 & 0.313 & 0.649 & 0.300 & -- \\
2 & 0.183 & 0.700 & 0.248 & 0.223 & 0.730 & 0.338 & 3.2 \\
4 & 0.289 & 0.750 & 0.341 & 0.352 & 0.765 & 0.426 & 5.7 \\
8 & 0.340 & 0.792 & 0.409 & 0.404 & 0.810 & 0.501 & 11.0 \\
16 & 0.364 & 0.828 & 0.457 & 0.440 & 0.838 & 0.548 & 20.9 \\
32 & 0.377 & 0.857 & 0.493 & 0.447 & 0.859 & 0.584 & 40.7 \\
64 & 0.382 & 0.877 & 0.515 & 0.452 & 0.874 & 0.606 & 82.1 \\
post-trained inference & 0.393 & 0.894 & 0.524 & 0.446 & 0.880 & 0.606 & 0.60 \\
\hline
\end{tabular}
\end{table}

\FloatBarrier
\clearpage
\begin{table}[H]
\centering
\caption{{Baseline characteristics of the positive-control and negative-control study
populations.
Cohort composition and outcome counts for each control, in the reference study and in the
MarketScan MDCR cohort. The positive control compares initiators of a COX-2 inhibitor (celecoxib in our study period) with initiators of
nonselective NSAIDs for severe gastrointestinal complications. The negative control compares influenza
vaccination with no vaccination for hip fracture, with follow-up over the four months after the
exposure season in both designs (January--April 1997 in the reference, January--April 2012 in ours).
COX-2, cyclooxygenase-2; GI, gastrointestinal; NR, not reported; NSAID, nonsteroidal anti-inflammatory drug; PACE, Pennsylvania Pharmaceutical Assistance Contract for the Elderly, linked to Medicare; PUD, peptic ulcer disease;
SD, standard deviation.}}
\label{tab:baseline_controls}
\footnotesize
\setlength{\tabcolsep}{4pt}
\renewcommand{\arraystretch}{1.05}
\begin{tabular}{@{}l r r@{}}
\toprule
\textbf{Characteristic}
 & \textbf{\shortstack{PACE \\ (reference)}}
 & \textbf{\shortstack{MarketScan \\ MDCR}} \\
\midrule
\multicolumn{3}{c}{\textbf{Positive control}} \\
\midrule
Index period                         & 1999--2002    & 2011--2014 \\
Patients, $n$                        & 49,653        & 977,551 \\
COX-2 inhibitor initiators, $n$ (\%) & 32,042 (64.5) & 84,773 (8.7) \\
Age at index, years                  & NR            & 72.9 (7.3) \\
Female sex, \%                       & 84.2          & 57.7 \\
Gastroprotective drug use, \%        & 24.9          & 16.5 \\
GI hemorrhage, \%                    & 1.5           & 1.3 \\
GI complication events, $n$ (\%)     & 552 (1.1)     & 13,152 (1.3) \\
Events in exposed arm, $n$ (\%)      & 367 (1.1)     & 1,222 (1.4) \\
\midrule
\multicolumn{3}{c}{\textbf{Negative control}} \\
\midrule
Index period                         & Oct--Dec 1996 & Oct--Dec 2011 \\
Patients, $n$                        & 147,583       & 3,040,767 \\
Vaccinated, $n$ (\%)                 & NR (42.0)     & 463,216 (15.2) \\
Age at index, years                  & NR            & 74.9 (7.5) \\
Female sex, \%                       & NR            & 52.9 \\
Osteoporosis, \%                     & NR            & 2.9 \\
Glucocorticoid use, \%               & NR            & 6.1 \\
Hip fracture events, $n$ (\%)        & 710 (0.48)    & 8,559 (0.28) \\
Events in vaccinated arm, $n$ (\%)   & NR            & 1,173 (0.25) \\
\bottomrule
\end{tabular}
\end{table}

\FloatBarrier
\clearpage
\begin{table}[H]
\centering
\caption{{Known-effect validation results for the positive and negative controls.
Decile-adjusted risk ratios with 95\% confidence intervals for each propensity-score
specification. The positive control compares initiators of a COX-2 inhibitor with initiators of
nonselective NSAIDs for severe gastrointestinal complications, for which a moderate protective effect is expected from randomized trials~\cite{schneeweiss2009high,moore2005tolerability}. The negative control compares influenza vaccination
with no vaccination for hip fracture under a landmark design, in which the expected relative risk is 1.0, and all negative-control 95\% confidence intervals in this table include 1.0. MarketScan estimates are from all eligible MarketScan MDCR enrollees (Methods). Reference values are from Table 3 of Schneeweiss et al.~\cite{schneeweiss2009high}. $c$, c-statistic of the propensity-score model (area under the receiver operating characteristic
curve), reported beside each estimate; it is undefined for the unadjusted analysis, and the reference reports
a c-statistic only for its hdPS model in the positive control. Dashes denote cells that
are undefined by design or for which no reference estimate is reported. CI, confidence interval; hdPS, high-dimensional propensity score; NSAID,
nonsteroidal anti-inflammatory drug; PACE, Pennsylvania Pharmaceutical Assistance Contract for the
Elderly, linked to Medicare; PS, propensity score; RR, risk ratio; $c$, c-statistic.}}
\label{tab:benchmark_controls}
\footnotesize
\setlength{\tabcolsep}{4pt}
\renewcommand{\arraystretch}{1.15}
\begin{tabular}{@{}l cc cc@{}}
\toprule
\textbf{PS specification} & \multicolumn{2}{c}{\textbf{MarketScan MDCR}} & \multicolumn{2}{c}{\textbf{PACE (reference)}} \\
\cmidrule(lr){2-3}\cmidrule(lr){4-5}
 & RR (95\% CI) & $c$ & RR (95\% CI) & $c$ \\
\midrule
\multicolumn{5}{c}{\textbf{Positive control}} \\
\midrule
Patients, $n$ & \multicolumn{2}{c}{977,551} & \multicolumn{2}{c}{49,653} \\
Unadjusted & 1.080 (1.018--1.146) & --- & 1.09 (0.91--1.30) & --- \\
hdPS       & 1.017 (0.957--1.081) & 0.690 & 0.87 (0.72--1.05) & 0.71 \\
Delphi     & 1.009 (0.951--1.072) & 0.655 & --- & --- \\
ReClaim-S  & 0.924 (0.870--0.983) & 0.744 & --- & --- \\
ReClaim-M  & 0.913 (0.859--0.972) & 0.755 & --- & --- \\
ReClaim-L  & 0.911 (0.857--0.969) & 0.751 & --- & --- \\
\midrule
\multicolumn{5}{c}{\textbf{Negative control}} \\
\midrule
Patients, $n$ & \multicolumn{2}{c}{3,040,767} & \multicolumn{2}{c}{147,583} \\
Unadjusted & 0.991 (0.932--1.054) & --- & 0.93 (0.80--1.08) & --- \\
hdPS       & 1.008 (0.947--1.074) & 0.665 & 1.02 (0.85--1.21) & --- \\
Delphi     & 1.011 (0.950--1.076) & 0.621 & --- & --- \\
ReClaim-S  & 1.021 (0.956--1.091) & 0.798 & --- & --- \\
ReClaim-M  & 1.033 (0.966--1.104) & 0.810 & --- & --- \\
ReClaim-L  & 1.009 (0.944--1.079) & 0.803 & --- & --- \\
\bottomrule
\end{tabular}
\end{table}

\FloatBarrier
\clearpage
\begin{table}[H]
\centering
\caption{{Dimension-matched principal component compressions of the ReClaim
representations.
Decile-adjusted risk ratios with 95\% confidence intervals in the positive- and
negative-control validation cohorts, for each ReClaim backbone compressed to $d$ principal
components and run through the unchanged analysis ladder. Each value of $d$ matches the dimension of 
a named baseline specification, whose own estimate is given for comparison. 
Cumulative variance explained by the compression is 82--85\% at $d$=120 and 93--96\% at $d$=500.
For the positive control, the unadjusted estimate is 1.080 and randomized trials establish a moderate protective effect; for the negative control, the known null is exactly 1.0 and the unadjusted
estimate is 0.991. CI, confidence interval; hdPS, high-dimensional
propensity score; PCA, principal component analysis; RR, risk ratio.}}
\label{tab:t10_pca_compressions}
\footnotesize
\setlength{\tabcolsep}{4pt}
\renewcommand{\arraystretch}{1.15}
\begin{tabular}{@{}c l c c c@{}}
\toprule
\textbf{$d$} & \textbf{\shortstack[l]{Dimension-matched \\ baseline}}
 & \textbf{\shortstack{PCA \\ ReClaim-S}} & \textbf{\shortstack{PCA \\ ReClaim-M}} & \textbf{\shortstack{PCA \\ ReClaim-L}} \\
\midrule
\multicolumn{5}{c}{\textbf{Positive control}} \\
\midrule
120  & Delphi: 1.009 (0.951--1.072) & 0.930 (0.875--0.988) & 0.933 (0.878--0.991) & 0.932 (0.878--0.991) \\
500  & hdPS: 1.017 (0.957--1.081)   & 0.931 (0.876--0.990) & 0.930 (0.875--0.989) & 0.927 (0.872--0.986) \\
Full & Native width                 & 0.924 (0.870--0.983) & 0.913 (0.859--0.972) & 0.911 (0.857--0.969) \\
\midrule
\multicolumn{5}{c}{\textbf{Negative control}} \\
\midrule
120  & Delphi: 1.011 (0.950--1.076) & 1.027 (0.961--1.097) & 1.036 (0.971--1.107) & 1.058 (0.991--1.129) \\
500  & hdPS: 1.008 (0.947--1.074)   & 1.024 (0.958--1.094) & 1.021 (0.955--1.091) & 1.019 (0.954--1.089) \\
Full & Native width                 & 1.021 (0.956--1.091) & 1.033 (0.966--1.104) & 1.009 (0.944--1.079) \\
\bottomrule
\end{tabular}
\end{table}

\FloatBarrier
\clearpage
\begin{table}[H]
\centering
\caption{{Doubly robust analysis of the known-effect validation.
Risk ratios with 95\% confidence intervals for the effect in the treated, estimated by cross-fitted
augmented inverse probability weighting with three propensity models, each alone (PS only) or paired with
one of four outcome models (OM), as in Supplementary Fig.~\ref{fig:doubly_robust}. PS only is the same weighting
without an outcome model, using the propensity score alone. Comparator weights are truncated
at their 1st and 99th percentiles. Positive control: all MarketScan MDCR enrollees ($n$ = 977{,}551;
84{,}773 treated; unadjusted risk ratio 1.080). Negative control: enrollees still enrolled when the
outcome window opens ($n$ = 2{,}513{,}894; 346{,}786 treated; unadjusted risk ratio 0.992; the known
null is exactly 1.0). hdPS, high-dimensional propensity score; OM, outcome model; PS, propensity score.}}
\label{tab:doubly_robust}
\footnotesize
\setlength{\tabcolsep}{5pt}
\renewcommand{\arraystretch}{1.15}
\begin{tabular}{@{}l l c c@{}}
\toprule
\textbf{Propensity model} & \textbf{Outcome model} & \textbf{Positive control} & \textbf{Negative control} \\
 & & RR (95\% CI) & RR (95\% CI) \\
\midrule
hdPS      & PS only         & 1.015 (0.956--1.078) & 1.000 (0.939--1.065) \\
          & + Delphi OM     & 0.991 (0.934--1.051) & 0.990 (0.930--1.054) \\
          & + ReClaim-S OM  & 0.937 (0.883--0.994) & 1.001 (0.940--1.065) \\
          & + ReClaim-M OM  & 0.937 (0.883--0.993) & 1.005 (0.944--1.070) \\
          & + ReClaim-L OM  & 0.935 (0.882--0.992) & 0.992 (0.932--1.057) \\
\midrule
Delphi    & PS only         & 1.013 (0.954--1.075) & 0.994 (0.934--1.058) \\
          & + Delphi OM     & 1.016 (0.958--1.078) & 1.003 (0.943--1.067) \\
          & + ReClaim-S OM  & 0.938 (0.885--0.994) & 1.012 (0.951--1.077) \\
          & + ReClaim-M OM  & 0.939 (0.885--0.995) & 1.012 (0.952--1.077) \\
          & + ReClaim-L OM  & 0.939 (0.886--0.996) & 0.997 (0.937--1.061) \\
\midrule
ReClaim-S & PS only         & 0.926 (0.871--0.984) & 0.982 (0.918--1.050) \\
          & + Delphi OM     & 0.932 (0.878--0.990) & 0.986 (0.922--1.054) \\
          & + ReClaim-S OM  & 0.926 (0.872--0.983) & 0.994 (0.930--1.063) \\
          & + ReClaim-M OM  & 0.923 (0.869--0.980) & 0.992 (0.928--1.061) \\
          & + ReClaim-L OM  & 0.923 (0.869--0.980) & 0.988 (0.924--1.056) \\
\bottomrule
\end{tabular}
\end{table}

\FloatBarrier
\clearpage
\begin{table}[H]
\centering
\caption{{Baseline characteristics and cumulative incidence in the Penn Medicine, MarketScan
and Yale New Haven Health cohorts, by emulation.
Cohort composition and unadjusted cumulative incidence over follow-up for the GLP-1RA arm and the
comparator arm of each emulation, in three data sources. The index period is 2019--2024 for all
twelve arms. The GLP-1RA columns of the two emulations are separate cohorts, not the same patients
analyzed twice. The reference study originally reports incidence as rates per 1,000 person-years.
We convert them to cumulative-incidence percentages for direct comparison.
Penn Medicine values are as reported by Tang et al.~\cite{tang2025association}.
DPP4i, dipeptidyl peptidase-4 inhibitor; GLP-1RA, glucagon-like peptide-1 receptor agonist; MS,
MarketScan 2-million combined held-out set; PM, Penn Medicine electronic health records; SD,
standard deviation; SGLT2i, sodium-glucose cotransporter-2 inhibitor; YNHH, Yale New Haven Health
electronic health records.}}
\label{tab:emulation_cohorts}
\footnotesize
\setlength{\tabcolsep}{3pt}
\renewcommand{\arraystretch}{1.1}
\begin{tabular}{@{}l r r r r r r@{}}
\toprule
& \multicolumn{2}{c}{\textbf{PM}} & \multicolumn{2}{c}{\textbf{MS}}
& \multicolumn{2}{c@{}}{\textbf{YNHH}} \\
\cmidrule(lr){2-3} \cmidrule(lr){4-5} \cmidrule(l){6-7}
\textbf{Characteristic}
 & \textbf{GLP-1RA} & \textbf{Comparator} & \textbf{GLP-1RA} & \textbf{Comparator}
 & \textbf{GLP-1RA} & \textbf{Comparator} \\
\midrule
\multicolumn{7}{c}{\textbf{Comparator: SGLT2i}} \\
\midrule
$n$                  & 7,308       & 10,119      & 3,888       & 2,778       & 7,532       & 8,763 \\
Age at index         & 58.3 (12.7) & 64.7 (12.0) & 52.8 (11.3) & 56.6 (11.8) & 57.6 (13.6) & 64.5 (12.9) \\
Female               & 49.3\%      & 33.6\%      & 49.0\%      & 34.2\%      & 50.9\%      & 32.2\% \\
Anxiety              & 10.9\%      & 9.6\%       & 9.75\%      & 8.42\%      & 9.56\%      & 8.36\% \\
Depression           & 6.7\%       & 6.7\%       & 5.32\%      & 4.50\%      & 4.83\%      & 4.28\% \\
Bipolar disorder     & 0.59\%      & 0.38\%      & 0.23\%      & 0.32\%      & 0.05\%      & 0.11\% \\
Schizophrenia        & 0.34\%      & 0.23\%      & 0.10\%      & 0.18\%      & 0.32\%      & 0.29\% \\
Other mood disorders & 0.64\%      & 0.68\%      & 0.54\%      & 0.36\%      & 0.56\%      & 0.54\% \\
Suicidal ideation    & 0.14\%      & 0.13\%      & 0.08\%      & 0.11\%      & 0.24\%      & 0.40\% \\
\midrule
\multicolumn{7}{c}{\textbf{Comparator: DPP4i}} \\
\midrule
$n$                  & 7,300       & 7,394       & 3,894       & 1,540       & 7,531       & 7,046 \\
Age at index         & 58.2 (12.7) & 67.0 (13.1) & 52.8 (11.3) & 58.6 (13.3) & 57.6 (13.6) & 65.3 (13.7) \\
Female               & 49.3\%      & 43.6\%      & 48.9\%      & 40.4\%      & 50.6\%      & 41.6\% \\
Anxiety              & 12.8\%      & 11.1\%      & 9.71\%      & 10.71\%     & 9.45\%      & 11.61\% \\
Depression           & 7.8\%       & 8.1\%       & 5.32\%      & 6.49\%      & 4.77\%      & 6.88\% \\
Bipolar disorder     & 0.55\%      & 0.60\%      & 0.23\%      & 0.71\%      & 0.05\%      & 0.20\% \\
Schizophrenia        & 0.38\%      & 0.54\%      & 0.10\%      & 0.65\%      & 0.35\%      & 0.67\% \\
Other mood disorders & 0.66\%      & 0.78\%      & 0.54\%      & 0.71\%      & 0.53\%      & 0.70\% \\
Suicidal ideation    & 0.16\%      & 0.26\%      & 0.08\%      & 0.32\%      & 0.24\%      & 0.48\% \\
\bottomrule
\end{tabular}
\end{table}

\FloatBarrier
\clearpage
\begin{table}[H]
\centering
\caption{{The reference study's covariate list and its implementation on each of our data
sources.
For each covariate in the reference specification (Supplementary Table 5 of Tang et al.~\cite{tang2025association}), we report whether it could be constructed in each data source and how it was implemented. The rule is to drop rather than substitute: any
covariate that cannot be built the same way on both sources (MarketScan and Yale New Haven Health) is omitted, so the specification is identical across sources apart from the insurance factor, which has three levels on claims and four
on the electronic health record. ACEi, angiotensin-converting enzyme inhibitor; ARB, angiotensin
receptor blocker; BMI, body mass index; COPD, chronic obstructive pulmonary disease; eGFR,
estimated glomerular filtration rate; EHR, electronic health record; HbA1c, glycated hemoglobin;
ICD-10-CM, International Classification of Diseases, Tenth Revision, Clinical Modification; MS, MarketScan 2-million combined
held-out set; NSAID, nonsteroidal anti-inflammatory drug; OMOP, Observational Medical Outcomes
Partnership; YNHH, Yale New Haven Health electronic health records.}}
\label{tab:rwe_application_reference_spec}
\scriptsize
\setlength{\tabcolsep}{5pt}
\renewcommand{\arraystretch}{1.2}
\begin{tabular}{@{}>{\raggedright\arraybackslash}p{1.8cm} >{\raggedright\arraybackslash}p{5.4cm} c c >{\raggedright\arraybackslash}p{5.4cm}@{}}
\toprule
\textbf{Group} & \textbf{Reference covariate} & \textbf{MS} & \textbf{YNHH} & \textbf{Implementation note} \\
\midrule
\multirow{3}{=}{Demographics}
 & Age at index & Yes & Yes & Continuous, as in the reference \\
\addlinespace[3pt]
 & Sex & Yes & Yes & \\
\addlinespace[3pt]
 & Race or ethnicity & No & No & Not recorded in MarketScan; dropped on both sources to keep the specification identical \\
\midrule
Insurance & Health insurance & Yes & Yes & Claims: payer database (CCAE, MDCR, MDCD). EHR: the OMOP payer plan period at index, mapped to commercial, Medicare, Medicaid or other/unknown \\
\midrule
Index year & Index year, 2019 to 2024 & Yes & Yes & \\
\midrule
Comorbidity index & Charlson comorbidity index & No & No & Coding algorithm and weights are unstated; every constituent condition already enters individually \\
\midrule
\multirow{3}{=}{Utilization}
 & Emergency visits ($\geq$ 1) & No & No & No native emergency-department flag in claims; dropped on both sources \\
\addlinespace[3pt]
 & Inpatient visits ($\geq$ 1) & Yes & Yes & Binary indicator \\
\addlinespace[3pt]
 & Outpatient visits (0, 1--20 or $>$ 20) & Yes & Yes & Banded as in the reference \\
\midrule
\multirow{4}{=}{Comorbidities}
 & Acute myocardial infarction, headache, menstrual disorders, arrhythmias, asthma, biliary tract disease, bronchitis, cardiomyopathy, cerebrovascular disease, chronic kidney disease, COVID-19, cognitive impairment, fractures, glaucoma, hypertension, obesity, osteoarthritis, osteoporosis, pneumonia, skin cancers, thyroid disorders, malnutrition, benign neoplasms & Yes & Yes & Same ICD-10-CM cluster dictionary on both sources, over the 365-day baseline \\
\addlinespace[3pt]
 & Peripheral vascular disease, COPD, coronary artery disease, lipid disorders, pancreatic disorders, seizures or epilepsy & Yes & Yes & Mapped one to one from a different dictionary name \\
\addlinespace[3pt]
 & Diabetes severity: nephropathy, retinopathy, neuropathy, other specified complications, other unspecified complications & Yes & Yes & Defined on the E11 code tree \\
\addlinespace[3pt]
 & Substance use disorders & Yes & Yes & Union of the substance-specific clusters, excluding tobacco \\
\midrule
Medications & ARB or ACEi, calcium-channel blockers, diuretics, lipid-lowering drugs, corticosteroids, NSAIDs, anticoagulants, antiplatelet agents, proton pump inhibitors, opioids, other anti-obesity medication, hormone replacement therapy, insulin, metformin, sulfonylureas, thiazolidinediones, alpha-glucosidase inhibitors & Yes & Yes & Same ingredient dictionary on both sources, over the 365-day baseline \\
\midrule
Laboratory and vitals & HbA1c, BMI, systolic and diastolic blood pressure, eGFR & No & No & Not available in claims; dropped on both sources to keep the specification identical \\
\bottomrule
\end{tabular}
\end{table}

\FloatBarrier
\clearpage
\begin{table}[H]
\centering
\caption{{Outcome-agnostic propensity-score diagnostics by specification: discrimination, covariate
overlap, residual systematic bias and weights.
c-statistic of the propensity-score model (area under the receiver operating characteristic curve),
covariate overlap, expected absolute systematic error (EASE) on the 24-outcome negative-control panel
of the reference study~\cite{tang2025association}, effective sample size (ESS) of each arm and maximum weight, for each specification in each comparator and data source. All estimates use stabilized inverse probability of treatment weighting, with weights truncated at their 1st and 99th percentiles. Lower EASE
indicates less residual systematic bias. Equipoise is the proportion of patients with a preference
score between 0.3 and 0.7. ESS, $(\sum w)^2/\sum w^2$ over the truncated weights, is the number of equally weighted patients carrying the same information as the weighted arm; $n$ is the number of patients in each arm before weighting. Maximum weight is the largest weight in the cohort before and after truncation. EASE, expected absolute systematic error; ESS, effective sample size; hdPS, high-dimensional propensity score; 
MS, MarketScan 2-million combined held-out set; Ref.\ spec, reference specification; trunc., truncation; 
YNHH, Yale New Haven Health electronic health records.}}

\label{tab:t5_report_card}
\footnotesize
\setlength{\tabcolsep}{4pt}
\renewcommand{\arraystretch}{1.05}
\begin{tabular}{@{}l l c c c c c c c@{}}
\toprule
& & & & & \multicolumn{2}{c}{\textbf{ESS}} & \multicolumn{2}{c@{}}{\textbf{Maximum weight}} \\
\cmidrule(lr){6-7} \cmidrule(l){8-9}
\textbf{Source} & \textbf{Specification} & \textbf{c-statistic} & \textbf{Equipoise} & \textbf{EASE} & \textbf{GLP-1RA} & \textbf{Comparator} & \makecell{\textbf{Before}\\\textbf{trunc.}} & \makecell{\textbf{After}\\\textbf{trunc.}} \\
\midrule
\multicolumn{9}{c}{\textbf{Comparator: SGLT2i}} \\
\midrule
\multirow{6}{*}{MS}
 & $n$        &       &      &       & 3{,}888 & 2{,}778 & -- & -- \\
 & Ref.\ spec & 0.702 & 0.72 & 0.165 & 3{,}493 & 2{,}343 & 8.26 & 2.58 \\
 & hdPS       & 0.755 & 0.58 & 0.130 & 3{,}309 & 2{,}106 & 64.10 & 3.24 \\
 & ReClaim-S  & 0.746 & 0.79 & 0.139 & 3{,}650 & 2{,}492 & 4.73 & 2.03 \\
 & ReClaim-M  & 0.780 & 0.76 & 0.130 & 3{,}662 & 2{,}481 & 5.22 & 1.96 \\
 & ReClaim-L  & 0.777 & 0.79 & 0.130 & 3{,}688 & 2{,}521 & 3.64 & 1.88 \\
\cmidrule(l){1-9}
\multirow{6}{*}{YNHH}
 & $n$        &       &      &       & 7{,}532 & 8{,}763 & -- & -- \\
 & Ref.\ spec & 0.735 & 0.64 & 0.034 & 5{,}949 & 7{,}386 & 26.89 & 3.32 \\
 & hdPS       & 0.767 & 0.59 & 0.050 & 5{,}956 & 6{,}987 & 86.26 & 3.48 \\
 & ReClaim-S  & 0.748 & 0.71 & 0.009 & 6{,}504 & 7{,}860 & 10.87 & 2.54 \\
 & ReClaim-M  & 0.761 & 0.73 & 0.009 & 6{,}570 & 7{,}903 & 19.26 & 2.50 \\
 & ReClaim-L  & 0.759 & 0.74 & 0.013 & 6{,}586 & 7{,}943 & 10.28 & 2.49 \\
\midrule
\multicolumn{9}{c}{\textbf{Comparator: DPP4i}} \\
\midrule
\multirow{6}{*}{MS}
 & $n$        &       &      &       & 3{,}894 & 1{,}540 & -- & -- \\
 & Ref.\ spec & 0.802 & 0.41 & 0.059 & 3{,}339 & 896 & 20.79 & 3.81 \\
 & hdPS       & 0.856 & 0.33 & 0.026 & 3{,}108 & 762 & 17.83 & 4.46 \\
 & ReClaim-S  & 0.842 & 0.48 & 0.007 & 3{,}576 & 1{,}052 & 9.48 & 2.75 \\
 & ReClaim-M  & 0.858 & 0.47 & 0.003 & 3{,}641 & 1{,}094 & 7.34 & 2.46 \\
 & ReClaim-L  & 0.858 & 0.52 & 0.001 & 3{,}664 & 1{,}122 & 6.73 & 2.50 \\
\cmidrule(l){1-9}
\multirow{6}{*}{YNHH}
 & $n$        &       &      &       & 7{,}531 & 7{,}046 & -- & -- \\
 & Ref.\ spec & 0.775 & 0.51 & 0.068 & 5{,}622 & 5{,}262 & 25.29 & 3.83 \\
 & hdPS       & 0.814 & 0.44 & 0.082 & 5{,}284 & 4{,}821 & 50.04 & 4.37 \\
 & ReClaim-S  & 0.803 & 0.54 & 0.072 & 6{,}034 & 5{,}672 & 13.62 & 3.16 \\
 & ReClaim-M  & 0.816 & 0.54 & 0.075 & 6{,}084 & 5{,}750 & 20.89 & 3.10 \\
 & ReClaim-L  & 0.820 & 0.56 & 0.064 & 6{,}185 & 5{,}827 & 16.97 & 3.00 \\
\bottomrule
\end{tabular}
\end{table}

\FloatBarrier
\clearpage
\begin{table}[H]
\centering
\caption{{Psychiatric outcomes of GLP-1RA initiation across three data sources.
Hazard ratios with 95\% confidence intervals for each outcome, by comparator, data source and
covariate specification, in new users of a GLP-1RA versus each comparator. Confidence intervals are
Wald intervals from Cox proportional hazards models fitted in the propensity-score-weighted cohort,
with robust standard errors. Treated events are the number of patients in the GLP-1RA arm who
experienced the outcome during follow-up; outcomes are ordered by the reference study's treated
event count, from most to least frequent. Ref.\ denotes the covariate specification of the
reference paper. Ref.\ calibrated and ReClaim-L calibrated denote the corresponding estimates after
negative-control calibration. Penn Medicine values are the
reference results as reported by Tang et al.~\cite{tang2025association}. MS, MarketScan 2-million
combined held-out set; PM, Penn Medicine electronic health records; YNHH, Yale New Haven Health
electronic health records.}}
\label{tab:t4_outcomes_both_sources}
\scriptsize
\setlength{\tabcolsep}{2.2pt}
\renewcommand{\arraystretch}{1.15}
\begin{tabular}{@{}l l cccccc@{}}
\toprule
\textbf{Source} & \textbf{Specification}
 & \textbf{Anxiety} & \textbf{Depression} & \textbf{Other mood}
 & \textbf{Bipolar} & \textbf{Schizophrenia+} & \textbf{\shortstack{Suicidal \\ ideation}} \\
\midrule
\multicolumn{8}{c}{\textbf{Comparator: SGLT2i}} \\
\midrule
\multirow{3}{*}{PM}
 & Treated events   & 799 & 492 & 47 & 43 & 25 & 10 \\
 & Ref.             & 1.15 (1.04--1.26) & 0.99 (0.88--1.12) & 0.93 (0.64--1.35) & 1.55 (1.00--2.41) & 1.48 (0.84--2.59) & 1.08 (0.47--2.47) \\
 & Ref.\ calibrated & 1.06 (0.84--1.35) & 0.92 (0.72--1.18) & 0.86 (0.56--1.33) & 1.44 (0.88--2.35) & 1.37 (0.75--2.49) & 1.00 (0.43--2.36) \\
\cmidrule(l){1-8}
\multirow{5}{*}{MS}
 & Treated events   & 379 & 207 & 21 & 9 & 4 & 3 \\
 & Ref.             & 1.11 (0.93--1.32) & 1.15 (0.90--1.46) & 1.64 (0.70--3.82) & 0.73 (0.28--1.90) & 0.54 (0.14--2.11) & 0.49 (0.09--2.80) \\
 & Ref.\ calibrated & 0.94 (0.74--1.19) & 0.97 (0.73--1.29) & 1.39 (0.59--3.29) & 0.62 (0.23--1.64) & 0.46 (0.12--1.80) & 0.42 (0.07--2.40) \\
 & ReClaim-L        & 1.11 (0.94--1.31) & 1.19 (0.94--1.49) & 1.86 (0.86--4.02) & 0.66 (0.25--1.72) & 0.67 (0.17--2.58) & 0.67 (0.13--3.47) \\
 & ReClaim-L calibrated & 0.97 (0.81--1.16) & 1.04 (0.82--1.32) & 1.63 (0.75--3.54) & 0.58 (0.22--1.51) & 0.59 (0.15--2.27) & 0.59 (0.11--3.05) \\
\cmidrule(l){1-8}
\multirow{5}{*}{YNHH}
 & Treated events   & 720 & 364 & 42 & 4 & 24 & 18 \\
 & Ref.             & 1.01 (0.90--1.14) & 1.07 (0.90--1.26) & 1.23 (0.76--1.99) & 0.26 (0.08--0.91) & 1.33 (0.70--2.51) & 0.55 (0.29--1.02) \\
 & Ref.\ calibrated & 1.04 (0.92--1.18) & 1.10 (0.93--1.30) & 1.28 (0.79--2.06) & 0.27 (0.08--0.94) & 1.37 (0.72--2.60) & 0.57 (0.30--1.06) \\
 & ReClaim-L        & 1.09 (0.97--1.21) & 1.13 (0.97--1.31) & 1.08 (0.69--1.68) & 0.38 (0.11--1.24) & 1.37 (0.76--2.49) & 0.62 (0.34--1.14) \\
 & ReClaim-L calibrated & 1.10 (0.98--1.23) & 1.14 (0.98--1.33) & 1.09 (0.70--1.70) & 0.38 (0.12--1.26) & 1.39 (0.77--2.52) & 0.63 (0.34--1.15) \\
\midrule
\multicolumn{8}{c}{\textbf{Comparator: DPP4i}} \\
\midrule
\multirow{3}{*}{PM}
 & Treated events   & 932 & 568 & 48 & 40 & 28 & 12 \\
 & Ref.             & 1.17 (1.07--1.29) & 0.97 (0.87--1.09) & 0.84 (0.57--1.23) & 0.93 (0.61--1.43) & 0.74 (0.45--1.19) & 0.63 (0.31--1.30) \\
 & Ref.\ calibrated & 1.12 (1.02--1.23) & 0.93 (0.83--1.04) & 0.80 (0.55--1.18) & 0.89 (0.58--1.37) & 0.70 (0.43--1.14) & 0.60 (0.29--1.24) \\
\cmidrule(l){1-8}
\multirow{5}{*}{MS}
 & Treated events   & 378 & 207 & 21 & 9 & 4 & 3 \\
 & Ref.             & 1.21 (0.97--1.51) & 1.03 (0.77--1.39) & 0.71 (0.31--1.66) & 0.41 (0.15--1.12) & 0.18 (0.05--0.67) & 0.15 (0.03--0.73) \\
 & Ref.\ calibrated & 1.14 (0.92--1.42) & 0.97 (0.72--1.31) & 0.67 (0.29--1.56) & 0.39 (0.15--1.05) & 0.17 (0.05--0.63) & 0.14 (0.03--0.69) \\
 & ReClaim-L        & 1.17 (0.95--1.44) & 1.06 (0.81--1.38) & 0.90 (0.40--2.01) & 0.35 (0.14--0.89) & 0.25 (0.07--0.90) & 0.21 (0.04--1.09) \\
 & ReClaim-L calibrated & 1.17 (0.95--1.44) & 1.06 (0.81--1.38) & 0.90 (0.40--2.01) & 0.35 (0.14--0.89) & 0.25 (0.07--0.90) & 0.21 (0.04--1.09) \\
\cmidrule(l){1-8}
\multirow{5}{*}{YNHH}
 & Treated events   & 712 & 359 & 40 & 4 & 26 & 18 \\
 & Ref.             & 0.98 (0.88--1.11) & 0.90 (0.77--1.05) & 1.14 (0.71--1.84) & 0.34 (0.11--1.07) & 0.99 (0.58--1.67) & 0.97 (0.50--1.86) \\
 & Ref.\ calibrated & 1.05 (0.90--1.22) & 0.95 (0.79--1.15) & 1.22 (0.75--1.98) & 0.37 (0.12--1.15) & 1.05 (0.62--1.79) & 1.03 (0.53--2.00) \\
 & ReClaim-L        & 1.01 (0.90--1.13) & 0.94 (0.81--1.09) & 1.13 (0.71--1.80) & 0.43 (0.13--1.38) & 0.99 (0.58--1.67) & 0.88 (0.45--1.74) \\
 & ReClaim-L calibrated & 1.07 (0.91--1.24) & 0.99 (0.82--1.19) & 1.19 (0.74--1.93) & 0.45 (0.14--1.47) & 1.04 (0.61--1.78) & 0.93 (0.47--1.85) \\
\bottomrule
\end{tabular}
\end{table}

\FloatBarrier
\clearpage
\begin{table}[H]
\centering
\caption{{Psychiatric outcomes of GLP-1RA initiation under 1:1 propensity-score matching.
Uncalibrated hazard ratios with 95\% confidence intervals for each outcome, by comparator, data
source and covariate specification. Confidence intervals are Wald intervals from Cox proportional
hazards models fitted in the matched cohort, with robust standard errors. Our estimates match 1:1
on each specification's own propensity score with a caliper of 0.2 standard deviations. ReClaim-L, the largest model, is shown for brevity. Treated
events are counted in each specification's own matched cohort, so they differ between specifications; outcomes are ordered by the reference study's treated
event count, from most to least frequent. Ref.\ denotes the reference paper's covariate
specification. Penn Medicine values are as reported by Tang et
al.~\cite{tang2025association}. MS, MarketScan 2-million combined held-out set; NE, not estimable
because a matched arm had no events; PM, Penn Medicine electronic health
records; YNHH, Yale New Haven Health electronic health records.}}
\label{tab:t4b_psm_sensitivity}
\scriptsize
\setlength{\tabcolsep}{2.5pt}
\renewcommand{\arraystretch}{1.15}
\begin{tabular}{@{}l l cccccc@{}}
\toprule
\textbf{Source} & \textbf{Specification}
 & \textbf{Anxiety} & \textbf{Depression} & \textbf{Other mood}
 & \textbf{Bipolar} & \textbf{Schizophrenia+} & \textbf{\shortstack{Suicidal \\ ideation}} \\
\midrule
\multicolumn{8}{c}{\textbf{Comparator: SGLT2i}} \\
\midrule
\multirow{2}{*}{PM}
 & Treated events & 644 & 401 & 36 & 28 & 25 & 9 \\
 & Ref.           & 1.17 (1.05--1.31) & 1.07 (0.93--1.23) & 1.16 (0.71--1.87) & 1.55 (0.86--2.80) & 1.56 (0.84--2.93) & 1.13 (0.43--2.92) \\
\cmidrule(l){1-8}
\multirow{4}{*}{MS}
 & \shortstack[l]{Treated events\\(Ref.)} & 244 & 133 & 11 & 7 & 3 & 1 \\
 & \shortstack[l]{Treated events\\(ReClaim-L)} & 213 & 110 & 14 & 6 & 3 & 0 \\
 & Ref.           & 1.17 (0.98--1.40) & 1.25 (0.98--1.61) & 1.18 (0.51--2.77) & 0.92 (0.33--2.52) & 1.04 (0.21--5.24) & 0.37 (0.04--3.48) \\
 & ReClaim-L      & 1.12 (0.92--1.36) & 1.07 (0.82--1.40) & 2.00 (0.81--4.98) & 0.85 (0.29--2.54) & 0.60 (0.14--2.53) & NE \\
\cmidrule(l){1-8}
\multirow{4}{*}{YNHH}
 & \shortstack[l]{Treated events\\(Ref.)} & 540 & 272 & 34 & 2 & 22 & 15 \\
 & \shortstack[l]{Treated events\\(ReClaim-L)} & 482 & 252 & 33 & 3 & 19 & 14 \\
 & Ref.           & 1.11 (0.98--1.25) & 1.09 (0.92--1.28) & 1.21 (0.73--1.99) & 0.25 (0.05--1.19) & 1.50 (0.78--2.92) & 0.67 (0.35--1.28) \\
 & ReClaim-L      & 1.04 (0.91--1.17) & 1.04 (0.87--1.24) & 1.00 (0.62--1.62) & 0.44 (0.11--1.70) & 1.64 (0.79--3.39) & 0.60 (0.31--1.16) \\
\midrule
\multicolumn{8}{c}{\textbf{Comparator: DPP4i}} \\
\midrule
\multirow{2}{*}{PM}
 & Treated events & 549 & 345 & 35 & 33 & 21 & 7 \\
 & Ref.           & 1.18 (1.05--1.34) & 1.01 (0.87--1.18) & 1.01 (0.63--1.61) & 1.44 (0.85--2.46) & 1.25 (0.66--2.36) & 0.64 (0.25--1.65) \\
\cmidrule(l){1-8}
\multirow{4}{*}{MS}
 & \shortstack[l]{Treated events\\(Ref.)} & 157 & 83 & 7 & 4 & 2 & 1 \\
 & \shortstack[l]{Treated events\\(ReClaim-L)} & 132 & 64 & 6 & 4 & 4 & 1 \\
 & Ref.           & 1.09 (0.87--1.36) & 1.07 (0.79--1.44) & 0.66 (0.26--1.70) & 0.41 (0.13--1.32) & 0.23 (0.05--1.09) & 0.21 (0.03--1.82) \\
 & ReClaim-L      & 1.06 (0.82--1.36) & 0.94 (0.66--1.32) & 0.83 (0.28--2.46) & 0.42 (0.13--1.38) & 0.64 (0.18--2.29) & 0.31 (0.03--2.87) \\
\cmidrule(l){1-8}
\multirow{4}{*}{YNHH}
 & \shortstack[l]{Treated events\\(Ref.)} & 492 & 257 & 33 & 4 & 21 & 14 \\
 & \shortstack[l]{Treated events\\(ReClaim-L)} & 438 & 226 & 26 & 4 & 20 & 11 \\
 & Ref.           & 0.98 (0.87--1.11) & 0.88 (0.75--1.04) & 1.05 (0.65--1.71) & 0.52 (0.15--1.74) & 0.88 (0.49--1.58) & 0.73 (0.37--1.45) \\
 & ReClaim-L      & 1.03 (0.91--1.17) & 0.94 (0.79--1.13) & 1.22 (0.70--2.12) & 0.53 (0.16--1.78) & 1.07 (0.57--2.00) & 0.85 (0.39--1.90) \\
\bottomrule
\end{tabular}
\end{table}

\FloatBarrier
\clearpage
\begin{table}[H]
\centering
\caption{{Covariate balance by specification, comparator and data source.
Number of covariates with an absolute standardized mean difference below 0.1 over the number of covariates evaluated, with the percentage in parentheses, before adjustment, after propensity-score weighting and after 1:1 matching, in new users of a GLP-1RA versus each comparator; percentages are rounded down, so that a cell reads 100\% only when no covariate exceeds 0.1. Reference covariates: the reference study's covariates (Supplementary Table~\ref{tab:rwe_application_reference_spec}), evaluated under the weights and matched set of every specification; the denominator is 64 in MarketScan and 65 in YNHH for every specification. Own covariates: the covariates of each specification's own propensity-score model (clinical covariates, hdPS dimensions or representation components); the denominator is the number of covariates in that model, 64 in MarketScan and 65 in YNHH for the reference specification, 507 for hdPS, 1{,}026 for ReClaim-S and 2{,}050 for ReClaim-M and ReClaim-L. hdPS, high-dimensional propensity score; MS, MarketScan 2-million combined held-out set; Ref.\ spec, reference specification; YNHH, Yale New Haven Health electronic health records.}}
\label{tab:balance_by_specification}
\footnotesize
\setlength{\tabcolsep}{4pt}
\renewcommand{\arraystretch}{1.15}
\begin{tabular}{@{}l l c c c c@{}}
\toprule
& & \multicolumn{2}{c}{\textbf{Reference covariates}} & \multicolumn{2}{c@{}}{\textbf{Own covariates}} \\
\cmidrule(lr){3-4} \cmidrule(l){5-6}
\textbf{Source} & \textbf{Specification} & \textbf{Weighted} & \textbf{Matched} & \textbf{Weighted} & \textbf{Matched} \\
\midrule
\multicolumn{6}{c}{\textbf{Comparator: SGLT2i}} \\
\midrule
\multirow{6}{*}{MS}
 & Unadjusted & 41/64 (64.0\%) & -- & -- & -- \\

 & Ref.\ spec & 64/64 (100\%) & 64/64 (100\%) & 64/64 (100\%) & 64/64 (100\%) \\

 & hdPS & 63/64 (98.4\%) & 63/64 (98.4\%) & 507/507 (100\%) & 507/507 (100\%) \\

 & ReClaim-S & 56/64 (87.5\%) & 62/64 (96.8\%) & 1{,}024/1{,}026 (99.8\%) & 1{,}026/1{,}026 (100\%) \\

 & ReClaim-M & 54/64 (84.3\%) & 63/64 (98.4\%) & 2{,}048/2{,}050 (99.9\%) & 2{,}050/2{,}050 (100\%) \\

 & ReClaim-L & 54/64 (84.3\%) & 61/64 (95.3\%) & 2{,}048/2{,}050 (99.9\%) & 2{,}050/2{,}050 (100\%) \\
\cmidrule(l){1-6}
\multirow{6}{*}{YNHH}
 & Unadjusted & 35/65 (53.8\%) & -- & -- & -- \\

 & Ref.\ spec & 65/65 (100\%) & 65/65 (100\%) & 65/65 (100\%) & 65/65 (100\%) \\

 & hdPS & 63/65 (96.9\%) & 64/65 (98.4\%) & 454/507 (89.5\%) & 507/507 (100\%) \\

 & ReClaim-S & 56/65 (86.1\%) & 61/65 (93.8\%) & 1{,}025/1{,}026 (99.9\%) & 1{,}026/1{,}026 (100\%) \\

 & ReClaim-M & 54/65 (83.0\%) & 61/65 (93.8\%) & 2{,}048/2{,}050 (99.9\%) & 2{,}050/2{,}050 (100\%) \\

 & ReClaim-L & 53/65 (81.5\%) & 60/65 (92.3\%) & 2{,}048/2{,}050 (99.9\%) & 2{,}050/2{,}050 (100\%) \\
\midrule
\multicolumn{6}{c}{\textbf{Comparator: DPP4i}} \\
\midrule
\multirow{6}{*}{MS}
 & Unadjusted & 42/64 (65.6\%) & -- & -- & -- \\

 & Ref.\ spec & 64/64 (100\%) & 62/64 (96.8\%) & 64/64 (100\%) & 62/64 (96.8\%) \\

 & hdPS & 57/64 (89.0\%) & 58/64 (90.6\%) & 499/507 (98.4\%) & 504/507 (99.4\%) \\

 & ReClaim-S & 56/64 (87.5\%) & 60/64 (93.7\%) & 1{,}014/1{,}026 (98.8\%) & 1{,}026/1{,}026 (100\%) \\

 & ReClaim-M & 56/64 (87.5\%) & 58/64 (90.6\%) & 2{,}039/2{,}050 (99.4\%) & 2{,}044/2{,}050 (99.7\%) \\

 & ReClaim-L & 54/64 (84.3\%) & 57/64 (89.0\%) & 2{,}047/2{,}050 (99.8\%) & 2{,}008/2{,}050 (97.9\%) \\
\cmidrule(l){1-6}
\multirow{6}{*}{YNHH}
 & Unadjusted & 39/65 (60.0\%) & -- & -- & -- \\

 & Ref.\ spec & 65/65 (100\%) & 65/65 (100\%) & 65/65 (100\%) & 65/65 (100\%) \\

 & hdPS & 64/65 (98.4\%) & 64/65 (98.4\%) & 506/507 (99.8\%) & 507/507 (100\%) \\

 & ReClaim-S & 61/65 (93.8\%) & 63/65 (96.9\%) & 1{,}025/1{,}026 (99.9\%) & 1{,}026/1{,}026 (100\%) \\

 & ReClaim-M & 61/65 (93.8\%) & 62/65 (95.3\%) & 2{,}049/2{,}050 (99.9\%) & 2{,}050/2{,}050 (100\%) \\

 & ReClaim-L & 58/65 (89.2\%) & 61/65 (93.8\%) & 2{,}049/2{,}050 (99.9\%) & 2{,}050/2{,}050 (100\%) \\
\bottomrule
\end{tabular}
\end{table}

\FloatBarrier
\clearpage
\begin{table}[H]
\centering

\caption{{Normalized Mutual Information (NMI) and Adjusted Rand Index (ARI) of Delphi and ReClaim learned disease embeddings against ICD-10-CM chapter labels. $\Delta$ is ReClaim minus Delphi.}}
\label{tab:clustering_metrics}
\begin{tabular}{lcccccc}
\toprule
& \multicolumn{2}{c}{Delphi} & \multicolumn{2}{c}{ReClaim} & \multicolumn{2}{c}{Difference} \\
\cmidrule(lr){2-3} \cmidrule(lr){4-5} \cmidrule(lr){6-7}
Representation & NMI & ARI & NMI & ARI & $\Delta$NMI & $\Delta$ARI \\
\midrule
UMAP & $0.442 \pm 0.001$ & $0.222 \pm 0.003$ & $0.524 \pm 0.002$ & $0.365 \pm 0.017$ & $+0.082$ & $+0.142$ \\
\textit{t}-SNE & $0.453 \pm 0.002$ & $0.234 \pm 0.004$ & $0.573 \pm 0.003$ & $0.479 \pm 0.028$ & $+0.120$ & $+0.245$ \\
\bottomrule
\end{tabular}
\end{table}

\FloatBarrier
\clearpage
\begin{table}[H]
\centering

\newcommand{\inputposttraincaption}{\textbf{Effects of input representation and post-training on disease prediction performance.} Mean ROC-AUC across 1,208 retrospective MarketScan disease endpoints. To account for differences in input representation, we further evaluated ReClaim-S using diagnosis-only trajectories matched as closely as possible to Delphi. Diagnosis-only ReClaim-S achieved a mean ROC-AUC of 0.672 before post-training, below Delphi (0.694), but improved to 0.737 after post-training for disease prediction. These results highlight the contribution of ReClaim's two-stage training framework beyond differences in input representation.}
\newcommand{\inputposttrainbody}{%
\begin{tabular}{lcc}
\toprule
\textbf{Input representation} &
\textbf{Pre-trained only} &
\textbf{Pre-trained + Post-trained} \\
\midrule
Diagnosis-only     & 0.672 & 0.737 \\
Full claims inputs & 0.616 & 0.743 \\
\bottomrule
\end{tabular}}

\caption{\inputposttraincaption}
\label{tab:input_posttrain}
\inputposttrainbody
\end{table}

\begin{table}[H]
\centering

\caption{{\textbf{Separate EHRShot comparison with ETHOS.} Values are mean disease-level ROC-AUCs. ETHOS and ReClaim are compared on the same 1,010 ICD-10-CM categories. The remaining 14 benchmark categories are absent from the ETHOS vocabulary. ETHOS is a frozen, publicly released pre-trained checkpoint. Monte Carlo (MC) inference samples 64 future trajectories per prediction point (temperature 1, top-$p$ 1) and scores each diagnosis by the fraction of trajectories containing it after the prediction point. Post-trained ReClaim uses diagnosis-token logits. Results for the full 1,024-endpoint benchmark are reported separately and are not directly comparable with the ETHOS result. All ReClaim, GPT-2 and LLaMA-3.1 training used MarketScan only. On the full benchmark, ReClaim-S post-trained inference required 39.9 GPU-seconds versus 20,160 for MC on the same H200 system and prediction points, approximately $505\times$ lower cumulative GPU time.}}
\label{tab:ethos_ehrshot}
\small
\begin{tabular}{llc}
\toprule
\textbf{Model} & \textbf{Inference} & \textbf{Mean ROC-AUC} \\
\midrule
\multicolumn{3}{l}{\textbf{a. ETHOS-overlap set: 1,010 endpoints}} \\
ETHOS & Pre-trained MC & 0.562 \\
ReClaim-S & Pre-trained MC & 0.628 \\
ReClaim-S & Post-trained logits & 0.733 \\
ReClaim-L & Post-trained logits & 0.753 \\
\midrule
\multicolumn{3}{l}{\textbf{b. Full EHRShot benchmark: 1,024 endpoints}} \\
GPT-2 & Post-trained logits & 0.716 \\
LLaMA-3.1 & Post-trained logits & 0.724 \\
ReClaim-S & Pre-trained MC & 0.627 \\
ReClaim-S & Post-trained logits & 0.733 \\
ReClaim-L & Post-trained logits & 0.753 \\
\bottomrule
\end{tabular}

\end{table}

\FloatBarrier
\clearpage
\subsection*{Supplementary Text}
\subsection{Data Processing}
To ensure comprehensive representation and compatibility across the three MarketScan payer databases, we selected a harmonized set of demographic, enrollment, clinical, and financial variables.
For demographic information, we collected biological sex and year of birth, as MarketScan does not provide additional demographic attributes such as race or ethnicity. For enrollment information, we extracted payer type, insurance plan type, indicator of capitated coverage, and state-level geographic location. 
Clinical information was collected according to claim type, including outpatient pharmaceutical, outpatient visit, and inpatient admission claims. Pharmacy claims included National Drug Codes (NDCs). Outpatient claims included diagnosis codes from the International Classification of Diseases, Tenth Revision, Clinical Modification (ICD-10-CM), and procedure codes primarily from Current Procedural Terminology, Fourth Edition (CPT-4). Inpatient admission records included diagnosis codes, procedure codes primarily based on the International Classification of Diseases, Tenth Revision, Procedure Coding System (ICD-10-PCS), length of stay, and discharge status. For financial variables, we used the gross payment amount associated with each claim.

We adapted the tokenization framework proposed in prior work \cite{renc2024zero,shmatko2025learning} to transform each enrollee’s longitudinal claims history into a structured, chronologically ordered token sequence suitable for autoregressive modeling. Missing or unavailable values across all variables were standardized and assigned the value \texttt{MISSING}. Tokens in the resulting sequence are categorized into two types depending on whether they carry explicit temporal information: static tokens and temporal tokens aggregated at a specified temporal resolution.

\paragraph{Temporal resolution and data aggregation}
Although the Merative MarketScan Research Databases record claims at daily resolution, latency and irregular reporting delays can distort the true temporal ordering of clinical events. Aggregation to a coarser temporal scale mitigates this noise; among daily, weekly, and monthly granularities, monthly aggregation yielded the best downstream performance in ablation studies. Accordingly, temporal event tokens were grouped by calendar month within each claim type. Within each month, tokens were aggregated either by summation or by deduplication based on first occurrence, then ordered according to their original finer-resolution timestamps (dates) when available. Finally, tokens occurring at the same timestamp were deterministically ordered according to the rules illustrated in Supplementary Table~\ref{tab:simultaneous_token_order}.

\paragraph{Static tokens}
Static tokens are time-invariant and positioned independently of temporal events. The sex token is treated as static and positioned at the start of the sequence. Additional static tokens include special instruction tokens that are manually inserted to signal the initiation of post-training tasks.

\paragraph{Enrollment token}
Each enrollment episode is represented by an enrollment-start token group and a corresponding enrollment-end token. The start group captures key attributes of the healthcare plan, consisting of an enrollment-start flag followed by tokens encoding plan type, capitated coverage status, and geographic location, all aligned to the calendar month of the enrollment start date. The enrollment-end token is aligned to the calendar month of the enrollment end date and marks the termination of the episode.

\paragraph{Clinical tokens}
\label{para:clinical_tokens}
MarketScan encodes clinical information using heterogeneous coding systems that vary across care settings and time periods. To enable unified representation learning and control vocabulary size, each code domain was standardized while preserving high mapping coverage and minimizing information loss. Diagnosis codes recorded in ICD-9-CM before 2015 were mapped to ICD-10-CM using the Centers for Medicare \& Medicaid Services General Equivalence Mappings. ICD-10-CM codes were then decomposed into up to three hierarchical components, major, minor and suffix, according to code granularity. Procedure codes from CPT-4, ICD-9 Procedure Coding System (ICD-9-Proc) and ICD-10-PCS were harmonized to Systematized Nomenclature of Medicine Clinical Terms (SNOMED CT) concepts using the Observational Health Data Sciences and Informatics Athena standard vocabulary (v20250827), with CPT-4 codes retained only when no SNOMED CT mapping was available, as this approach provides more consistent cross-mapping to ICD-10-PCS while yielding a smaller effective vocabulary. Drug product codes recorded as 11-digit NDC values were first mapped to RxNorm drug-level concepts and then collapsed to RxNorm active ingredients. Multi-target mappings (for example, one-to-many SNOMED CT procedure mappings or combination drugs) were enclosed by special combination start and end tokens to preserve atomic groupings. Tokens in each atomic group are sorted alphabetically to form a unique, canonical representation, enabling one-to-one mapping to the original code. Within each monthly aggregated encounter, repeated occurrences of the same standardized clinical code are collapsed by retaining the earliest instance, and the resulting unique codes are ordered by their original dates. Principal diagnosis and procedure codes are indicated by a preceding principal marker token, as are secondary diagnosis and procedure codes. Codes that fail to map to the standardized vocabulary are assigned the value \texttt{NOMAP}.

\paragraph{Cost token}
We used gross payment as the measure of healthcare expenditure in foundation model pre-training. Gross payment reflects the amount eligible for payment after applying pricing rules, such as fee schedules and discounts, but before deductibles, copayments and coordination of benefits. It therefore provides a stable estimate of expenditure magnitude that is closely tied to the recorded clinical event and less sensitive to patient-specific or plan-specific payment variation, which can differ substantially across insurance benefit designs and cost sharing arrangements. Incorporating this expenditure signal into longitudinal token sequences requires a representation that can accommodate the wide dynamic range of healthcare costs, which span several orders of magnitude from small pharmacy payments to high-cost inpatient stays. We therefore discretized expenditure into tokens that capture approximate magnitude. For each monthly timestep, gross payments within the same claim type were summed, rounded to one significant digit using standard half-up rounding, and encoded as a two-digit scientific-notation-style value, with the first digit representing the leading significant digit and the second representing the order of magnitude. For example, \$2,400 is rounded to 2,000 and encoded as 23, corresponding to $2\times10^3$, while \$859 is rounded to 900 and encoded as 92, corresponding to $9\times10^2$; zero or negative values are encoded as 0. This logarithmic discretization yields a compact vocabulary of 101 cost tokens covering values up to $9\times10^9$, including a token for missing expenditure. It preserves relative magnitude while assigning finer resolution to lower-cost events and coarser resolution to high-cost events, where order-of-magnitude differences are more informative. {We audited the discretization on the held-out evaluation split for the expenditure forecasting task (2.08 million enrollee-years): tokenized annual totals reproduce the exact totals reconstructed from the raw claims with $R^2$ 0.993, and the HNHC label differs for 0.42\% of enrollee-years (Supplementary Fig.~\ref{fig:discretization} and Supplementary Table~\ref{tab:discretization}), so the discretization is effectively lossless for the annual forecasting target.}

\paragraph{Interval and anchor tokens}
Temporal information is encoded through a combination of artificial time tokens and anchor tokens. Artificial time tokens $\langle\text{ATT-}N\rangle$ are inserted between consecutive monthly event-token groups, where $N$ denotes the number of elapsed months. To bound $N \in \{0,1,\ldots,12\}$ and maintain alignment with calendar time, a \emph{New Year} token $\langle\textsc{NY}\rangle$ is introduced at the first month of each calendar year, serving as a recurring calendar anchor. Additional anchor tokens provide absolute reference points: a year-of-birth token placed at the first month of the birth year, and an age token encoding the enrollee’s age at their first observed claim, assigned to the first month of that year. Together, these tokens encode time in a relative form while preserving the ability to recover absolute calendar time.
   
\paragraph{Vocabulary construction}
To control vocabulary size while maintaining adequate coverage, clinical code tokens were populated only from codes observed in the training data rather than from the full set of theoretically possible codes in the original coding systems. The vocabulary was further substantially compressed through the standardized mapping procedures described in the Clinical tokens paragraph above. For other token categories, the vocabulary was constructed as the union of values observed in the training data and all allowable values exhaustively defined in the MarketScan data dictionary. A summary of the resulting vocabulary is provided in Supplementary Table~\ref{tab:vocab_summary}.

\paragraph{Sequence assembly}
We organize temporal token groups into six major event categories that collectively define each enrollee’s longitudinal healthcare trajectory: (1) anchor events, including age tokens that encode temporal context and enable alignment to calendar time; (2) enrollment start, capturing plan attributes and coverage characteristics; (3) outpatient visits, including diagnosis and procedure information; (4) outpatient pharmaceutical visits, including prescription records; (5) inpatient admissions, including diagnosis, procedures, discharge status and length of stay; and (6) enrollment end. For each individual, tokens are assembled into a single longitudinal sequence that begins with static demographic tokens, such as sex, followed by temporally ordered predefined events aggregated at monthly resolution. Consecutive monthly events are separated by artificial time tokens encoding the number of elapsed months to the next observed event. Within each month, events and their associated tokens are deterministically ordered according to the hierarchy defined in Supplementary Table~\ref{tab:simultaneous_token_order}, ensuring a unique and reproducible sequence representation. This assembly yields a compact, information-complete representation that preserves temporal structure and healthcare event heterogeneity in a form suitable for autoregressive modeling. A synthetic example of the resulting token sequence is shown in Supplementary Table~\ref{tab:plain_token_sequence}, with a detailed step-by-step decoding provided in Supplementary Table~\ref{tab:example_walkthrough_decode} to illustrate how individual tokens map back to the original records.

The pre-trained disease embedding spaces of Delphi and ReClaim were visualized with UMAP (Supplementary Fig.~\ref{fig:embedding_umap}) {and \textit{t}-SNE (Supplementary Fig.~\ref{fig:embedding_tsne}), with broadly concordant patterns across the two methods.} Delphi produced a relatively diffuse arrangement {with several isolated satellite groups}, whereas ReClaim showed clearer large-scale organization. In the ReClaim {embedding space}, major ICD-10-CM chapters occupied distinct regions, {while} clinically related conditions formed compact local neighborhoods, indicating that pre-training on large-scale claims trajectories organizes diseases by both semantic similarity and shared longitudinal context. {This structure was further quantified by comparing the alignment between the learned representations and the ICD-10-CM chapter labels: ReClaim showed higher Normalized Mutual Information (NMI) and Adjusted Rand Index (ARI) than Delphi under both projections (UMAP, NMI 0.524 versus 0.442, and ARI 0.365 versus 0.222; \textit{t}-SNE, NMI 0.573 versus 0.453, and ARI 0.479 versus 0.234) (Supplementary Table~\ref{tab:clustering_metrics}).} Common disease endpoints were concentrated near the centers of chapter-level clusters, whereas rarer diagnoses were more often located at the periphery or in smaller satellite groups, consistent with a hierarchy extending from broad disease families to finer subtypes.

The ReClaim {embedding space} was not separated strictly by chapter. Selected groups showed partial continuity, including metabolic and cardiovascular disorders as well as respiratory and infectious diseases, suggesting that the model captures recurring cross-chapter relationships {reflected in longitudinal} care trajectories. Compared with Delphi's narrower disease-focused representation, ReClaim appeared to encode a broader view of longitudinal health states by integrating diagnoses, procedures, medications, and expenditure signals.

\end{appendices}

\end{document}